\PassOptionsToPackage{hypertexnames=false}{hyperref}
\documentclass[12pt]{report}

\usepackage{tikz}
\usepackage{amsmath}
\usepackage{amssymb}
\usepackage{caption}
\usepackage{subcaption}
\usepackage{enumitem}
\usepackage{hyperref}
\usepackage{fancyvrb}
\usepackage{graphics}
\usepackage{graphicx}
\usepackage{rotating}
\usepackage{xcolor}
\usepackage{multirow}
\usepackage{booktabs}

\usepackage{amssymb}
\usepackage{pifont}
\usepackage[numbers,sort&compress]{natbib}

\usepackage{NMTthesis2020}
\usepackage[hypertexnames=false]{hyperref}  

\newcommand{\FIXME}[1]{{\color{black}#1}}
\newcommand{\FIX}[1]{{\color{black}#1}}
\newcommand{\sm}[1]{{\color{black}#1}}

\hypersetup{
        colorlinks = true,
        linkcolor = black,
        anchorcolor = black,
        citecolor = black,
        filecolor = black,
        urlcolor = black,
	    pdfnewwindow = true,
        extension = pdf
} 
\dissertation
\author{Sushil Thapa}

\title{On effects of Knowledge Distillation on Transfer Learning
}
  
\degreeSought{Master of Science in Computer Science\\
  } 
  
\thesiscompletiondate{August, 2022}

\graduationdate{August, 2022}

\academicadvisor{Subhashish Mazumdar \textit{(Affiliation: NMT)}}
\coresearchadvisors{Subhashish Mazumdar \textit{(Affiliation: NMT)}}{Tanmoy Bhattacharya \textit{(Affiliation: LANL)}}
\committeesize{5}

\committeememberone{Benjamin Hamilton McMahon \textit{(Affiliation: LANL)}}
\committeemembertwo{Hamdy Soliman \textit{(Affiliation: NMT)}}
\committeememberthree{Ramyaa Ramyaa \textit{(Affiliation: NMT)}}
\begin{document}
%
%
%
%
\titlepage


\begin{dedication}
This thesis is dedicated to my family for their endless love and support. 
\end{dedication}

\begin{acknowledgments}
I would like to express my sincere gratitude to my academic and research advisor Dr. Subhashish Mazumdar for supporting my research and providing continuous guidance throughout my Master's study. Second, I would like to thank my research advisor, Dr. Tanmoy Bhattacharya from Los Alamos National \-Laboratory for his unparalleled research mentorship and guidance. His immense knowledge, research ideas, amazing thought processes, and ability to explain complex topics in simple words have been very valuable in this research. I thank my advisors for their empathy, motivation, encouragement, and patience with me.

Besides my advisors, I would like to thank all my committee members - Dr. Benjamin McMahon from Los Alamos National Laboratory, Dr. Hamdy Soliman, and Dr. Ramyaa for agreeing to sit on my committee and providing very \-insightful feedback and comments. All the comments provided by each committee member were very helpful in developing a coherent story and \-polishing my thesis.

I would like to especially acknowledge Dr. Benjamin McMahon for acting as the mediator for the collaborative research between New Mexico Tech and Los Alamos National Laboratory and making this research possible. His valuable advice on research directions and insights from a different research perspective helped push my research to be more robust and generalizable.

Lastly, I would like to thank Dr. Judith D. Cohn from Los Alamos National Laboratory for her invaluable time discussing my research and providing astute suggestions based on her amazing research experience.

\end{acknowledgments}
%
%
\begin{abstract}
Knowledge distillation is a popular \sm{machine learning} technique that aims to transfer knowledge from a large `teacher' network to a smaller `student' network and improve the student's performance by training it to emulate the teacher. In recent years, there has been significant progress in novel distillation techniques that push performance frontiers across multiple problems and benchmarks. Most of the \sm{reported} work focuses on achieving state-of-the-art results on the specific problem. However, there has been a significant gap in understanding the process and how it behaves under certain training scenarios. Similarly, transfer learning \sm{(TL)} is an effective technique in training neural networks on a limited dataset faster by reusing representations learned from a different but related problem. Despite its effectiveness and popularity, there has not been much exploration of knowledge distillation on transfer learning. \sm{In this thesis, we propose a machine learning architecture we call TL+KD that combines knowledge distillation with transfer learning; we then present a quantitative and qualitative comparison of TL+KD with TL in the domain of image classification.} Through this work, we show that using guidance and knowledge from a larger teacher network during fine-tuning, we can improve the student network to achieve better validation performances like accuracy. We characterize the improvement in the validation performance of the model using a variety of metrics beyond just accuracy scores, and study its performance in scenarios such as input degradation. 

\keywords{Deep Learning; Transfer Learning;
Knowledge Distillation;\hspace*{0pt plus 1fil} \linebreak \hspace*{63pt}Interpretability; Image Classification; Vision Transformers}
\end{abstract}

\tableofcontents
%
%
\listoftables
%
\listoffigures

%
\listofabbrs
%
\signaturepage
\chapter{Introduction}
In recent years, deep learning has become the gold standard in solving various machine learning problems. \label{chap:intro}Neural networks have been very effective in learning various modalities/types of data \sm{like speech~\cite{pmlr-v48-amodei16}, image~\cite{goodfellow2014generative}, text~\cite{abstention_ood}, etc.} The data-intensive nature of deep learning makes neural networks trained with large datasets capable of learning rich representations that often generalize well. However, smaller networks trained on smaller datasets are considered less equipped with the capacity to learn similar rich representations~\cite{mhaskar2017and, Brutzkus2019WhyDL, bubeck2021universal}. Similarly, another major challenge for the deep learning solution is the availability of sufficiently good quality data, which is very expensive, time-consuming, and difficult to obtain in some scenarios, such as medical diagnoses.
Since advanced compute and massive data are not accessible to everyone, we can pre-train a network and reuse it to fine-tune smaller data multiple times when needed with Transfer Learning~\cite{yosinski2014transferable}. It could be a data-efficient way to achieve faster training and inference on problems with limited available resources. Such a method has already been widely applied in different modalities of datasets such as images~\cite{pmlr-v27-bengio12a}, language~\cite{devlin-etal-2019-bert}, genomics~\cite{arik2019tabnet}, audio~\cite{Lan2020ALBERT}, etc. For example, a network can be pre-trained with a large dataset like ImageNet~\cite{imagenet} once, and its feature extractor or representations can be reused to learn another smaller dataset like MNIST~\cite{lecun-mnisthandwrittendigit-2010} effectively. 

Similarly, a large previously trained network, called the teacher network, can also be used to teach another network, called the student network, by emulating the teacher through Knowledge Distillation~\cite{hinton_kd}. This helps a smaller student network achieve the higher performance of a larger teacher network. \sm{This process works across many domains, modalities, or problems already like image classification~\cite{chen2020big}, audio recognition~\cite{Chen_2021_CVPR}, image captioning~\cite{Rennie2017SelfCriticalST}, etc.} This addresses the problem of requiring large computation by providing a way to combat the requirement to train large networks every time we need them. 

Despite the successes in both research directions, to our knowledge, there have not been \sm{any} attempts to apply KD to improve transfer learning performance or to find empirical evidence of how KD behaves in certain experimental settings. \sm{Some of the papers that explore knowledge distillation for training image classifiers from scratch without transfer learning are discussed in the Related work section~\ref{related_work_kd}}. In this thesis, we use knowledge distillation with transfer learning to obtain higher validation performance over the vanilla transfer learning setup for image classification. We also explore how knowledge distillation with transfer learning affects the quantitative and qualitative performance of transfer learning. To characterize gains and performance, we employ a variety of metrics beyond accuracy scores, such as confusion matrix, interpretations, \sm{number of epochs needed for training}, Precision, Recall, and F1 score. We choose image classification here for a couple of reasons: 1) It would be easier to adopt it to other classification problems such as tissue prediction with genes as input \sm{because both problems are similar i.e. classification}. Tissue prediction was the problem we worked on initially for a long time but did not choose for the thesis because of the lack of an intuitive way to get, understand, and verify determinants of classification like in the image. 2) Variants of image classification datasets are publicly available, and the image domain is popular as a benchmark for classification. 3) Pioneering research papers on our thesis for both knowledge distillation~\cite{hinton_kd, kd_survey} and transfer learning~\cite{pmlr-v27-bengio12a} also use image classifiers. 

In this thesis, Chapter~\ref{chap:intro} explains the background of the techniques used in this thesis and the proposal and motivation of the thesis. Chapter~\ref{chap:related_work} explains the previous work in this research direction and introduces our contribution to addressing the literature gap. Chapter~\ref{chapter:dataset} introduces and provides details of the datasets used in this thesis. Chapter~\ref{chap:baseline_model} introduces the progression, motivation, and details of the models used. Chapter~\ref{chap:experiments} first explains the framework of our application of knowledge distillation in transfer learning. Then, it provides details of the training to test each of the hypotheses listed in Section~\ref{hypothesis_list}. In chapter~\ref{chap:results} we present our findings and discuss the results based on the list of research questions related to the list of hypotheses in Section~\ref{hypothesis_list}. Chapter~\ref{chap:conclusion} summarizes our results again based on the same hypotheses. Finally, Chapter~\ref{chap:future_work} lists some of the research directions we think would be worth pursuing in the future. For simplicity, vanilla transfer learning is sometimes referred to as TL, and transfer learning with knowledge distillation is referred to as TL+KD in subsequent chapters. 

\section{Background}
\label{sec_background}
We now describe the background required for this thesis and explain the ideas used throughout the paper. First, we introduce machine learning and describe its working principle. Then we explain the need for deep learning and introduce its significance. We then go through Transfer Learning, its motivation, advantages, and the process of fine-tuning the model. We then explain the Knowledge Distillation process and its building blocks. We also discuss some of the data-augmentation techniques that will be used throughout the experiments. Finally, we explain the motivation for interpretability and use cases. We give some intuitive examples to help us understand the popular and effective interpretation techniques used in this thesis. 

\subsection{Machine Learning}
Artificial intelligence is a tool or technique for building machines as intelligent as humans (if not more). Machine learning is a branch of artificial intelligence that addresses this problem by allowing machines to learn from data without explicit programming. For example, let us take a chair recognition problem; recognizing a picture of a chair might not seem like intelligence to us because it is so effortless for human beings, but it is very difficult to teach the same idea to a computer because there could be various types of the chair in the real world. With machine learning, instead of defining a set of rules for what a chair should look like, we use images of chairs to teach the machine to learn what the chair looks like with examples. Machine learning is used in so many daily applications that we use today, ranging from email spam classification to face recognition to digital assistants like Siri, Alexa, etc. It has been used successfully to progress on problems like recommendation systems, self-driving cars, disease prediction, language translation, etc. 

There are different types of machine learning systems divided into supervised learning, unsupervised learning, and reinforcement learning. In supervised learning, we have an input dataset \sm{with associated labels for all of the samples. An example of this dataset would be a list  of movie summaries with their genre. The task of the machines is to look at the summary of a sample and its genre to learn an association between the sample and its genre. Next time, when the trained machine is given a movie summary, it learns to predict the genre automatically.} In unsupervised learning, we simply have input data without labels.
An example would be just the summary of a movie without genre, and we need to infer the genre type from the summary. In reinforcement learning, the machine learns with feedback or reward for its actions. An example would be a system that learns to play chess automatically by winning or losing~\cite{sutton2018reinforcement,mnih2013playing}. Recently, self-supervised learning is also emerging as a hybrid type that supervises itself by generating pseudo-labels~\cite{chen2020simple, he2020momentum}.

\begin{figure}[!htb]
\centering
\includegraphics[width=0.85\linewidth]{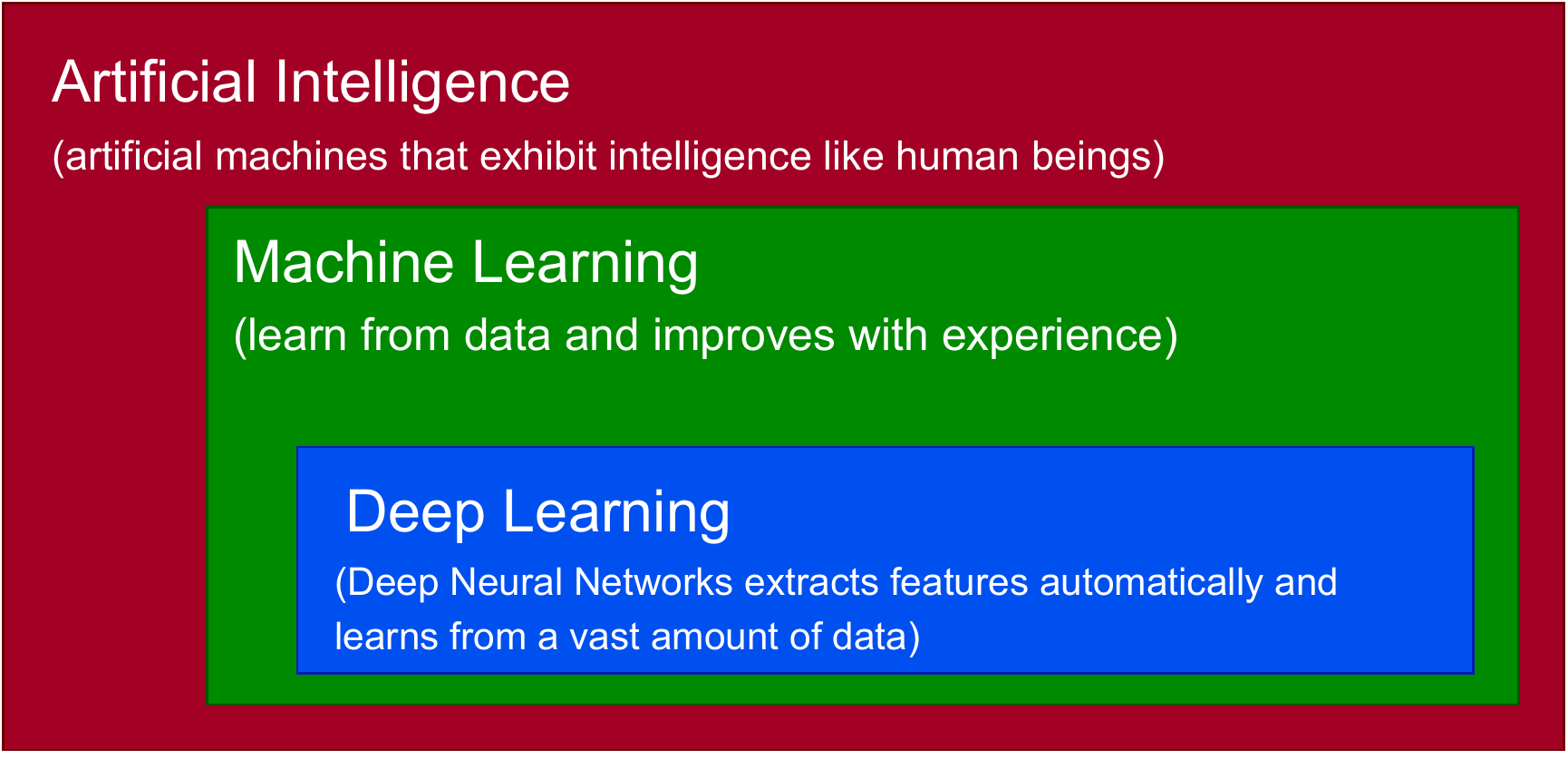}
\caption[Hierarchy representation of artificial intelligence, machine learning, and deep learning]{Hierarchy representation of artificial intelligence, machine learning, and deep learning. Artificial intelligence is an umbrella term that describes the ability to learn and reason like humans. Machine learning is a possible way to build an AI system that learns without explicit programming. Deep learning is a special case of ML in which we use deep models \sm{with many layers}, such as deep neural networks, to learn representations directly from huge data. This figure is adapted from~\cite{ai_ml_dl}.}\label{fig:ai_ml_dl}
\end{figure}

Let us discuss some of the terminology used in machine learning and used in different sections of this thesis later. In supervised learning, we have data with data samples and \sm{their target labels}. Based on those data and targets, we build a model that learns the association of those data. A model can be considered a black box or a function that takes data samples as input and gives predictions \sm{on what the target of those samples looks like}. Such a black box can be of different types, natures, and sizes. One simple example of a model can be as simple as \sm{looking at the closest subset of neighbors  and returning a class that is repeated the most in that subset}. Machine learning generally works when raw datasets are preprocessed to get simpler features. For example, images could be passed through an edge detector to get simpler features as edges compared to raw values and passed on to machine learning algorithms. 

The machine learning process in general can be summarized into 7 main steps~\cite{keytodatascience_2020}.
\begin{enumerate}
    \item Data collection
    \item Data cleaning and feature extraction 
    \item Model selection
    \item Model training
    \item Model evaluation
    \item Hyper-parameter tuning and model improvements
    \item Model inference
\end{enumerate}

In supervised learning, a dataset is divided into three parts, i.e., training set, validation set, and test set. We teach the model by showing the examples from the training set and using the unseen validation set to validate the learning of the model during training. Test sets \sm{are generally predefined to report benchmark results for a dataset. Generally, a validation set is used to validate the changes we make in the model during training whereas a test set is only used after the model is in its final form to calculate final accuracy.} If the splits are not already divided in datasets, random splits are done maintaining the ratio of classes in each split. Sometimes validation is optional; in that case, the model is validated and tested on the test split.

\begin{figure}[!htb]
\centering
\includegraphics[width=0.85\linewidth]{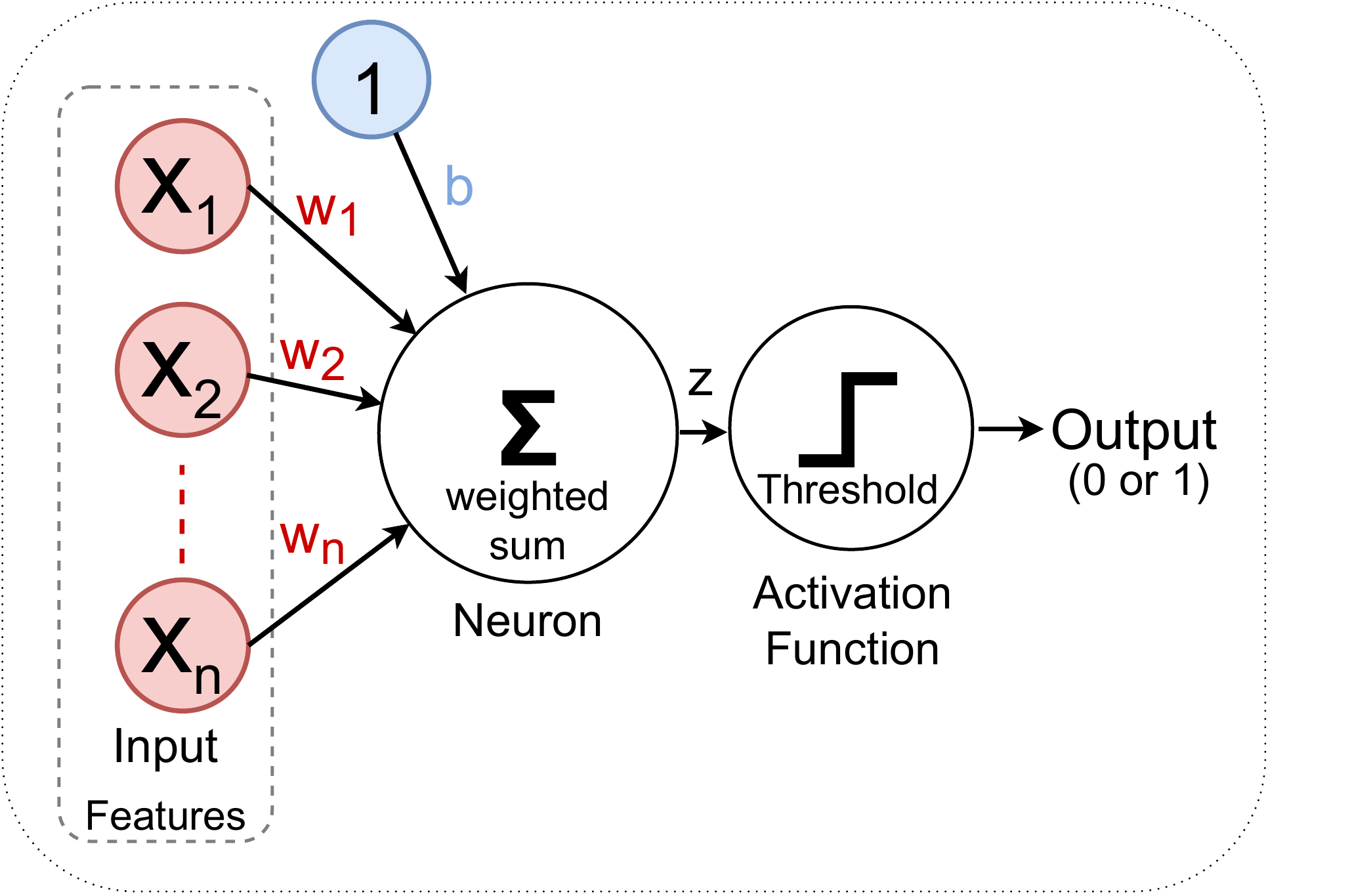}
\caption[Example of Perceptron network]{Example of Perceptron network. Input \sm{features} \{$X_1, X_2, .. X_n$\} are multiplied with corresponding weights \{$w_1, w_2, .. w_n$\} and added together with bias value $b$ to get a value $z$. $z$ is passed through a simple activation function that compares the values to a threshold. If the score $z$ is higher than the threshold value, it fires and outputs a 1. Otherwise, the output is 0. Figure is adapted from~\cite{Rosenblatt58theperceptron}.}\label{fig:perceptron_network}
\end{figure}

One example of a model is Perceptron~\cite{Rosenblatt58theperceptron} which works by finding the value of the weighted sum of all features and seeing if that value is enough to cross the threshold as summarized in Figure~\ref{fig:perceptron_network}. Perceptron has a node also called neuron and two types of values in the model which need to be learned called weights and bias. Input is multiplied with weights and added with a bias to get \sm{raw outputs $z$}. Weights provide individual \sm{strength} to inputs whereas \sm{bias adds or declines the value $z$ by a fixed value}. Raw outputs are then passed through an activation function like the unit step function to get an output. If the output does not match the \sm{real labels from dataset}, we update the weights of the layer continuously till it matches the output. This process of finding the optimal value of weights and biases forms \sm{the basis of training the model.}

\begin{figure}[!htb]
\centering
\includegraphics[width=0.65\linewidth]{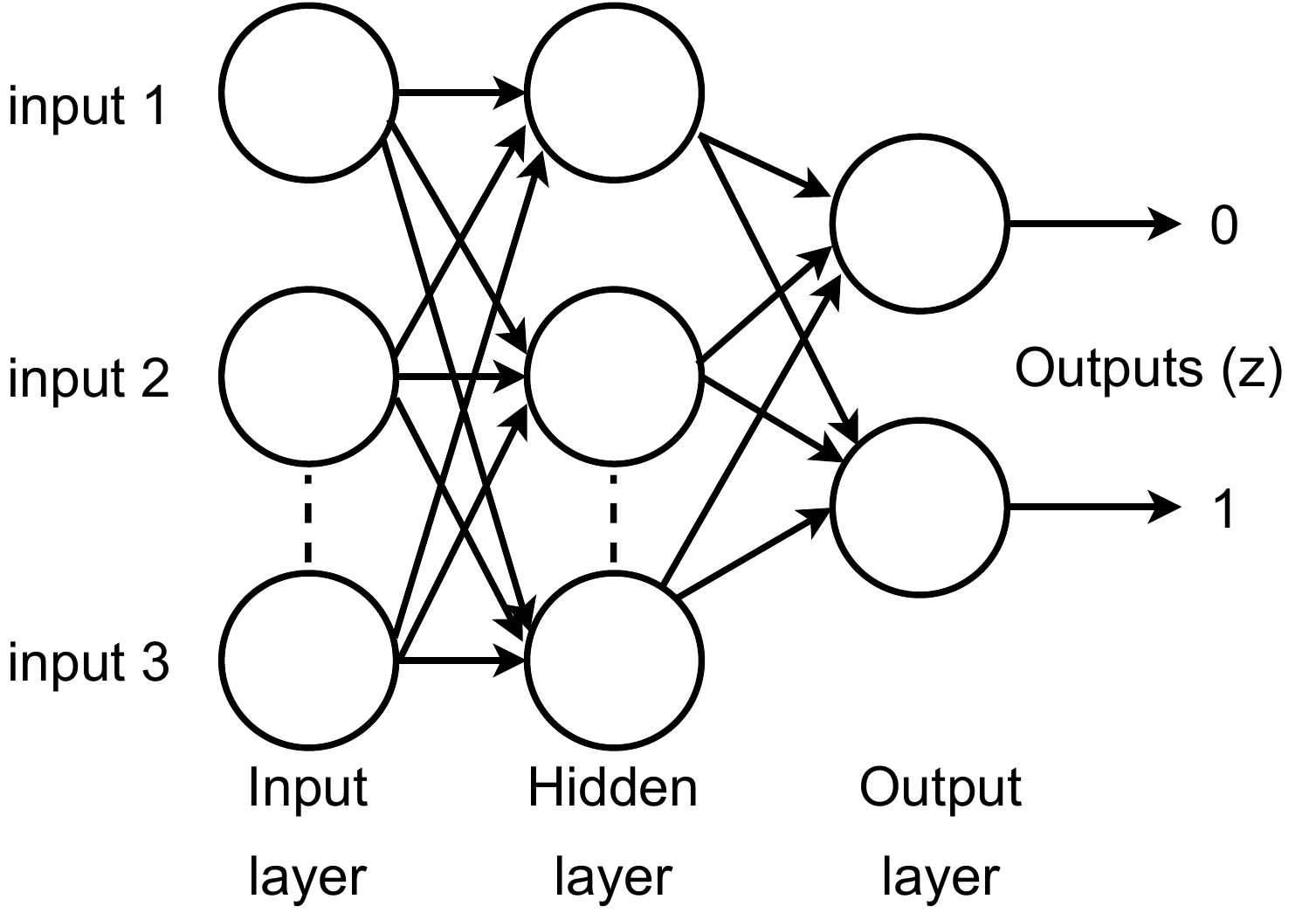}
\caption[Example of Multilayer Perceptron Network]{Example of Multilayer Perceptron Network. Inputs are treated as nodes in the input layer. They are connected to a hidden layer which in turn is connected to the output layer. Each node in a subsequent layer is connected to every node in previous layers. Each node in the output layer is treated as output. There can be any number of hidden layers more than 1 with any number of nodes in multilayer perception. The number of nodes in output and input layers is fixed and depends on the dataset and task at hand. This figure is adapted from~\cite{HASHEMIFATH202080}.}\label{fig:multi_perceptron_network}
\end{figure}

\sm{Perceptron can only learn linear functions~\cite{Rosenblatt58theperceptron}. To remove this inability of Perceptron to learn nonlinear function}, multilayer perceptron network~\cite{haykin1994neural} was introduced. Perceptron is also known to be roughly based on biological neural networks. There are three types of layers of nodes here. Input nodes in input layers are connected to input whereas output nodes in output layers are connected to outputs. This network is also called a fully connected neural network because all the nodes in a layer are connected to every node before and after the layer. \sm{The number of nodes in the output layer is the same as the number of classes in the dataset.} A variable number of hidden layers with variable hidden nodes are connected with input \sm{simultaneously}.
Weights and biases are used in the same way as perceptron above. \sm{Each connection arrow in Figure~\ref{fig:multi_perceptron_network} has a weight value associated with it. Each node in the hidden layer has a bias value associated with it.} A non-linear activation function like \sm{Rectified Linear Unit, also called, ReLu~\cite{nair2010rectified}} is used here. ReLu is a \sm{piecewise-linear} function that ignores negative values formulated as $z=max(0, z)$. \sm{Each output of hidden layers is passed through the activation function.} The output layer is normally selected to be the softmax layer which converts the set of values to a different set of probabilities whose sum adds to 1. 

To update weights and biases, an algorithm called stochastic gradient descent (SGD)~\cite{10.1214/aoms/1177729586} is used in multilayer perceptron. Once we obtain the prediction ($z$) after a forward pass from input to output, \sm{the actual labels ($y$) and prediction ($z$) are compared to obtain the error. The error is then propagated through each layer of the network to update the values of weights and bias.}
By calculating the contribution of each weight and biases in each layer to the error in the form of gradients through partial differentiation, the weights and biases are updated. The learning rate, generally less than 1, \sm{controls the step size of the update of the model's weights and biases}. This whole process is called backpropagation explained further in~\cite{10.1214/aoms/1177729586}. \sm{In SGD, instead of doing backpropagation using the whole dataset at once, we use a random sample of the dataset to calculate the estimation of total error at a time. This forward and backward pass is repeated as specified by iterations and epochs in SGD. Updating the model using all the datasets once is, one epoch whereas updating the model once for subset of input samples specified by batch size is one iteration. Sometimes, momentum is a term in SGD that accelerates parameter updates in SGD in a relevant direction.} Hyperparameters like the number of layers, the number of nodes in each layer, choice of activation function, etc. are predetermined before training and are selected by the user. In contrast, parameters like weights and biases are learned by the network during training. 

\begin{figure}[!htb]
\centering
\includegraphics[width=\linewidth]{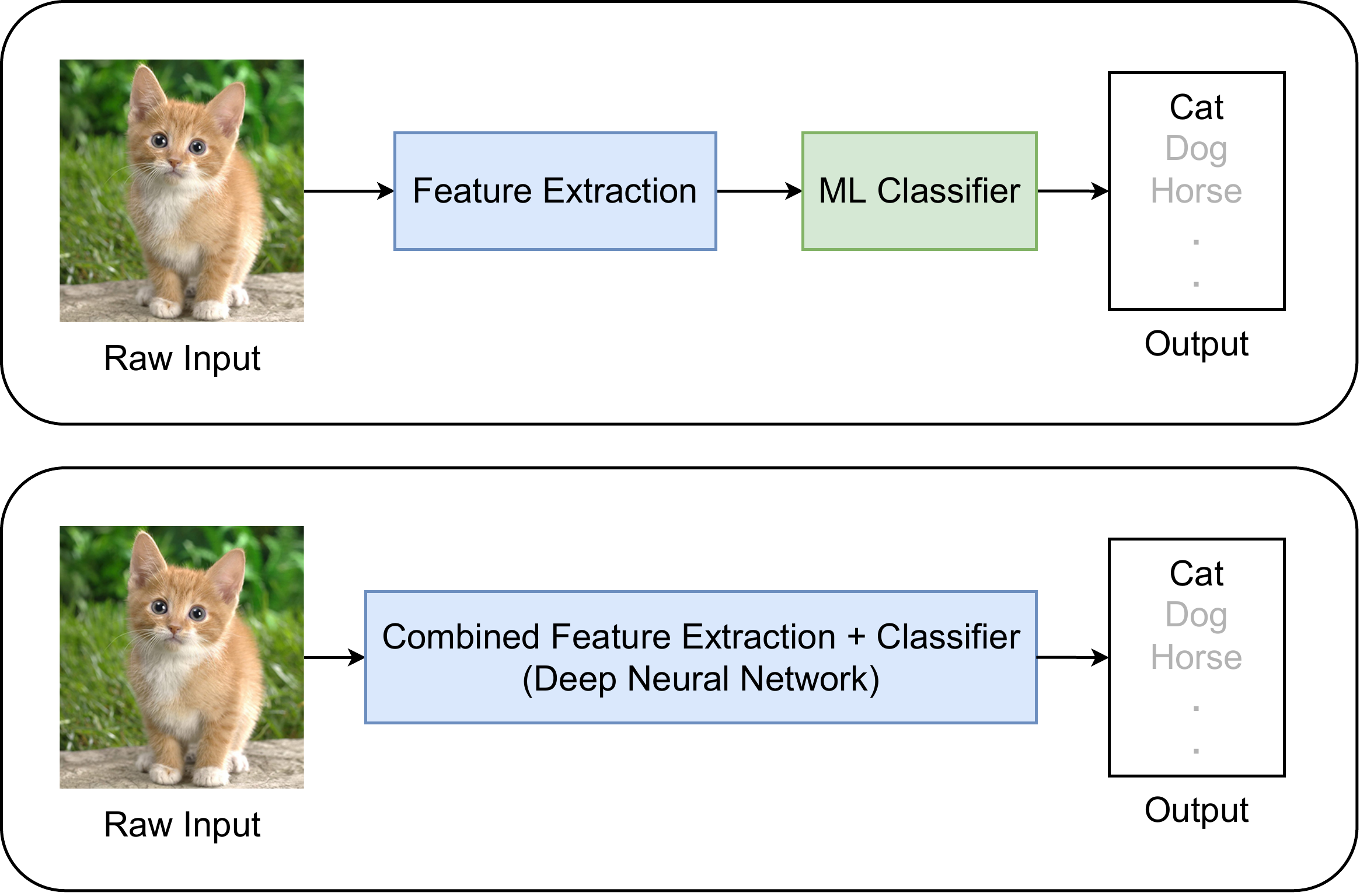}
\caption[Difference between ML and DL based on Representation Learning]{Difference between ML and DL based on Representation Learning. The upper section represents the ML pipeline, where feature extraction is done separately before passing the features to the ML algorithm. The lower section represents how deep neural networks work well with raw data and have a feature extractor built into the algorithm. This figure is adapted from~\cite{dl_ml}.}\label{fig:mlvsdl}
\end{figure}

Deep learning is a special case of machine learning that uses large-scale neural networks. With the advancement of computing resources and massive datasets, deeper neural networks have become a go-to method for solving problems. Deep learning automatically learns features without requiring feature extraction, such as general machine learning, as shown in Figure~\ref{fig:mlvsdl}. In summary, artificial intelligence, machine learning, and deep learning and their hierarchy are shown in Figure~\ref{fig:ai_ml_dl}. 
\sm{Details of the deep learning architectures used in this thesis will be explained} later in chapter~\ref{chap:model}.

\begin{figure}[!htb]
\centering
\includegraphics[width=\linewidth]{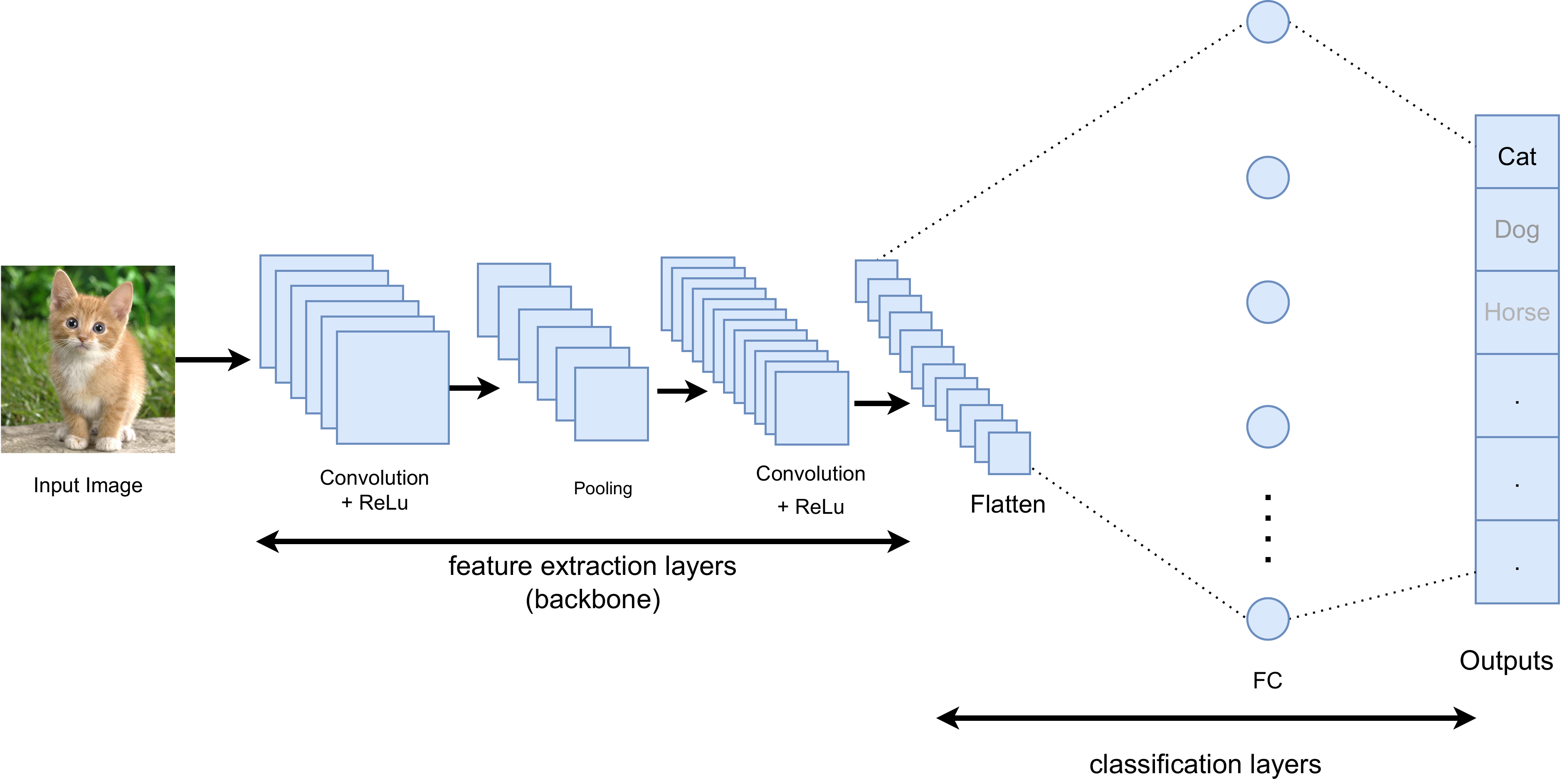}
\caption[Convolutional Neural Networks block diagram]{Convolutional Neural Networks block diagram. Input is passed through multiple convolutional layers with max-pooling to get the features and learn a representation of the image. Such representation is then flattened to get a one-dimensional array of images passed through fully connected layers, also called classification layers. Finally, the classifier gives a prediction for all classes, and the class with the highest output score is selected as the prediction class of the model. There can be any number of Convolution+pooling layers in a backbone or FC classification layers in a CNN network, and it varies according to the architecture of CNN. This figure is adapted from~\cite{lecun1998}.}\label{fig:cnn_network}
\end{figure}

Images are used as input to deep neural networks in our thesis. Images are nothing but three arrays of pixel intensity values for red, green, and blue channels.  Such input is used by a type of neural network called convolution neural network (CNN), as shown in Figure~\ref{fig:cnn_network}. Details on types of CNN and descriptions of different architectures are given in Chapter~\ref{chapter:model}. Instead of fully connected layers like before, CNN has convolutional layers which \sm{moves a fixed-sized matrix called a \textit{kernel} over images and multiplies with the subsection of image and obtains a value for each move. The combination of \textit{kernel} applied across different input channels is called a \textit{filter}. Learning the filter parameters and applying them over images extracts different features from the images.
A parameter, called stride, controls the number of pixels the filter is moved during convolution}. Sometimes we add pixels called \textit{padding} around the images to pass the filter over the edges. 

When convolution is applied in an input, we get an output matrix from the convolution layer followed by a nonlinear activation layer. \sm{Instead of a single weight connection in perceptron, the filter matrix here has multiple weight parameters the model learns during training. For a single filter, we learn \textit{(number of previous channels * height of filter * width of filter + 1)} parameters where 1 is for the bias term for each filter. Such filters are applied multiple times on an input which also multiplies the number of parameters.} To decrease the spatial size of output, we apply a max-pooling layer to decrease the input size by half and apply dimensionality reduction. \sm{The combination of convolution, ReLu, and max-pooling is applied multiple times on the image to obtain features. The number and size of such layers define the architecture of the network. Once we select an architecture for the feature extraction layer, we flatten the output array into a 1-dimensional vector which is now input to our fully connected classification layers. FC layer essentially classifies these flattened features into different classes, as shown in Figure~\ref{fig:cnn_network}.}

\begin{figure}[!htb]
\centering
\includegraphics[width=0.8\linewidth]{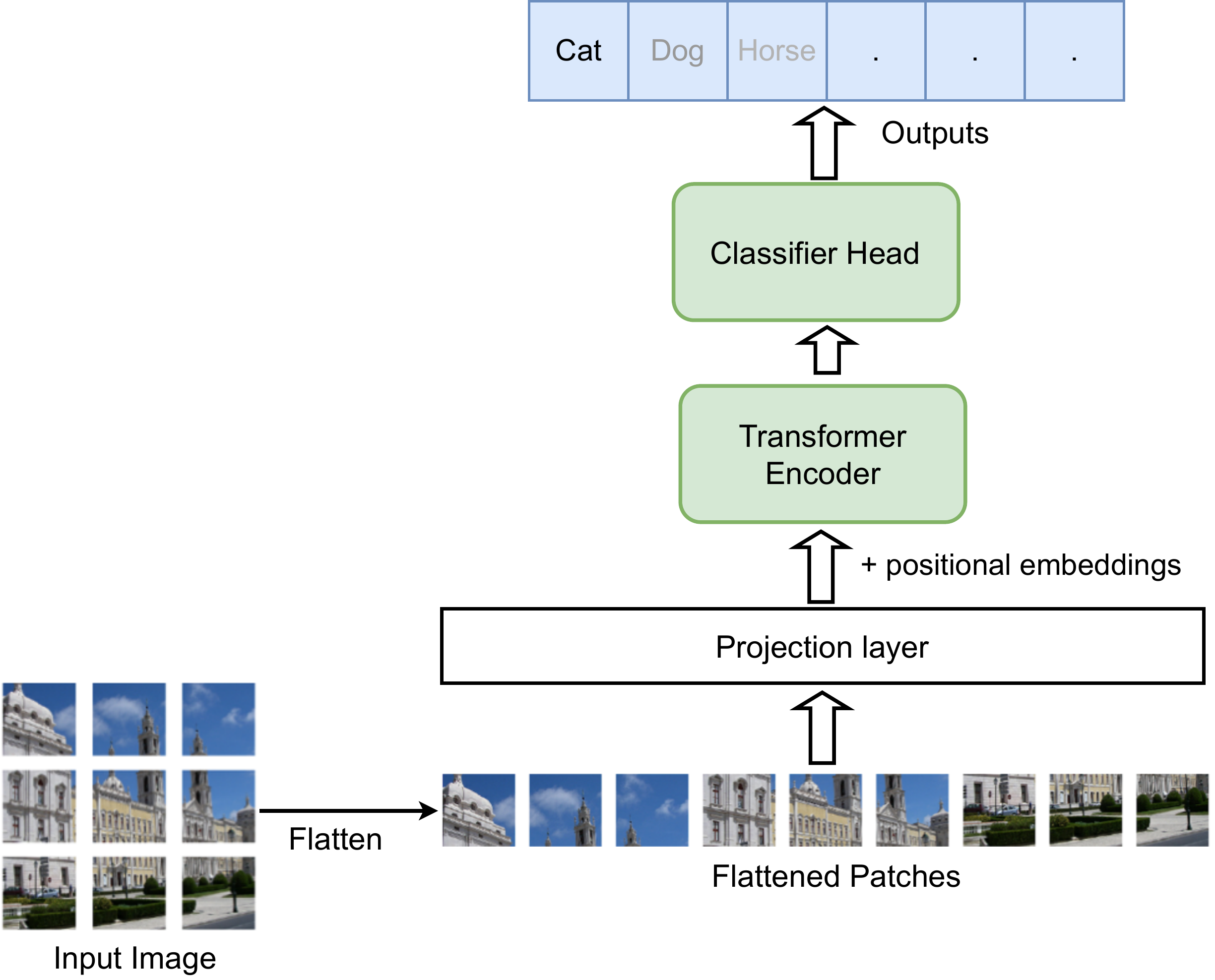}
\caption[Vision Transformers Networks block diagram]{Vision Transformers Networks block diagram. The image is divided into nine patches and is flattened in a 1-dimensional fashion. Flattened patches are passed through the projection layer to get nine smaller image representations. All of them, along with the position embedding, are passed through the transformer layer to encode the learned representation. Multiple transformer encoders can be used in a vision transformer. The encoded representations are then passed through a classifier head like a neural network that outputs the prediction. This figure is adapted from ViT paper~\cite{dosovitskiy2020imagevit}.}\label{fig:vit_network}
\end{figure}

Another network called vision transformer is used to achieve high accuracy for image classification tasks. Firstly, the input image is divided equally into $(n*n)$ tokens. Figure~\ref{fig:vit_network} shows an \sm{n=3 i.e., 3*3 token} for simplicity, but we employ 16*16 tokens in our experiments. Doing this allows us to use fewer parameters to learn the representation of an image. \sm{Such tokens or patches are then flattened to get 1-dimensional vectors. We pass these vectors through the projection layer with fully connected layers to get smaller vectors hence reducing the size. Each of these projected vectors is combined with positional embedding that provides position information to all tokens. The middle encoder network, as shown in Figure~\ref{fig:vit_network}, is based on a Transformer~\cite{transformer} architecture. The transformer encoder transforms the projected input into a list of vectors that define the significance of all tokens onto a certain token. Doing this helps to extract a rich representation of each token with respect to other tokens. Lastly, the encoded representations are passed through a classifier head made up of a fully connected neural network that produces the output.} 

\subsection{Transfer Learning}
Transfer learning is a process of reusing the entire or part of a pre-trained network trained on the first task as a starting point of training for a second related task. First, a base model is trained on a base data set for a base task, such as image recognition. Next, as with lifelong learning algorithms~\cite{Thrun_1998}, the learned features are repurposed to be transferred to a second target model. This process is generally effective when the initial task is similar to the final task, and the learned features of one model are relevant to be used as features for the second network. Therefore, by applying transfer learning to novel tasks with a limited dataset, one can perform better than training on such limited data from scratch. Unlike traditional machine learning algorithms, using transfer learning in deep neural networks avoids training separate models independently, making it computationally efficient.  

\begin{figure}[!htbp]
\centering
\includegraphics[width=4.5in]{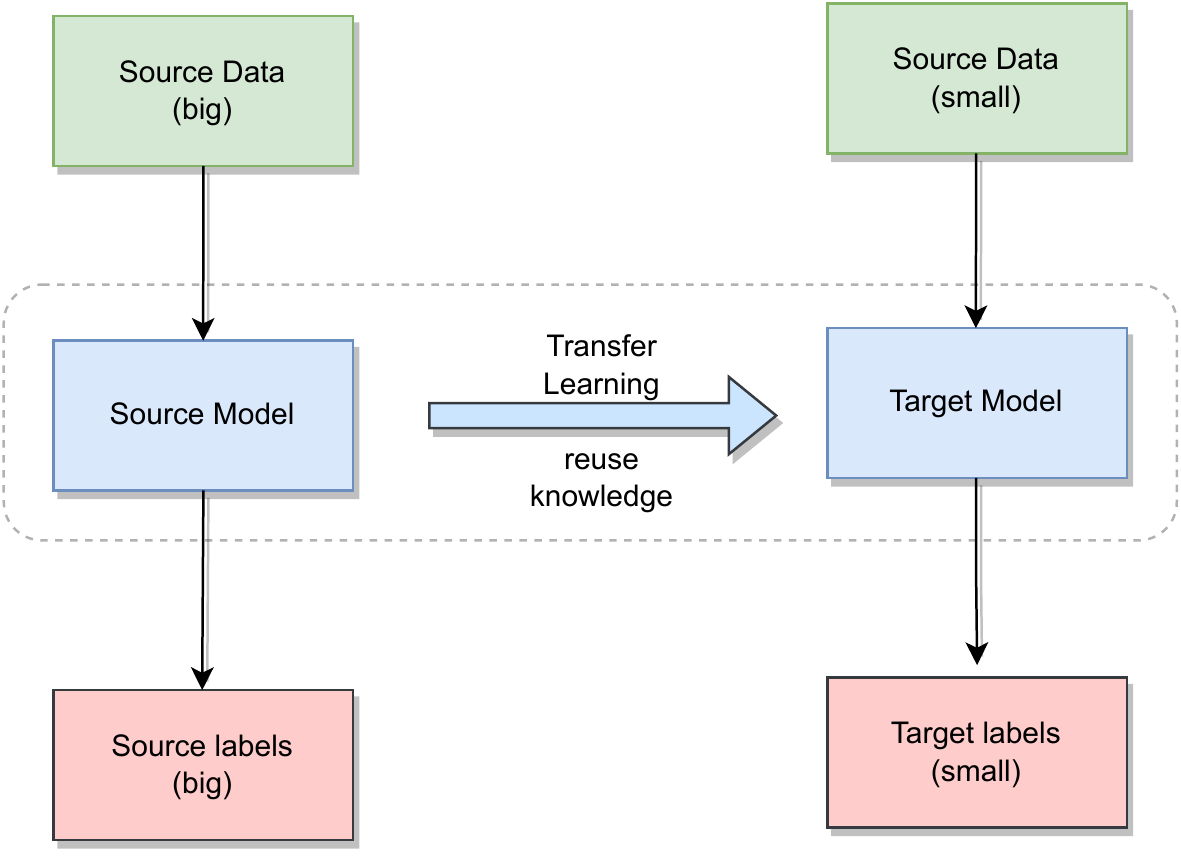}
\caption[Generic Transfer Learning Framework]{Generic Transfer Learning Framework. The big network is first trained on big datasets, and the knowledge learned by the big network after training is reused by a smaller target model which later learns a smaller dataset.}
\label{fig:tl}
\end{figure}

A high-level representation of the transfer learning framework from large input data to a smaller one is shown in Figure \ref{fig:tl}. Big networks are trained with a large data set to learn representation in the target task, which is then transferred and reused in smaller tasks with smaller networks. 

\begin{figure}[!htb]
\centering
\includegraphics[width=\linewidth]{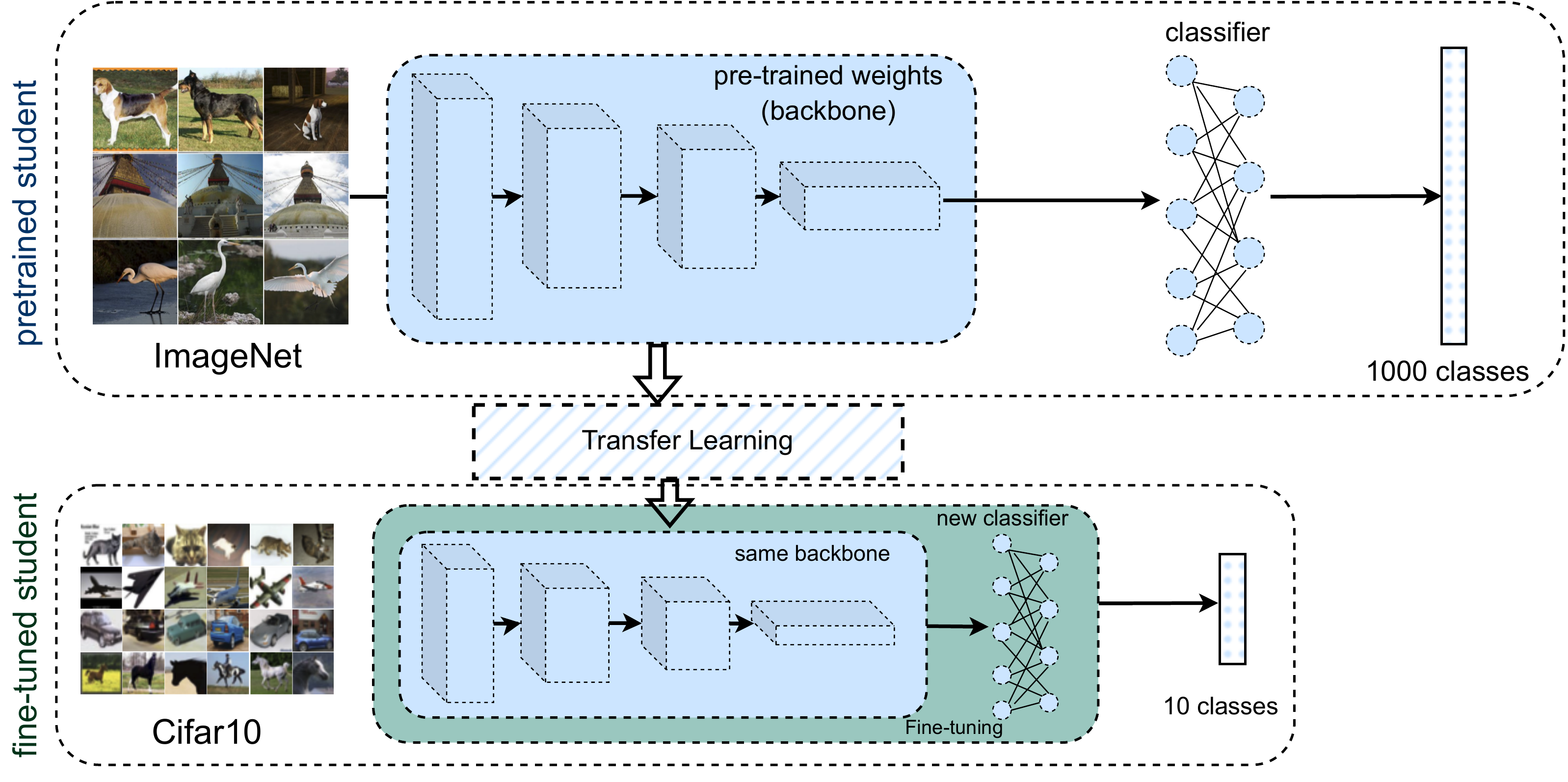}
\caption[Process of Transfer Learning from ImageNet to Cifar10]{Transfer Learning from ImageNet to CIFAR10. First, a pre-trained network with a backbone and classifier is trained with 1000 class ImageNet in the upper half. The backbone is reused with the same pre-trained weights, and a new classifier layer is appended to make a new student network. The new student network is fine-tuned with the new CIFAR10 data set to output its ten classes. Only the last layer(0.04\% of total parameters) is optimized from scratch during fine-tuning. Figure adapted from~\cite{Oquab_2014_CVPR}.}\label{fig:finetune_block}
\end{figure}

The source model is normally divided into two parts and is treated differently during transfer learning, as shown in Figure~\ref{fig:finetune_block}. Since the first layers of neural networks are more general than the final layers, the initial layers can be frozen to reuse feature extraction capabilities such as edge detection, line detection, etc. of neural networks~\cite{yosinski2014transferable}. The final layers that are more specific are generally replaced by a randomly initialized layer that learns the target-class mapping of the feature extractor during fine-tuning.

Generally, the transfer learning process can be summarized in these $4$ steps.\label{tf_process_step}
\begin{enumerate}[label=Step \arabic*:]
    \item Identify the source task, preferably in a similar domain. The availability of a massive dataset for this task is quintessential in this step.
    \item Train a model or obtain a pre-trained one for the source task.
    \item Reuse all or part of the source model as a starting point for the target model. Freeze the early feature extraction layers and replace the final layers of the target network with new classification layers that satisfy the number of classes for the target task.
    \item Fine-tune the whole network on a limited-available target dataset. 
\end{enumerate}

Figure \ref{fig:finetune_block} shows the actual process of transferring the representations from the CNN backbone along with data set examples, classifier changes, and the final output size. Details about these datasets and models are included in its own chapters ~\ref{chapter:dataset} and ~\ref{chapter:model}, respectively. 

Let us take an example of the transfer learning from a large image classification data set to another small image classification dataset. The computer science community has collected datasets such as ImageNet~\cite{imagenet} and CIFAR10~\cite{Krizhevsky2009LearningML} containing cropped images of different objects to perform the benchmark image classification task. As shown in Table~\ref{tab:dataset_details}, ImageNet is a large dataset with 128 thousand images per class (1.28 million total) sized 224*224 pixels. On the other hand, CIFAR10 is a comparatively smaller dataset with only five thousand images per class (only 50 thousand total) sized 32*32 pixels. The exact image samples from both datasets are also shown in Figure~\ref{fig:dataset_eg} as a reference. The primary motive here is to train a model to learn ImageNet image representations and reuse that representation while classifying the smaller CIFAR10 dataset. ImageNet is used here because it meets the criteria for the source data, as mentioned in Step 1 in Section~\ref{tf_process_step}. We know that earlier layers of networks learn general image characteristics such as edge detection, filters, etc. and these representations learned from a large dataset are reusable to be used in another similar task~\cite{pmlr-v27-bengio12a}. Since deep learning is data-hungry, such representations learned with large datasets with more samples are generally better than those learned from smaller datasets such as CIFAR~\cite{Oquab_2014_CVPR}. Reusing the feature extractor from the model trained with ImageNet would save the CIFAR10 model time and compute avoiding learning the same thing again target dataset is small. This helps to increase the efficiency of the data and results in faster training~\cite{Oquab_2014_CVPR} compared to training from scratch. Therefore, with transfer learning, we essentially get a better model faster with less computing compared to a CIFAR10 model trained from scratch~\cite{Oquab_2014_CVPR}. 

Let us formulate this task of fine-tuning the CIFAR10 dataset~\cite{Krizhevsky2009LearningML} into the steps mentioned above in Section~\ref{tf_process_step}. Again, the motivation for choosing datasets and the details of what these datasets are are explained above and in Chapter \ref{chapter:dataset}. For Step 2 the model with a large backbone and classifier is trained on ImageNet. Once trained, we reuse its backbone network and weight to make another model by appending a new classifier. We then fine-tune the newly generated model on a smaller CIFAR10 dataset where the weight for the backbone starts from the pre-trained weight, and the classifier is trained from scratch. More discussion of transfer learning along with its history is given in Section~\ref{related_work_tl}.
        
\subsection{Knowledge Distillation}
    \label{intro_kd}
\begin{figure}[!hbp]
\centering
\includegraphics[width=0.93\textwidth]{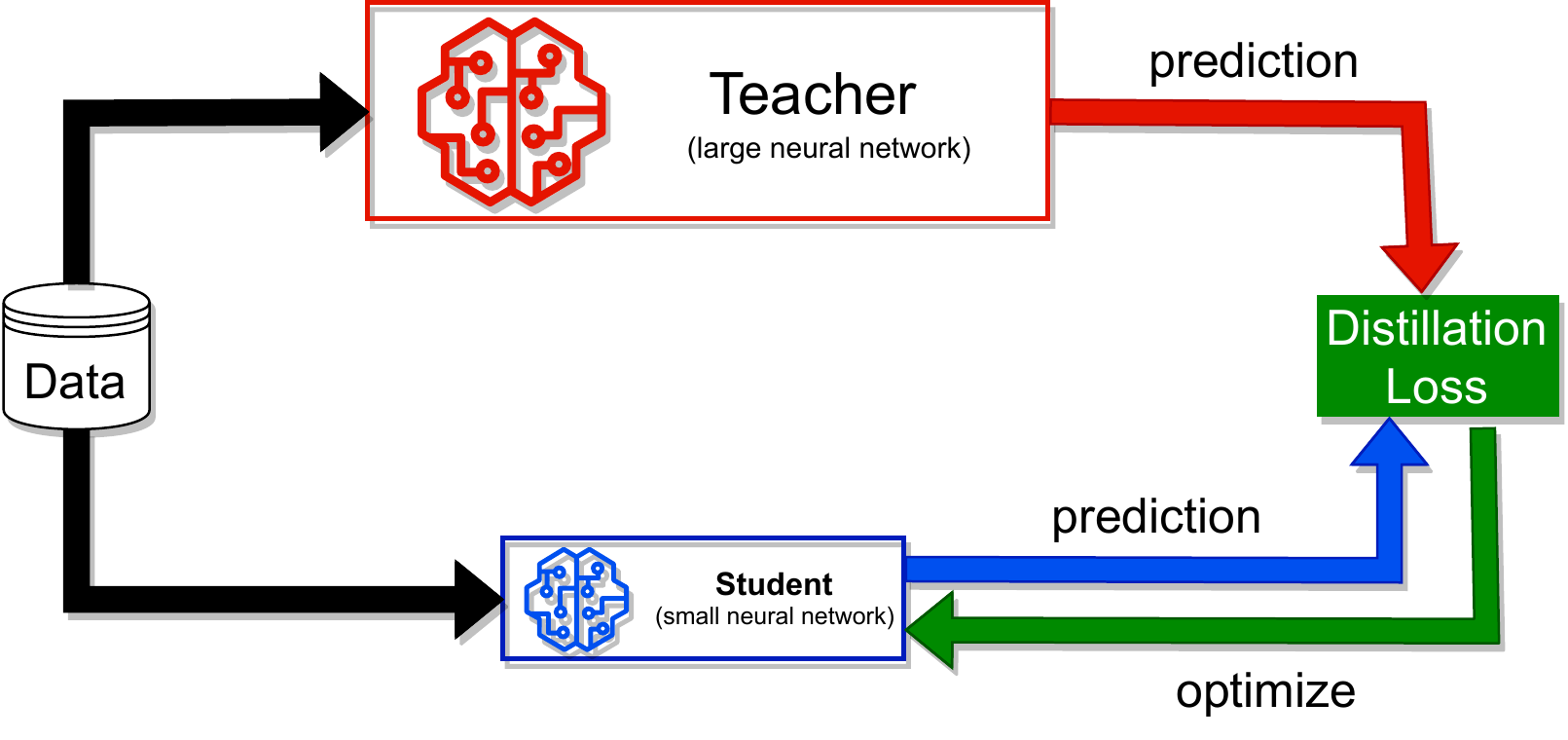}
\caption[Knowledge Distillation Architecture]{Knowledge Distillation Architecture. The teacher and student networks feed the same dataset to give predictions, which are compared with a loss function. One of the examples of this distillation loss function is explained later in Section~\ref{kldiv}. The student's output is compared with the teacher's output instead of the ground truth that generally occurs during transfer learning. The student network is then optimized and trained with the signal from the loss function. The teacher network is frozen and is not trained during this process. This figure is adapted from~\cite{kd_survey}.}\label{fig:KD_arch}
\end{figure}

Knowledge distillation is a way to transfer knowledge from large ensembles of strong models to smaller/faster networks. Knowledge distillation consists of two different networks, called the teacher network and the student network. Teacher networks are generally large ensembles of networks or a single neural network larger than student networks. The teacher network, generally pre-trained in vanilla form, takes in data and outputs logits, which the student network tries to mimic. Logits are the values we get as an output of the neural networks. Logits are generally passed through normalization layers such as Softmax. The softmax layer converts the real values to a distribution of probabilities for $K$ classes that add up to 1. The loss function compares the output of both teacher and student models and provides a loss value used to iterate and train the Student through Backpropagation~\cite{kelley1960gradientbackprop}. The architecture of vanilla knowledge distillation (or the teacher-student framework) is shown in Figure \ref{fig:KD_arch}. 

During training, a batch of images($x$) is passed through the network to produce a vector of raw outputs($z(x)$). If there are $K$ classes from which the model has to choose, the dimension of the raw output is also $K$. To give the probabilities for each class, the raw outputs are passed through normalization layers like Softmax. For an input $x$ and the raw output vector $z(x)= [z_1(x), z_2(x) ... z_K(x)]$ where $K$ is the number of classes, $i^{th}$ output from the Softmax function is:
\begin{equation}
    \hat{p_i}(x) = \frac{exp(z_i(x))}{\sum_j exp(z_j(x))}
\label{eq:softmax}
\end{equation}
where $j$ ranges from $0$ to $K$ and $i$ is an instance of $j$.

Furthermore, as proposed in a seminal paper~\cite{hinton_kd}, we can use the temperature scaling~\cite{hinton_kd} in Softmax to obtain softer probability predictions($\hat{p_i}(x:T)$) that result in a better distillation of knowledge compared to the hard labels of Equation~\ref{eq:softmax}.
\begin{equation}
    \hat{p_i}(x:T) = \frac{exp(z_i(x)/T)}{\sum_j exp(z_j(x)/T)}
\label{eq:temperature_kd}
\end{equation}
where $T$ refers to the temperature parameter. 

The temperature parameter $T$ in Equation~\ref{eq:temperature_kd} controls the softness of the output probabilities in the Softmax scores. $T=1$ is the special case that outputs vanilla softmax scores, i.e., hard labels from Equation~\ref{eq:softmax}. However, using values higher than $1$ produces a softer probability distribution between classes. For an intuitive understanding of the hardness and softness of logits, let us take an example of hard labels of a truck image sample with a score of $0.99$ for the truck class and $0$ or $1$ everywhere else. On the contrary, soft labels provide a more nuanced representation of classes with a probability distribution. The same truck image would have softer logits of $0.65$ for the truck, $0.25$ for the car, and a nominal score for other classes with higher temperatures. With this information, it is easier to learn that cars and trucks are more similar to each other compared to other classes. 

\begin{figure}[!htb]
\centering
\includegraphics[width=\textwidth]{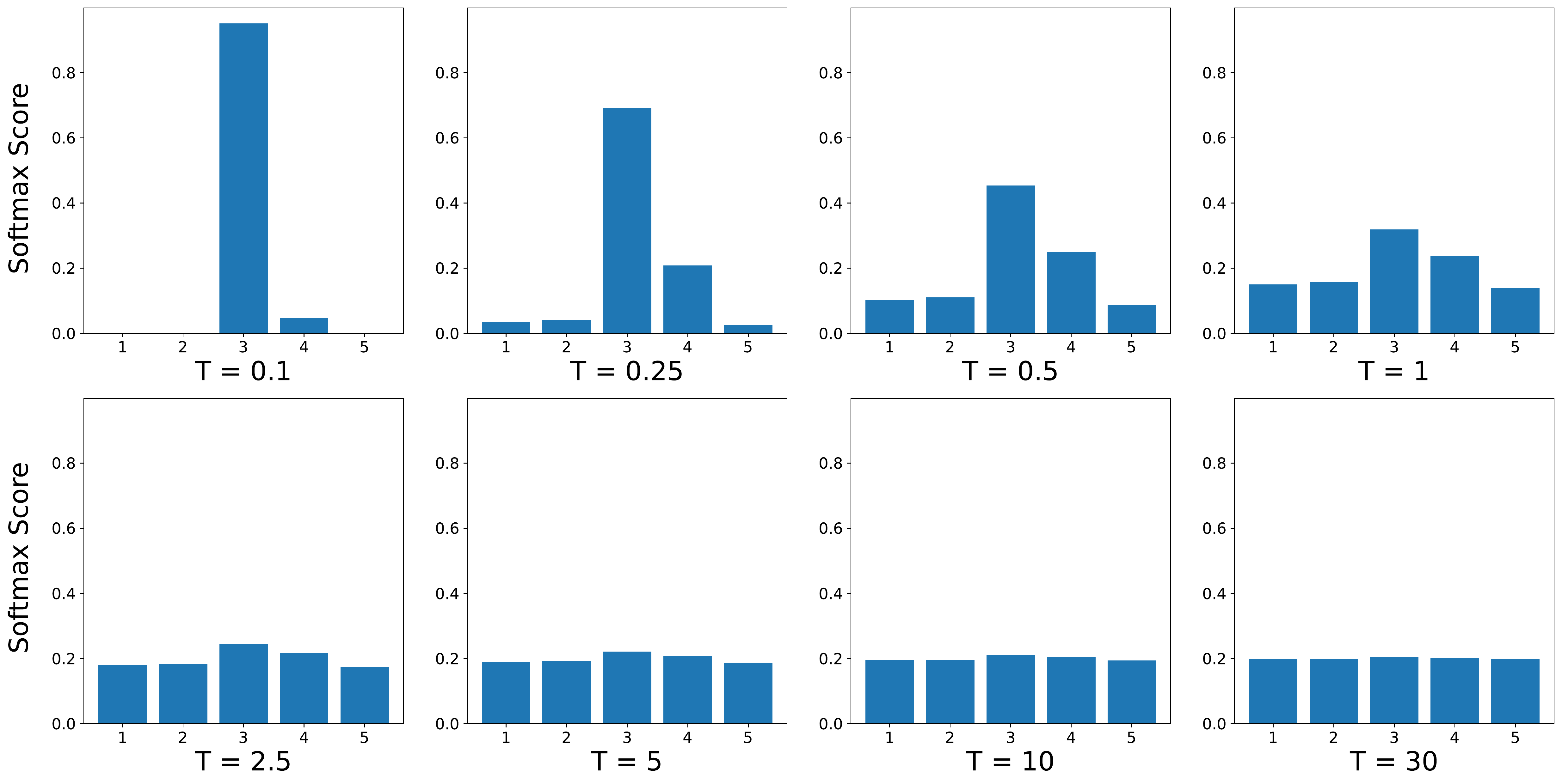}
\caption[Visualization of Softmax scores of a logit sample under different Temperatures]{Visualization of Softmax scores of a logit sample ([0.1, 0.14, 0.85, 0.55, 0.02]) at different softmax temperatures. Generally, we use vanilla softmax for $T=1$. Lowering $T$ makes the model more overconfident with harder labels, and increasing $T$ for knowledge distillation makes the logits softer. Note that the maximum score value is divided and decreases with higher temperatures.}\label{fig:logit_T}
\end{figure}

Further exploration of how the temperature value changes the output logits of the same network output is shown in Figure \ref{fig:logit_T}. Softmax with higher $T$ yields softer probabilities that are less confident in the model's prediction. With a lower value of $T$, i.e., more hard labels, the network tends to be more confident in its predictions. This happens because Softmax uses an exponential function, and the temperature value of Softmax penalizes larger logit values more than smaller logit values. 

\newpage
Vanilla form of knowledge distillation originally proposed in~\cite{hinton_kd} in classification problem learns with a combined total loss of hard and soft labels as:
\begin{equation}
    \mathcal{L}(x) = \alpha T^2 \mathcal{H}_{KL}(\sigma(z_s;t=T), \sigma(z_t;t=T)) + (1 - \alpha) \mathcal{H}_{CE}(y, \sigma(z_s;t=1)) 
\label{eq:kd_loss}
\end{equation}
Here, $z_s$ and $z_t$ are soft logits of the student and teacher network, respectively, with the temperature parameter $T$ in the Softmax function $\sigma$, $y$ is the ground truth, $\mathcal{H}$ is the loss function, and $\alpha$ is the hyperparameter for the weight of the distillation. $\alpha$ controls the weight and effect of distillation on the original loss. 

Similarly, $\mathcal{H}_{CE}$ refers to the standard cross-entropy loss, whereas $\mathcal{H}_{KL}$ refers to the \sm{Kullback–Leibler} divergence, also called KL divergence loss\label{kldiv}.
With KL Divergence loss($\mathcal{H}_{KL}$) we are trying to provide a measure of the relative distance between the probability distribution of soft softmax scores from the teacher and the student to help the student network mimic the teacher's soft labels. Cross-entropy loss is used to measure the discrepancy of prediction or softmax score with unit temperature and ground truth to help the student learn to match with ground truth. Student loss is a combined version of both losses and provides information on improving predictions with hard and soft labels.

\subsection{Data Augmentation}
Various augmentation techniques were used to introduce generalizability and robustness to the image recognition task. 
Augmentation is the process of general multiple versions of samples in datasets randomly applying various geometric, visual, context, and view-based transformations, as shown in figure \ref{fig:aug} below.

\begin{figure}[!htb]
\centering
\includegraphics[width=4.5in]{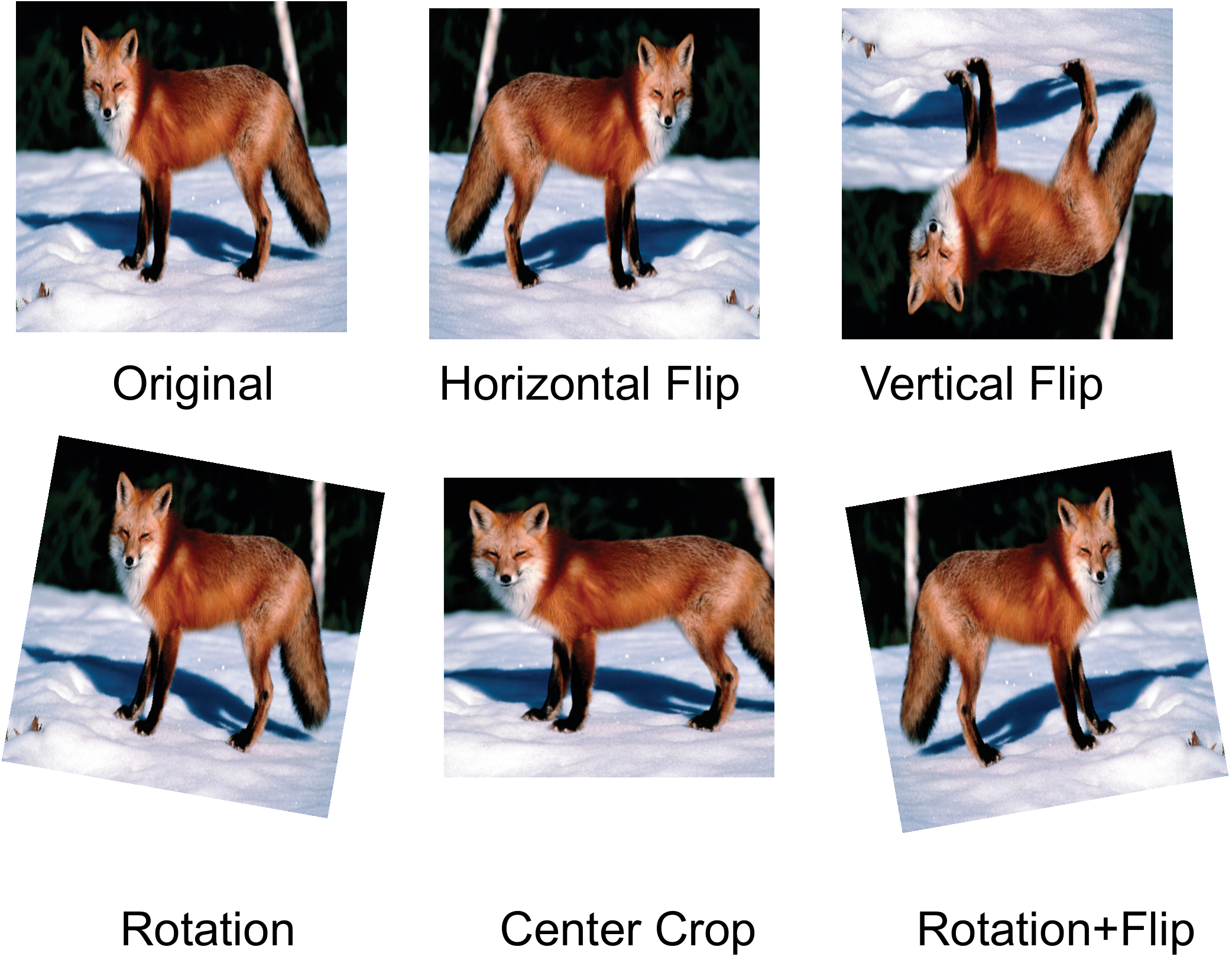}
\caption[Examples of Augmentations]{Examples of augmentations. The top left is the original image, and all other images are output after applying different types of augmentation. The images are only for illustration purposes. Not all augmentations are applied to all experiments. The sample is taken from ImageNet~\cite{imagenet} dataset.}\label{fig:aug}
\end{figure}

The datasets are randomly resized and cropped to obtain different views to augment the dataset. To avoid overfitting and introduce regularization, we use the dropout technique~\cite{JMLR:v15:srivastava14a} to disable the fraction of neurons randomly. This allows us to strengthen the representations learned by individual neurons. 
Two custom types of augmentation that we introduced were QuarterBlack and CenterBlack. For QuarterBlack, we blackened a random quarter or $\frac{1}{4}^{th}$ of the image with black pixels, i.e., value 0. CenterBlack\label{centerblack}, however, was more strict. We randomly blackened the central fraction of the image between a certain size and the full size of the image. The primary purpose of trying these harsh augmentations was because using simpler augmentation was not enough to add complexity to the network to degrade its performance by a significant margin. The primary purpose of blackening the center of the image with CenterBlack augmentation was to hide the foreground and only to make the background visible. This allows us to assess the effect of backgrounds on identifying and being confused with classes with similar backgrounds.   
Gaussian blur helps to add robustness against different types of blurring or noise prevalent in the dataset. 
Horizontal flip helps to double the size of the data set by introducing different views of the data set. Most of the time, objects are horizontally symmetric or maintain the class type even after flipping the image horizontally. In most cases, VerticalFlip does not provide a naturally occurring view, so we avoid applying VerticalFlip. 
Normalization is needed to speed up the optimization process by converting the pixel values into the range of $[0,1]$. Three different values of mean($\mu$) and standard deviation($\sigma$) are used normally to apply standard deviation normalization. The formula for this normalization is given below. 

\begin{equation}
    x = \frac{x - \mu}{\sigma}
\end{equation}
where $\mu$ and $\sigma$ are a list of the mean and standard deviation of the three channels of the dataset. 

\subsection{Interpretability}
    ~\label{intro_interpret}Apart from the quantitative results such as accuracy scores, it is also crucial to understand why a model makes a certain prediction. However, most of the time, the models providing the highest scores, like Deep Neural Networks, are inherently black-box, struggle to provide any interpretation, and force a trade-off between accuracy and interpretability. With interpretability, humans can understand the cause of prediction, also known as determinants of classification, and provide explanations. Interpretability and explainability terms are usually used interchangeably, but even if they are closely related concepts, some articles~\cite{doshivelez2017rigorous, 8466590} suggest otherwise. Interpretability generally focuses on the intuition behind the model's output, while explainability focuses on explaining the internal logic of the model itself. For example, in the image classification task, interpretation of the model prediction could show a certain relevant area of the image that the model focuses on classifying. However, this does not necessarily mean that one can explain the internal logic of the model that led to that decision. Therefore, the interpretability of the ML system does not necessarily mean the explainability or vice versa~\cite{survey_xai}. 

Various methods have been developed in response that range from model-based interpretable methods like Linear regression, Logistic Regression, Decision Trees, TabNet~\cite{arik2019tabnet} etc.  to model-agnostic methods~\cite{ribeiro2016should, Ribeiro2018AnchorsHM, lundberg2017unified}. Global interpretability~\cite{pdp,arik2019tabnet} explains the inherent nature of the network as a whole. On the contrary, the local interpretability~\cite{ribeiro2016should} focuses more on the sample level and attempts to explain the network predictions at the individual level. If the network is too complex to make it inherently explainable, Post-hoc interpretability techniques can be used after training the model. Such techniques attempt to quantify the contribution of each input feature to the prediction after it is trained and can be divided into perturbation-based, gradient-based, or attention-based techniques based on how they generate explanations. For example, LIME~\cite{ribeiro2016should} is called a perturbation-based method because it applies perturbations to individual instances to generate local interpretable approximations with linear models. Gradient-based methods, on the other hand, compute gradients of outputs concerning input for a particular class, which provides information regarding how the change in input changes the output prediction of a particular model. Among gradient-based techniques, Deep Learning Important FeaTures, also called \\DeepLIFT~\cite{shrikumar2017learning} backpropagates the contribution of all neurons to every feature of the input. It observes the activation of each neuron and compares it with some reference activation to provide positive and negative contributions based on that difference. On the other hand, the attention-based technique relies on the weight the models like Transformers (explained in Chapter~\ref{chapter:model}) assign to their set of input tokens. 

\subsection{Shapely Values}
~\label{intro_shap}
Shapley value was introduced in cooperative game theory~\cite{P-295} as a way of providing a fair solution to find the contribution of each member who collaborates with a team or coalition to provide a particular value. \sm{As shown in Figure~\ref{fig:shapely_value} below, let's assume a team of 3 players, i.e., Red $A$, Green $B$, and Blue $C$, are playing a game together and win some money as a prize.} Each of their contributions can be identified \sm{by analyzing} the prizes they get when they play individually, as a couple, and together as a single coalition. \sm{Order in which players are selected} is an important factor when making sub-teams since player $A$ might perform better when he plays with $B$. Colored players
within the box in Figure~\ref{fig:shapely_value} represent a coalition for a game that yields prizes on their right. For example, in the top-right box, Player $A$ wins \$5, whereas, in the bottom-left box, Players $A$ and $B$ together win \$15. Once the prizes for every single possible coalition of players are determined, the contribution of adding another player is calculated as specified in Figure~\ref{fig:shapely_value_permutation}. 

\begin{align*}
\text{Shapley value}(A) &= \frac{5+5+8+8+8+7}{6} = 6.83\\
\text{Shapley value}(B) &= \frac{10+11+7+7+10+11}{6} = 9.33 \\
\text{Shapley value}(C) &= \frac{6+5+6+6+3+3}{6} = 4.83
\end{align*}

\sm{In machine learning, features of the dataset are treated as players, prediction of the model as the payout, and a subset of features is treated as the coalition of players. For example, to interpret a model that outputs the price of a house, features such as number of rooms, location, size of yard, number of floors, color, etc. could be players. After analyzing the different subsets of features to predict house price, we will likely get the result that the contribution of the number of rooms would be higher than the contribution of color for the price of the house.}

\begin{figure}[!htb]
\centering
\includegraphics[width=4.5in]{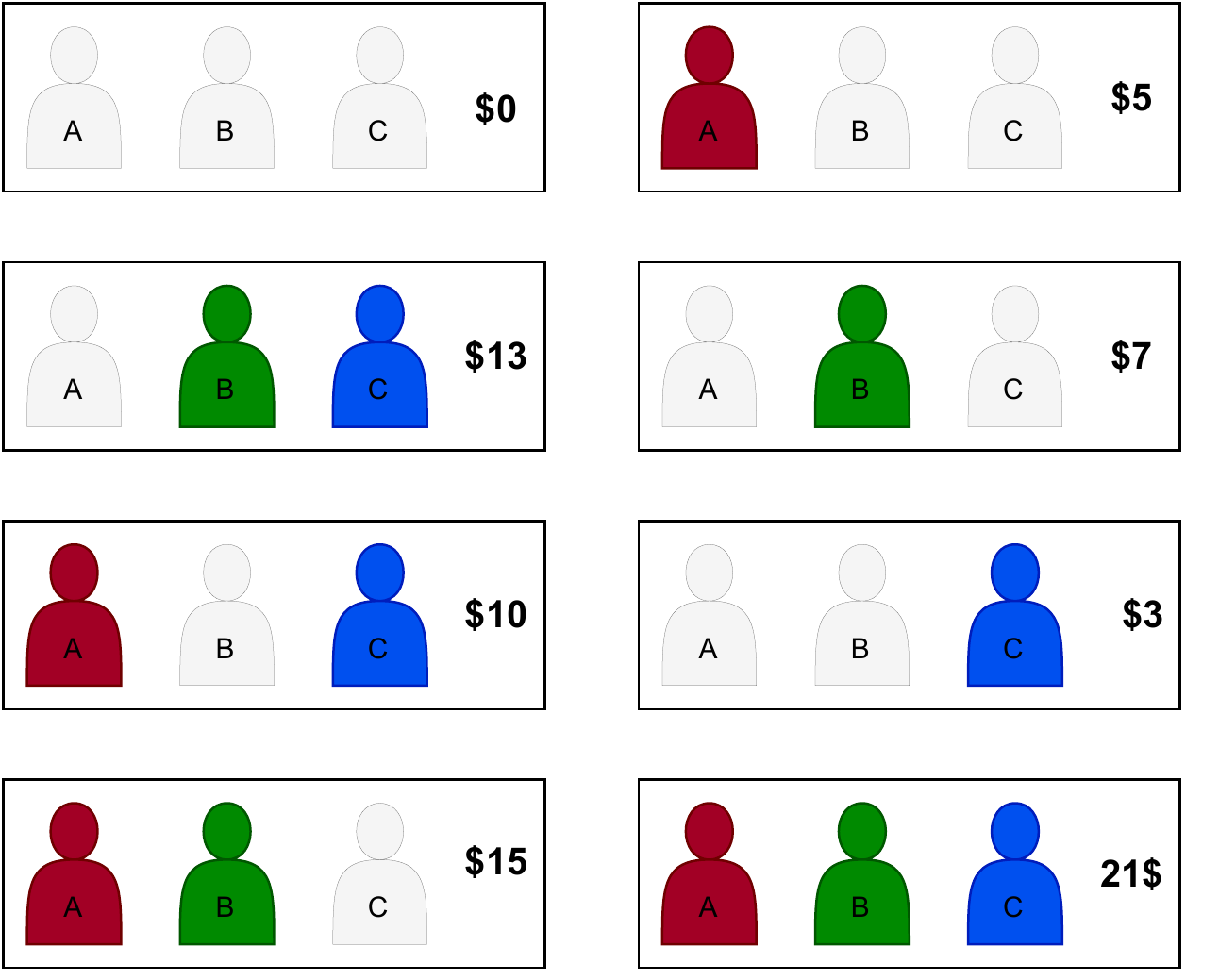}
\caption[3-player coalition example for Shapely values]{Example of a 3-player coalition for Shapely values. Colored players within a box represent a coalition for a game that yields prizes on their right. White players represent their absence. Using the prize for these coalitions, we calculate the individual contribution of each player in Figure~\ref{fig:shapely_value_permutation}. Example is adapted from~\cite{clearcode_2020_shapley}.}
\label{fig:shapely_value}
\end{figure}

\begin{figure}[!htb]
\centering
\includegraphics[width=4.5in]{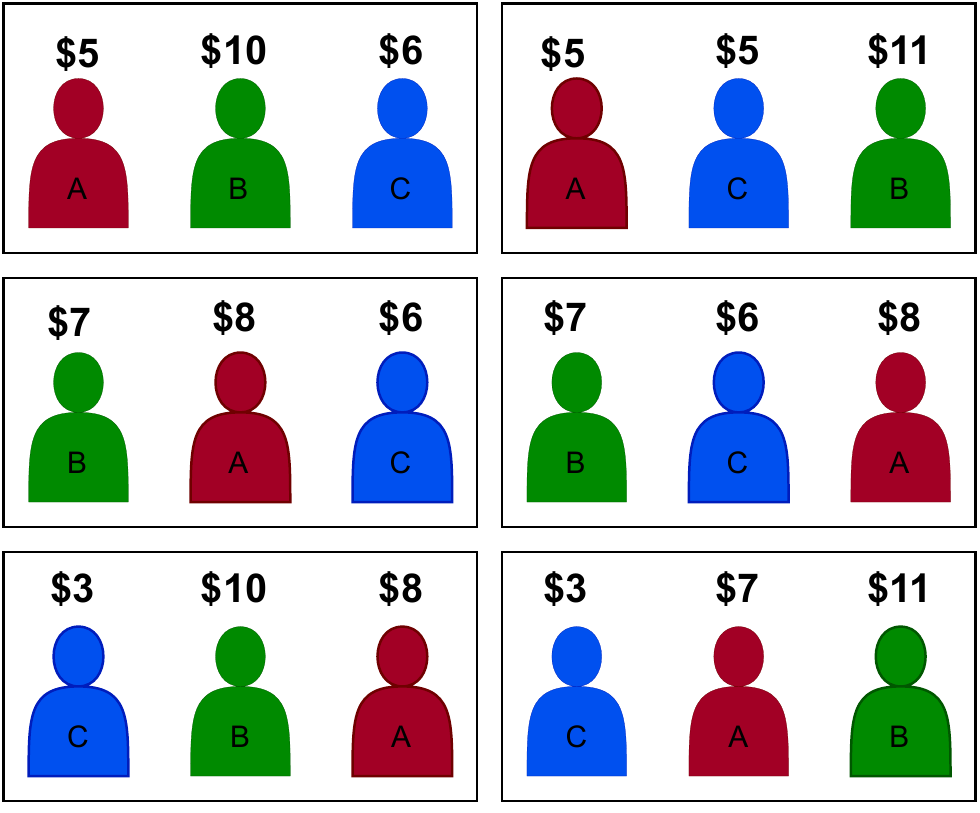}
\caption[Individual contributions of players for each permutation of players for Shapely values]{Individual contributions of players for each permutation of players is shown above all players. Taking the top-left box as an example, first Player $A$ wins \$5 when it is alone in Figure~\ref{fig:shapely_value}. Adding Player $B$ to previous $A$ adds another \$10 prize since they get a total of \$15 prize when only $A$ and $B$ are together above. Adding $C$ to the previous $A$ and $B$ adds \$6 and makes the total prize as \$21. All the permutation in all of these boxes here adds up to the same total of \$21. Average individual contribution is calculated by taking an average of each player in all these permutation settings. Shapely values for Players $A$, $B$, and $C$ in this example are calculated to be 6.83, 9.33, and 4.83, respectively.}
\label{fig:shapely_value_permutation}
\end{figure}

To interpret the vision classifier models, we can see an example in Figure \ref{fig:shapely_img_eg} below. Instead of players, coalitions A, B, and C would now be a region of Superpixels where Figure \ref{fig:shapely_img_eg} shows an example of Superpixel $A$ as the purple region in the fox. Superpixels are perceptually grouped regions, generally generated from image over-segmentation or clustering. Generating Superpixel patches instead of working directly on pixels would decrease the number of input features for SHAP. In general cases, instead of just three, there can be hundreds of Superpixels in an image that contribute to a certain prediction. 


\begin{figure}[!htbp]
\centering
\begin{subfigure}{.49\textwidth}
  \centering
  \includegraphics[width=\linewidth]{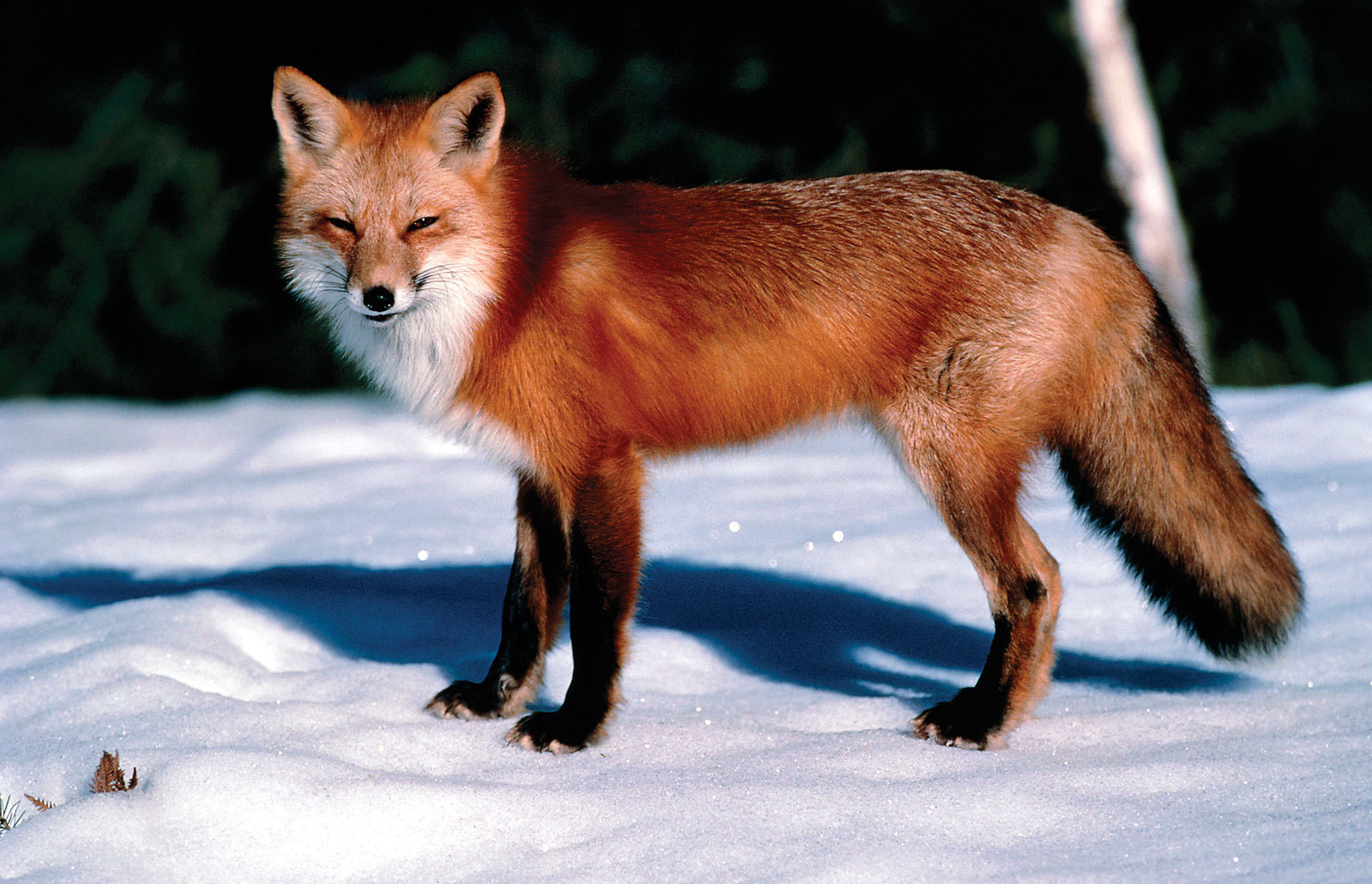}
  \caption{Sample Test Image}
  \label{fig:shap_eg_fox}
\end{subfigure}%
\begin{subfigure}{.49\textwidth}
  \centering
  \includegraphics[width=\linewidth]{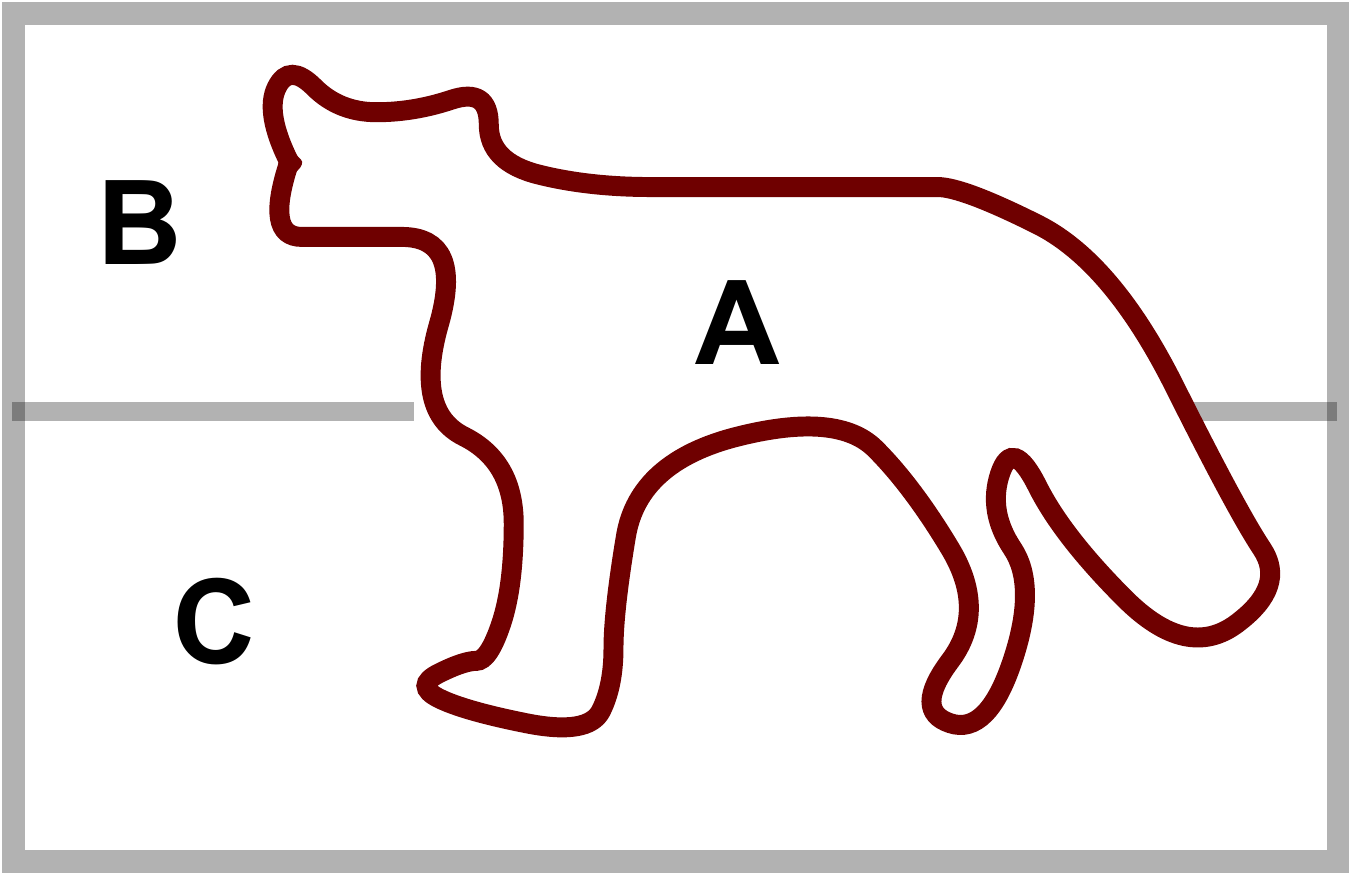}
  \caption{Oversimplified Superpixels}
  \label{fig:shap_eg_sp}
\end{subfigure}
\caption[3-player simplified coalition for image for Shapely values]{Representation of a simplified coalition of 3 players for an image sample. Assuming that we find 3 primary Superpixels A, B, and C with over-segmentation, we can treat these Superpixels as players as in Figure~\ref{fig:shapely_value} that yield \sm{different relevance scores. The full process of getting Superpixel contributions is described at the end of Section~\ref{shap_lift_process}}. This representation is only for illustrative purposes, and there could be hundreds of such Superpixels in reality.}
\label{fig:shapely_img_eg}
\end{figure}

In this way, the contribution of each member is calculated by adding and removing a particular member from all subsets of the remaining members. It is particularly tricky because the interaction of one member with the different permutations of the team members could be \sm{different in exponential ways because of the permutation}. Hence, computing Shapely values in the original form is NP-hard and \sm{has exponential time complexity}, and it is infeasible to be used in practice. To circumvent this, \sm{various approximation techniques for different types of models have been proposed in~\cite{lundberg2017unified} where the authors show that Shapley values satisfy different desired properties of explanation of any model.} To obtain the interpretations of our deep neural network models, we use Deep SHAP (also called `DeepExplainer') from the SHAP~\cite{lundberg2017unified} library, which is the combination of DeepLIFT~\cite{shrikumar2017learning} and Shapley values. Deep SHAP uses DeepLIFT's node-based attribution rules to approximate conditional expectations of Shapely values using a selection of background samples. When a large number of background samples is combined, differences between the expected values of the actual output we get with the input are obtained and contrasted with that of background samples.

\label{shap_lift_process}
\sm{To understand the process of getting the Deep SHAP interpretation for images with our deep neural networks, let's take the example from Figure~\ref{fig:shapely_img_eg} and its Superpixel representations. First, we get the reference prediction for the background $(\hat{y}_{background}$) by passing a black image or a random sample of training images through the neural network. Now we build different coalition of superpixels like in Figure~\ref{fig:shapely_value} to obtain the prediction for each subset as input $(\hat{y}_{actual})$. The score for this particular coalition is obtained by obtaining the difference between the actual prediction and the reference prediction, i.e., $(\hat{y}_{actual} - \mathbb{E}[\hat{y}_{background}])$ which also corresponds to the difference in the input pixels. If the difference is positive, the subset of superpixels in the coalition is considered to provide a positive contribution, as shown in red in Figures~\ref{fig:shap_all_3}-\ref{fig:shap_all_one} whereas negative values are considered negative contributions, as shown in blue superpixels. Adding all the positive \& negative SHAP contributions together with class expected values gives the model's output for that particular class. 
}
    
\section{Motivation and Proposal}
Over the years, there have been many attempts to develop novel techniques to apply and improve the performance of transfer learning and knowledge distillation separately, as shown in Chapter~\ref{related_work}. However, to our knowledge, \sm{this is the first time Knowledge Distillation is applied and explored the effects in a transfer learning setting.} Doing this would help us tackle the problem of not having a massive dataset with transfer learning and/or the problem of not having massive computing to train large models every time we need to build a well-performing model. Our main hypothesis is that knowledge distillation could be used to improve further the validation performance of a network in terms of evaluation metrics like accuracy after fine-tuning. Along with improved performance, we also dive into how knowledge distillation affects transfer learning in detail. With transfer learning, we can reuse the representation of a model trained on a related dataset and improve performance while training on a limited dataset. However, with knowledge distillation, we could again improve the actual prediction capacity of the network by training it to mimic a larger network. This means we are getting the best of both worlds, i.e., big model-like performance with the limited in-domain dataset. This will greatly reduce the dependence of the machine learning production environment and embedded systems on a large dataset and/or a large network architecture. In this thesis, we explore the effects of knowledge distillation on transfer learning for image classification both qualitatively and quantitatively and present our findings. 

We go over the background in Section~\ref{sec_background} and the previous literature in detail and discuss where our contribution fits in Chapter~\ref{related_work}.

We summarize our research and our findings in five different research questions involving TL+KD. A community standard image classification problem is used to test and validate all these hypotheses due to its popularity and seminal contribution to the transfer learning literature~\cite{Oquab_2014_CVPR, yosinski2014transferable}. Our list of hypotheses, along with the process of validating them, is described below.
\begin{itemize} \label{hypothesis_list}
    \item Hypothesis 1: \FIX{TL+KD} architecture improves image classification over the TL architecture on the CIFAR10 dataset.
    \begin{description}
        \item[\hspace*{19pt}] The validation performance of a \FIX{TL+KD network} should be more than TL only model in the target task. We plan to investigate this hypothesis by evaluating the validation performance of the student model on the same task, i.e., image classification \FIXME{of CIFAR-10 dataset}, first with vanilla transfer learning and then with transfer learning with knowledge distillation. We plan to evaluate both models on multiple metrics such as accuracy, precision, recall, and F1 score and look at the confusion matrix, as well as SHAP interpretability. Each of these evaluation methods should infer that the performance of TL + KD is better than that of TL-only training to validate this hypothesis successfully. 
    \end{description}
    \item Hypothesis 2: \sm{TL+KD} \FIX{provides more contribution on foreground image features measured by SHAP contribution} than TL only.
        \begin{description}
            \item[\hspace*{19pt}] TL+KD network's focus should increase in the foreground of the image for the CIFAR10 classifier. We plan to investigate this hypothesis by differentiating the foreground and background of the images and evaluating the SHAP interpretability scores of the models on each of them separately with and without KD. To successfully validate this hypothesis, the positive SHAP contribution given by TL+KD in the foreground should be greater than that of TL alone. 
        \end{description}
    \item Hypothesis 3: TL+KD achieves similar validation accuracy faster with fewer training epochs than TL only model. 
        \begin{description}
            \item[\hspace*{19pt}] To investigate this hypothesis, we look at the convergence of the same student image classifier with and without the KD and assess the epochs taken to achieve similar accuracy thresholds. TL-only model reaching an accuracy threshold slower with more epochs than the TL+KD setup would validate this hypothesis.
        \end{description}
    \item Hypothesis 4: \sm{TL+KD} improves validation accuracy even after adding training complexities such as training with a fraction of data or corrupted images.
        \begin{description}
            \item[\hspace*{19pt}] For this, we investigate the same student image classifier model with additional complexities while training. Additional complexities here mean providing more difficult data to the model to drive the validation accuracy downward so that we can see the effects of TL+KD for lower-accuracy spaces, too. More difficult data may refer to taking only a fraction of training samples or randomizing a fraction of training labels. Training both models would be evaluated based on the validation accuracy that students can achieve even with complexity. Differences in their validation accuracy or lack thereof with complexities would help validate or invalidate this hypothesis.
        \end{description}
\end{itemize}
We evaluated these hypotheses to explore the TL+KD model, validate claims, and present our results based on five corresponding research questions in the Results and Discussion Chapter~\ref{chap:results}.

\chapter{Related Work and Contribution}
\label{related_work}\label{chap:related_work}
\section{Image Classification}
Computer vision, similar to human vision, enables computers to see, recognize, and process visual elements of an image or video. Image classification or image recognition is one of the core requirements of computer vision models and is widely used in real-world applications like self-driving cars, facial recognition, medical imaging, tracking, etc. For this reason, there has been a great deal of fascination with solving vision-related tasks in the AI community. Previously, the image used to be transformed into simpler features~\cite{hog, lowe1999objectsift} and applied machine learning algorithms like K nearest neighbour(KNN)~\cite{knn}, Support Vector Machines(SVM)~\cite{svm}, Perceptron~\cite{haykin1994neural} etc. However, with massive data and more powerful computing, CNN~\cite{lecun1989, lecun-mnisthandwrittendigit-2010} based deeper neural networks are currently dominating the image classification space. In the ImageNet LSVRC~\cite{imagenet} benchmark, deep and wide CNNs~\cite{krizhevsky2012, vgg, he2016deepresnet, efficientnet} have been dominating the leaderboard for almost a decade. Recently, transformer-based deep neural networks~\cite{bit, dosovitskiy2020imagevit, wu2021cvt} have also achieved impressive benchmarks, often exceeding previous benchmarks. Transformers are generally known to improve out-of-distribution robustness~\cite{hendrycks2020pretrained} and vision transformers specifically are also more robust to corruptions, perturbations, and distribution shifts~\cite{paul2021vision}. The use of extra data on convolutional-based image classification methods is also shown to build more robust~\cite{hendrycks2018deep, abstention_ood, abstention_ood_or, ilyas2019adversarial} classifiers. For this reason, in our experiments, we explore two types of network architecture to explore the effects of TL+KD, a deep convolutional neural network, and a vision transformer. 

\section{Transfer Learning}\label{related_work_tl}
Transfer learning has decades of history long before deep learning was popular~\cite{tl_stevo, NIPS1994_0f840be9}. Transfer of learning as a paradigm has an even longer history in psychology, cognitive science, etc. In machine learning, transfer learning helps solve a problem with a limited dataset and improves performance with the help of other related problems and datasets. Because of its significance, Transfer Learning has been applied to various machine learning algorithms such as text classification, heterogeneous image classification, medical imaging, etc. For image classification,~\cite{Oquab_2014_CVPR} trained an ImageNet model to obtain a feature representation and reuse the initial layers. for another task that had fewer data. Parallel research has been conducted~\cite{10.5555/3016100.3016236} on domain adaptation and transfer learning when the source domain differs from the target domain in conjunction with transfer learning. Some research directions also focus on learning representations without supervision and reusing learned representations~\cite{pmlr-v27-bengio12a}. The initial layers are known to be generalists and learn task-agnostic filters or color blobs, whereas the higher layers are more problem-specific and feed on lower-level features. Generalist neurons are known to be most helpful for reuse for transfer learning, but it is not clear where or how the network chooses the specificity of the network. Some attempts have been made to understand transfer learning and explore what is being transferred.~\cite{yosinski2014transferable} quantifies each layer in terms of generality and specificity and reports on cases where transfer learning is good and bad. Similarly,~\cite{neyshabur2020being} finds that low-level statistics are primarily responsible for transfer learning and pre-training guides the target model towards a flat basin of energy loss landscape. The target model achieved similar accuracy even from earlier checkpoints of the pre-trained model and was generally better than training with randomly initialized weights~\cite{neyshabur2020being, yosinski2014transferable}. 

\section{Knowledge Distillation}\label{related_work_kd}
Ensembles of networks are usually trained with different architectures or initializations to minimize similarities in their learning and achieve maximum performance. However, running the ensembles in the production environment is extremely time-consuming. In an attempt to combat this, the idea of training a small production model by transferring knowledge from larger networks on the same training data has also been around for more than a decade~\cite{caruana_compression}. The primary purpose was to teach data mapping to the set of network labels or learning of big networks. This knowledge transfer with the help of soft labels was first introduced in~\cite{hinton_kd}. The basics of what it is and how it works are described in detail in Section~\ref{intro_kd}. Various knowledge distillation methods have been proposed in the literature since their inception. Typically, the progress of the type of distillation can be divided into three types of distillation, i.e., offline, online, and self-distillation~\cite{kd_survey}. In offline KD, the teacher network is typically pre-trained to an extent and then distilled into a student network separately~\cite{hinton_kd}.  In online distillation, teacher and student networks are simultaneously trained end-to-end and improved~\cite{mirzadeh2020improved, chen2020online}.
Self-distillation is a form of self-supervised learning in which the same network of previous time steps provides the knowledge to distill a newer version of the student~\cite{zhang2019your}. Sometimes self-distillation can also be used to distill knowledge from specialized deep layers to generalized shallow layers. On the other hand, actual knowledge can also be divided into three types, i.e., response-based, feature-based, and relation-based. In response-based knowledge, we try to mimic the teacher's predictions with the student~\cite{hinton_kd}. Feature-based knowledge uses features or parts of the teacher network to mimic representations~\cite{aguilar2020knowledge}. In relation-based knowledge distillation, the relationship between the different layers or data is explored, treating correlations between features as knowledge~\cite{gift_from_kd}.~\cite{Kim_2021_ICCV} explores Knowledge Distillation with Self-supervised learning or self-training for self-distillation without the need for explicit larger teacher networks. For a more comprehensive survey of KD methods, see~\cite{kd_survey}.

In summary, there are many ways in which knowledge can be distilled for a student. However, despite many variations and great successes in practice, there has been limited research attempts~\cite{phuong2019towards, efficacykd, cheng2020explaining, ji2020knowledge, stanton2021does} towards a theoretical or empirical understanding of knowledge distillation.
~\cite{phuong2019towards} exhibits theoretical characterization of the solution learned by the student by deriving bounds on the transfer risks.~\cite{ji2020knowledge} provides another theoretical perspective on KD and explores ways to achieve a perfect vs. imperfect teacher.~\cite{efficacykd} finds the dependence of student and teacher architecture and the advantage of early stopping to bridge the gap between the performance of the models.~\cite{cheng2020explaining} explores methods to analyze visual concepts with an intermediate neural net layer based on the relevance of the task. He also finds that KD yields more stable optimization directions. It also explores the changes in KD learning for different epochs. Finally,~\cite{stanton2021does} explore the discrepancies in student generalization with fidelity to match teachers and explore ways to improve the fidelity. However, none of them explores the effect of KD on transfer learning, nor does it explore the cases of training that we explore, like convergence, the dependence of complexity, etc. To our knowledge, this is the first instance of exploring the effects of knowledge distillation in the transfer learning scheme and quantifying qualitative results with SHAP values.


\section{Contribution}
The summary of our contributions, along with their significance, includes the following.
\begin{itemize}
    \item Our main contribution is applying knowledge distillation in transfer learning setup for image classification on the CIFAR10 dataset. We report results on various evaluation metrics beyond accuracy with interpretations. We see quantitative and qualitative gains in TL+KD supported by multiple runs. To our knowledge this is the first time, to our knowledge, that someone is applying knowledge distillation on transfer learning in any domain.
    \item We not only provide the explanations but also analyze how the TL+KD affects the model's features and quantify model interpretations based on foreground and background. We find that with TL+KD, the positive contributions the model provides in the foreground are always stronger. However, the positive contribution of the model in the background is inconsistent and varies depending on the instance of data being investigated.  To our knowledge, this style of quantifying the SHAP interpretations based on semantic segmentation of background and foreground is also done for the first time in the literature.
    \item We explore the effects of TL+KD on various training scenarios like using a fraction of training data and adding corruptions in images and labels for image classification tasks and provide empirical evidence for enhanced TL+KD performance. We find that training is initially adversely affected in terms of validation accuracy by adding KD but is quickly overcome in subsequent epochs. We also explore how TL+KD behaves with different added complexities of the training. There is a little work~\cite{efficacykd, stanton2021does} exploring the effect of TL+KD in certain other scenarios, but not in the complexities we explored. We find that, under the complex scenarios that we used, the gains with TL+KD are still seen based on validation performance.
\end{itemize}

\chapter{Baseline Dataset}\label{chapter:dataset}
In this work, we use two datasets called CIFAR10~\cite{Krizhevsky2009LearningML} and ImageNet~\cite{imagenet} to test our hypotheses. Both are in a similar domain and have images of nouns collected from the web. A detailed description of its origin and content is explained below.

\section{CIFAR10}
\begin{figure}[!htbp]
\centering
\begin{subfigure}{.6\textwidth}
  \centering
  \includegraphics[width=\linewidth]{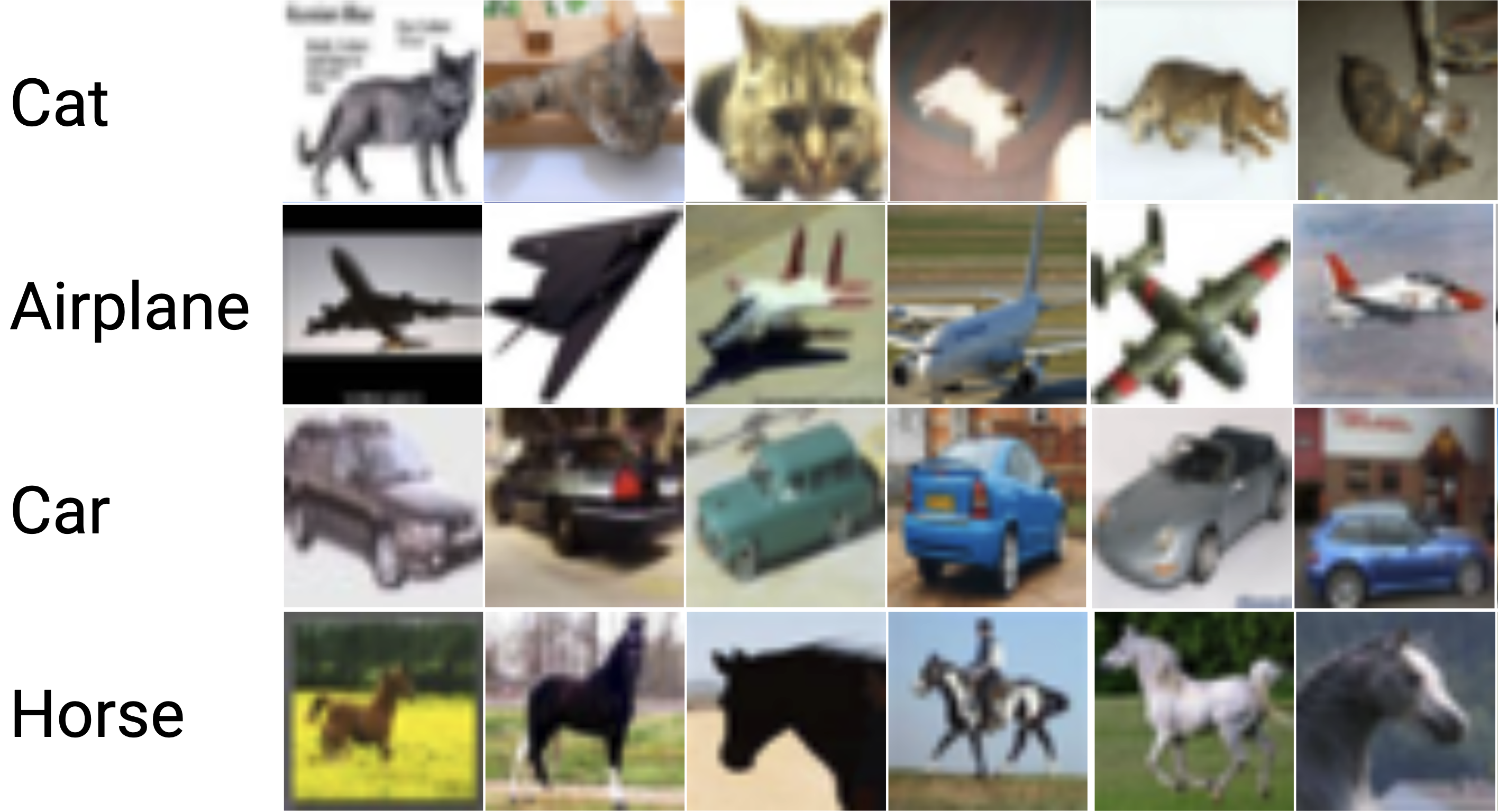}
  \caption{CIFAR10}
  \label{fig:dataset_eg_cifar}
\end{subfigure}%

\bigskip

\begin{subfigure}{.9\textwidth}
  \centering
  \includegraphics[width=\linewidth]{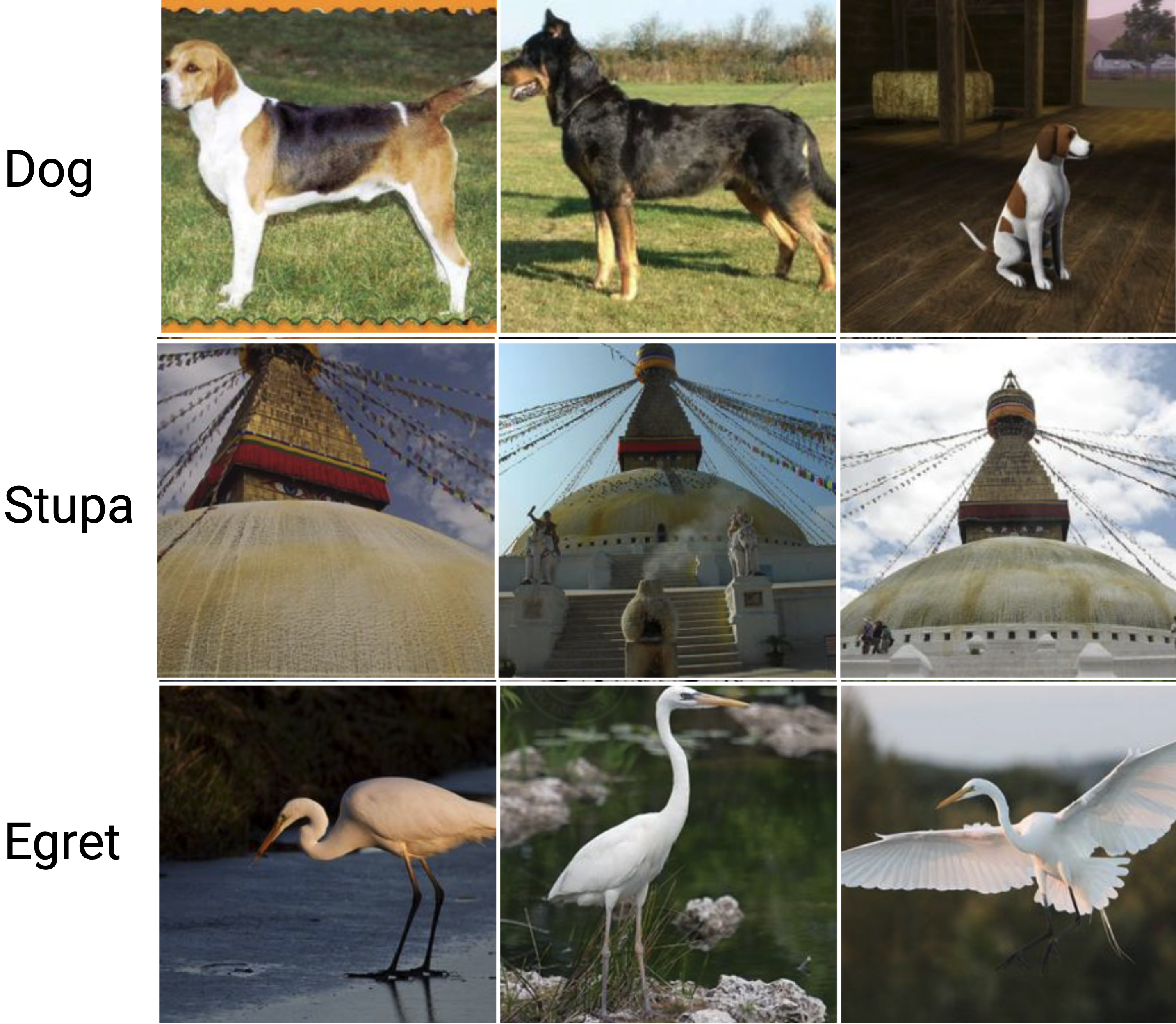}
  \caption{ImageNet}
  \label{fig:dataset_eg_imagenet}
\end{subfigure}

\caption[Examples of a few classes of Datasets]{Examples of a few classes of each dataset a) 4 out of 10 CIFAR10~\cite{Krizhevsky2009LearningML} classes and b) 3 out of 1000 ImageNet~\cite{imagenet} classes. ImageNet is significantly larger and has more resolution than CIFAR10.}\label{fig:dataset_eg}
\end{figure}

For fine-tuning, we use the CIFAR object recognition dataset introduced in 2009 through a technical report~\cite{Krizhevsky2009LearningML} and further summarized on the official website~\cite{cifar_web} for the target task. The CIFAR10 data set is the subset obtained from another data set called TinyImages~\cite{tinyimages}. For the TinyImages dataset, 80 million images were collected by searching 79,000 query terms extracted from all non-abstract nouns from an English lexical database called WordNet~\cite{wordnet}. Of these, the CIFAR10 team collected similar images for classes as shown in Figure \ref{fig:dataset_eg_cifar}, removed mislabeled images, and combined the hyponym of the query term from WordNet, filtered photorealistic images with single prominent instances only to obtain the CIFAR10 dataset. Classes in CIFAR10 are ordered alphabetically and are as follows: airplane, automobile, bird, cat, deer, dog, frog, horse, ship, and truck. The images in the dataset have 3 channels, have their classes in the middle surrounded by backgrounds, and tend to be symmetric about the horizontal axis~\cite{Krizhevsky2009LearningML}. 

\section{ImageNet}
For source tasks or pre-training purposes, we use the Large Scale Visual Recognition Challenge (ILSVRC2012), more commonly known as the ImageNet\-~\cite{imagenet} data\-set. Like CIFAR10, ImageNet is also collected based on 100k meaningful concepts in WordNet~\cite{wordnet} called the "synonym set" or "synset" but has a larger image size compared to CIFAR10. On average, a thousand images for each synset mostly nouns are provided by the ImageNet dataset. Samples of some classes of ImageNet ILSVRC2012 datasets are shown in Figure \ref{fig:dataset_eg_imagenet}. The primary purpose of choosing the ImageNet dataset is because of its wide range of 1000 classes, resemblance in the type of classes with the CIFAR dataset, and because it is one of the biggest and most popular datasets for image classification and pretraining.

More details of both the CIFAR10 and ImageNet datasets are shown in Table \ref{tab:dataset_details}. As Figure~\ref{fig:dataset_eg} shows, the samples  of objects in both of the datasets seem similar. We are not sure if the images overlap with ImageNet and CIFAR10, but since the transfer learning aspect of the network is common on TL only and TL+KD models and our focus is the application of KD, we do not explore this further. Along with similarities, there are many differences in the datasets which makes them ideal for our use case. ImageNet when downloaded is 150 GB whereas the CIFAR10 dataset is only 140MB. The ImageNet image size (i.e., 224 * 224) is 7 times larger than the CIFAR10 image size (i.e., 32 * 32). There are 1000 image classes in ImageNet and only ten classes in CIFAR10. The number of samples for each class is also around 1.88 million for ImageNet whereas it is just 60 thousand for the CIFAR10 dataset. Using such a massive ImageNet dataset helps us to provide representations to learn from and reuse in CIFAR10. Both datasets are open-source and free to use for educational purposes.
\begin{table}[!htbp]
\centering
\begin{tabular}{||l || r| r||} 
 \hline\hline
  & ImageNet & CIFAR10 \\ [0.5ex] 
 \hline
 Image Size & 224*224 & 32*32\\
 Number of Classes & 1000 & 10\\
 Training size & 1.28M & 50K \\ 
 Validation size & 50K & - \\ 
 Test Size & 100K & 10K \\[0.5ex] 
 \hline\hline
\end{tabular}
\caption[Baseline Dataset Details]{Details of the baseline dataset. ImageNet is 7 times larger in image size, has 100 times more classes, and almost 25 times more samples.}
\label{tab:dataset_details}
\end{table}

\chapter{Baseline Model}\label{chapter:model}
\label{chap:baseline_model}
For these experiments, we consider two different types of image-based networks to evaluate \sm{our setup explained in Chapter~\ref{chap:experiments} because of their popularity, simplicity, and performance. However, our entire pipeline, as shown in Figure~\ref{fig:KD_arch_approach}}, is network-agnostic and can also be applied to different types of modalities if required. 

\section{Convolutional Neural Networks}
Before diving into Convolutional Neural Networks (CNN), we describe what neural networks, in general, are and how they work. After going through standard neural networks, we explain the need for a convolutional neural network for images and how early CNN tackled image classification problems. We dive into deep CNNs from the modern era and go through recent well-performing architectures. 

\subsection{Feed-forward neural network}
Neural networks are computing agents inspired by and roughly modeled based on the operation of neurons in the human brain~\cite{keijsers2010}. Neural networks, also called artificial neural networks, are networks of neurons(nodes) where each node (except the input node) performs a weighted sum of values from all incoming nodes. A feed-forward neural network is commonly seen as the simplest form of neural network with an input layer, a number of hidden layers, and an output layer, where the weighted combination of the output of the previous layer is fed to the next layer until the output layer is reached. These networks are typically trained with gradient descent/ascent. Backpropagation algorithm~\cite{goodfellow2016} by comparing the model output with the target output and optimized with algorithms such as stochastic gradient descent (SGD)~\cite{kiefer1952}, adaptive moment estimation (ADAM)~\cite{kingma2014}. Adaptive gradient algorithm (AdaGrad)~\cite{duchi2011}.

\subsection{Early CNN}
Convolutional Neural Networks (CNNs) are feed-forward networks capable of considering the locality of the features. CNNs are very popular in computer vision  and natural language processing providing state-of-the-art results on problems such as image classification and text classification~\cite{khan2020}. Although vanilla neural networks are simple to use, they are not capable of capturing the spatial relationship between features, which is very important in computer vision tasks. CNNs are known to be very good at picking up patterns in the input image, such as lines, gradients, circles, or even eyes and faces. It has good image-specific inductive biases like locality, translation equivariance, 2D-neighborhood structure, etc. ~\sm{Inductive bias refers to the set of assumptions the model learns to make that can be generalized beyond a given training dataset. The inductive bias of locality learns that the pixels that are closer together are similar or closely related. Translation equivariance, on the other hand, describes that objects which are translated within an image give equivalent CNN representations. 2D neighborhood structure represents the CNN network maintaining the 2D neighborhood of pixels throughout the convolutional layers.}

LeNet~\cite{lecun1989, lecun1998} was one of the first convolutional neural networks to give a new direction to deep learning for image classification tasks. LeNet used a neural network trained by backpropagation algorithm~\cite{kelley1960gradientbackprop} to read handwritten numbers and successfully applied it to identify handwritten zip code numbers \sm{dataset} provided by the US Postal Service called MNIST~\cite{lecun-mnisthandwrittendigit-2010}. The availability of ImageNet data~\cite{deng2009} through ImageNet's large-scale visual recognition challenge (ILSVRC)~\cite{russakovsky2014} since 2010 has played an important role in the development of modern CNN architectures. AlexNet~\cite{krizhevsky2012} was the first CNN-based model to win the ImageNet (ILSVRC) challenge in 2012 and served as a huge milestone in deep learning for images. AlexNet has 8 layers in total, with 5 convolutional layers and 3 fully connected layers trained on the ImageNet dataset.  It had 60M parameters and achieved 63.3\% accuracy. It also introduced a normalization layer, called a response normalization layer, which normalized all values at a particular location across the channels in a given layer. Furthermore, it introduced the rectified linear unit (ReLU)~\cite{fukushima1975} as an activation function and has about 60 M parameters.

\label{chap:model}
\subsection{ResNet}
Much of the success of deep neural networks was attributed to the additional deep layers with the intuition that the layers progressively learn more complex features. Deeper networks are harder to train, both from a hardware and software perspective.~\cite{he2016deepresnet} showed that there is a threshold for the depth of traditional CNNs  empirically by plotting the error of a 20-layer CNN versus a 56-layer CNN~\cite{he2016}. Because of vanishing gradients, the performance would drop after a certain point if we keep adding layers to a Neural Network. To combat this, they introduced ResNet~\cite{he2016} with a residual learning framework as a better alternative to traditional CNN models. They reformulated the layers as learning residual functions with reference to the layer inputs rather than learning the original functions. The authors of the paper claim to have alleviated the difficulty of training deep networks with the introduction of \textit{the residual block}, shown in the equation below. 
\begin{equation}
    x' = I(x) + F(x, W_i)
\end{equation}
where $x$ refers to input to residual connection,  $I(x)$ as identity mapping, and F refers to the section of the network from the input of the residual block to the output where it satisfies the identity function. "This skip connection identity mapping does not have any parameters and is just there to add the output from the previous layer to the layer ahead"~\cite{skip_connection}. Adding the residual block makes it easier for the model to train deeper networks by retaining the information learned earlier in the networks without losing it due to depth. Due to the residual network, it is much easier to propagate gradients through deeper networks, as skip connections tend to copy gradients throughout the sub-network $F(x, W_i)$. It had more than 11 million parameters with 18 to 152 layers and was able to achieve 75\% accuracy on ImageNet. Visualizing loss landscapes~\cite{li2018visualizing} also shows an easier and smoother loss surface with skip connections. WideResnet~\cite{zagoruyko2016wide} showed wider networks compared employed better feature reuse and perform better than thinner and deeper counterparts.
    
\section{Image Transformers}
\subsection{Attention, Transformers, and BERT}
This section focuses on the origin and motivation for developing an attention mechanism in neural networks. \sm{Attention mechanism refers to the act of attending to just a relevant region of input instead of the whole input image or sentences. In contrast to encoding a variable-length sequence into a fixed-length vector~\cite{sutskever2014sequence}, Bahdanau et. al.~\citep{bahdanau2014neural}} introduced the attention mechanism \sm{with encoder-decoder architecture. In encoder-decoder architecture, a neural network called an encoder is used to encode the training input to a representation. After that, another network called a decoder is used to decode the representation to get the output. In attention-based networks, the decoder} can essentially attend/focus on a specific region of the input/source sequence directly. This allows the model to learn the corresponding mapping of relevant words in different sentences for natural language processing. 

Diving further into attention, \sm{Vaswani et. al.}~\citep{transformer} proposed to remove the whole sequence/recurrence bit from the model. Removing recurrence would enable models to be trained in parallel on massive datasets in significantly less time. They fundamentally only used a type of attention to process the whole sequence in parallel with the help of positional encoding. This model was called Transformers\citep{transformer} which would employ self-attention, which allows the model to learn representations by looking at the input sequence itself. This Transformer model\citep{transformer} actually revolutionized the NLP research as we know it today.

With the help of self-attention in transformers, Transformers encoders~\cite{transformer} learn unidirectional representations of words with the help of context. BERT~\cite{devlin-etal-2019-bert} learns representation looking at both forward and backward section inspired by Cloze procedure~\cite{taylor1953cloze}. To learn the word-word interaction, BERT masks a word and uses context around it to predict the masked word. The intuition is that the model will be able to predict masked words once it learns about the syntax and semantics  of the words. It learns more general sentence-level representation with the help of proxy tasks like Next Sentence Prediction (NSP). The representations that it learns were so rich that when applied to multiple downstream tasks ranging from machine translation, Question Answering, etc. BERT works in the form of transfer learning, i.e., pre-training with massive datasets and fine-tuning in downstream tasks with limited labels to gain significant improvements in performance.

   \subsection{Vision Transformers}
While self-attention-based architectures like Transformers~\cite{transformer} were becoming the go-to model of choice in NLP, in computer vision, convolution-based architectures were still the model of choice. \sm{Transformers architecture lacks inductive bias for images~\cite{bit} such as translation equivariance that is inherently present in CNN-based models. In spite of this, Vision Transformers~\cite{dosovitskiy2020imagevit} were able to achieve better performance on multiple image recognition tasks with the help of large-scale training, which allowed them to learn such inductive biases from scratch}. To avoid the quadratic cost in a number of pixels, ViT~\cite{dosovitskiy2020imagevit} approximates the input space by converting images into fixed-sized discrete patches and treated as input tokens of NLP Transformers~\cite{transformer}. Because of its formulation, ViT, through massive pertaining, was able to achieve state-of-the-art results on multiple image-related downstream tasks for the first time using Transformers~\cite{transformer}. ViT has more than 86 million parameters with 12 to 32 transformers layers and achieves 88\% accuracy~\cite{paul2021vision} on ImageNet. Like ViT, there are numerous Transformer based Foundation models~\cite{bommasani2021opportunities} for image classification as well as a wide range of other domains like NLP, Robotics, Image, etc. However, we leave exploring those models for Knowledge Distillation and Transfer Learning for future work. 


    
The details of the parameters of the three models used in this thesis are shown in Table \ref{tab:model_details}. During transfer learning, all or part of each network is retrained/fine-tuned, and the rest are frozen and are not updated during any iteration. ViT has a maximum number of total parameters but has the lowest trainable parameter from scratch to total parameter ratio. We reuse many layers and parameters because the model weights for earlier layers are generalizable across our source and target tasks. 
\begin{table}[!htb]
\centering
\resizebox{\linewidth}{!}{%
\begin{tabular}{||l ||r| r| r||} 
 \hline
  & Resnet18 & WideResnet & ViT\\  
 \hline\hline
 Number of trainable Parameters (scratch) & 5,130 & 20,490 & 7,690\\
 Number of Frozen Parameters (re-use) & 11,181,642 & 66,854,730 & 85,806,346\\
 \hline
 Total Parameters & 11,186,772  & 66,875,220 & 85,814,036 \\  
 \hline\hline
  ImageNet Acc@1 & 69.76 & 78.47 & 85.30\\
 ImageNet Acc@5 & 89.08 & 94.09 & 97.65\\
 \hline\hline
\end{tabular}}
\caption[Model Parameter Details]{Detail of the parameters of the baseline model. The student network Resnet18 has the lowest number of total parameters, while ViT has the highest. The number of trainable and frozen parameters is when only the last fully connected layer is fine-tuned with all the backbones frozen. The bottom rows show the Top-1 and Top-5 accuracy for each TorchVision pre-trained model~\cite{paszke2019pytorch} on ImageNet.}
\label{tab:model_details}
\end{table}

\chapter{Experimental Setup}
\label{chap:experiments}
\section{Transfer Learning Framework}
\begin{figure}[!htb]
\centering
\includegraphics[width=\textwidth]{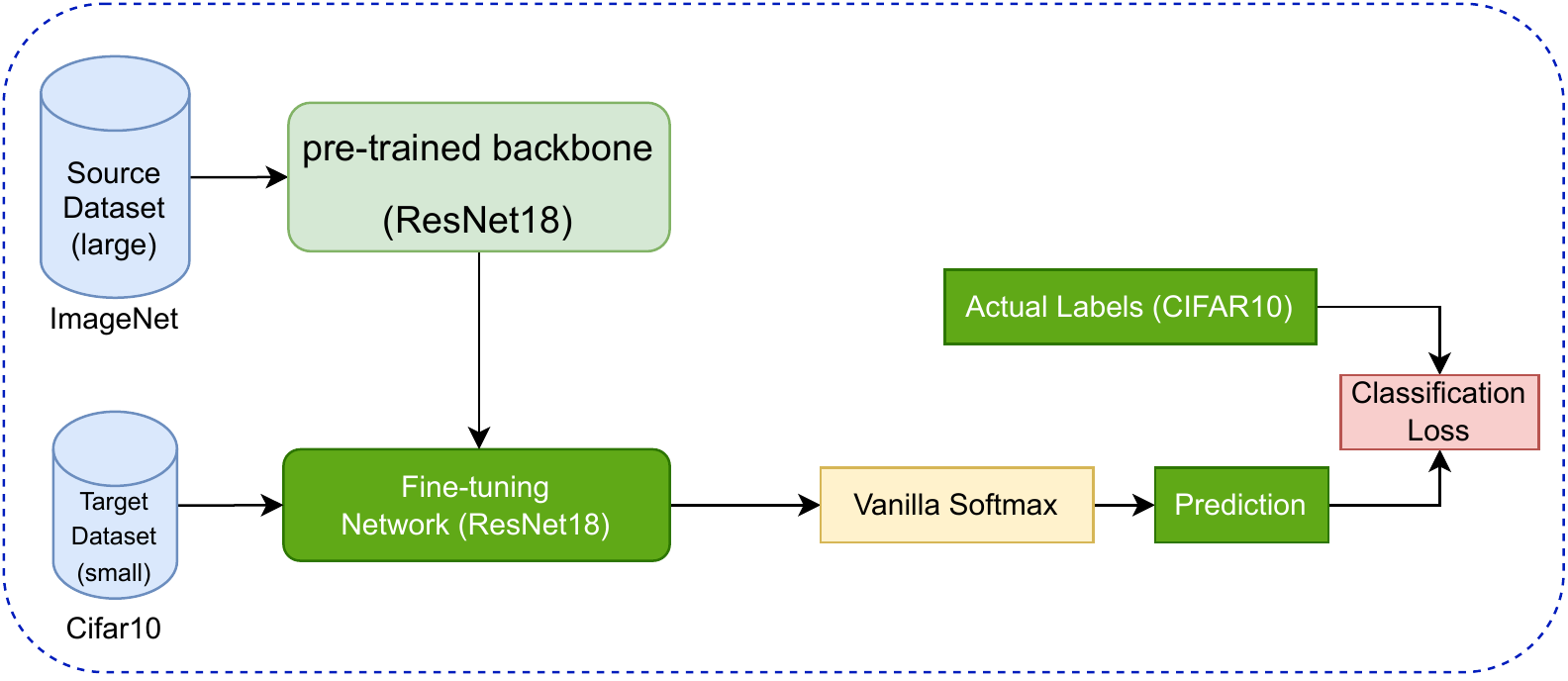}
\caption[Transfer Learning Block Diagram]{Transfer Learning Block Diagram. ImageNet pretrained network is fine-tuned with the CIFAR10 dataset. We used vanilla softmax scores with T=1 for this setup. Classification loss is used to compare and calculate loss between the prediction of the network with actual labels from the CIFAR10 dataset. This TL setup is also present in TL+KD models.}
\label{fig:tl_arch}
\end{figure}

In our experiments, we use the resnet18 model that is pre-trained on the ImageNet dataset with 1000 classes. All layers of resnet18 that were used to learn the ImageNet dataset except the last one are transferred to a different dataset classification problem. Here, we do not train the resnet18 model from scratch but reuse the pre-trained weights that have been trained on the ImageNet dataset previously and fine-tune them on the CIFAR10 dataset. Figure~\ref{fig:tl_arch} shows the block diagram for the transfer learning only model. 

\section{Transfer Learning with Knowledge Distillation Framework}
\label{ex_approach}
\begin{figure}[!htb]
\centering
\includegraphics[width=\textwidth]{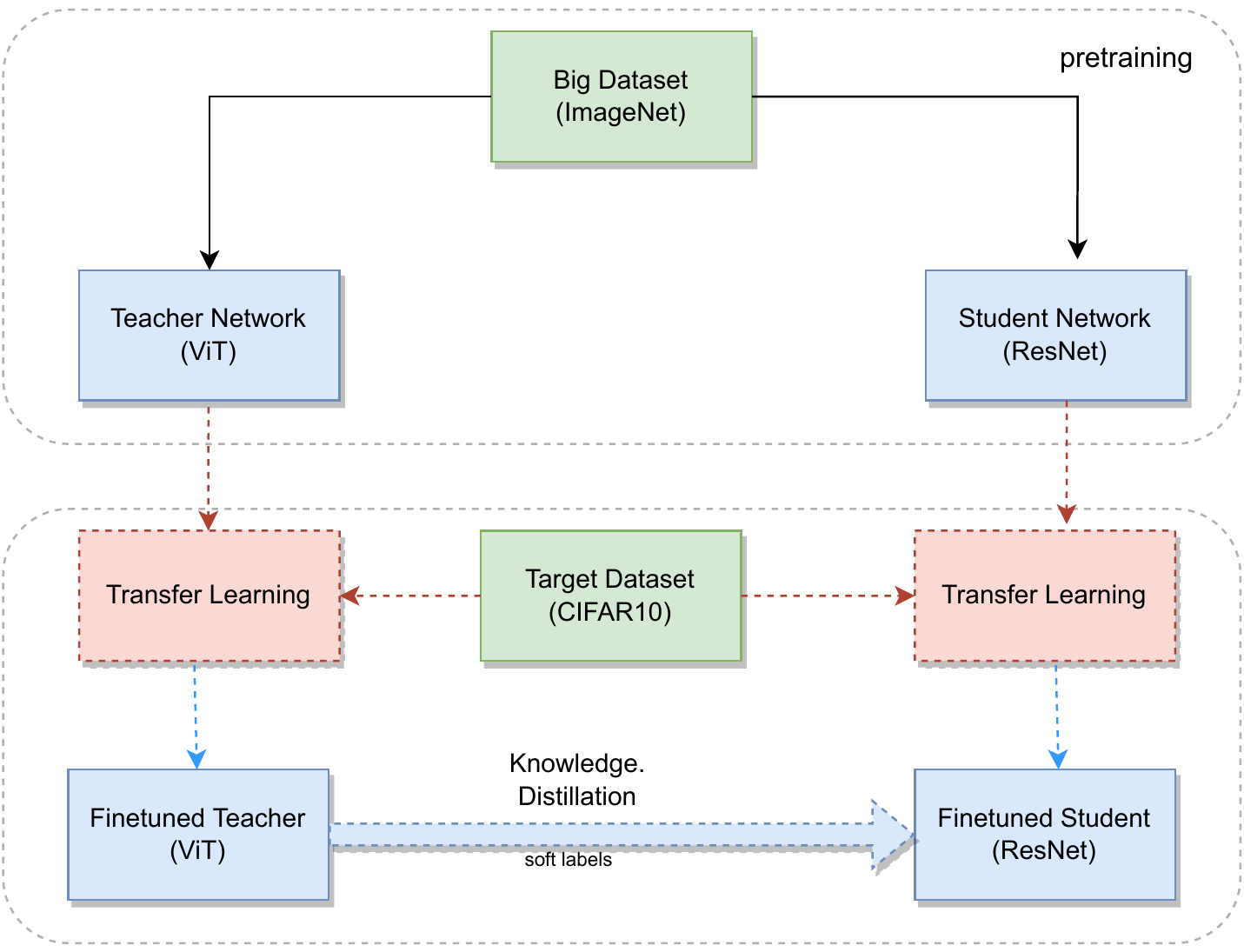}
\caption[TL+KD Block Diagram]{TL+KD Block Diagram. Once both the large teacher network and the smaller student network are fine-tuned with the target CIFAR10 dataset separately, the fine-tuned teacher network is used to provide soft labels to improve the student network further. \FIX{Both of these fine-tuned student network and fine-tuned teacher network are used in TL+KD fine-tuning process in Figure~\ref{fig:KD_arch_approach} }.}
\label{fig:kd_dist_arch}
\end{figure}

In our experiments, we use and explore the response-based offline knowledge distillation method to test our hypotheses on transfer learning due to its simplicity, popularity, and seminal contributions. The architecture of the methodology that describes the process of applying knowledge distillation to transfer learning for all hypotheses 1-4 in our thesis is shown in Figure \ref{fig:kd_dist_arch}. We have two types of networks, teacher network, and student network. The teacher network is generally deeper or larger than the student network. Here, we use large networks such as Vision Transformer~\cite{dosovitskiy2020imagevit} as a teacher network and ResNet18~\cite{he2016deepresnet} as a student network. Both are described in detail with their motivation in Chapter~\ref{chap:model}. Both are pre-trained with a large image recognition dataset called ImageNet~\cite{imagenet} with 1000 classes. Then both networks are fine-tuned with the baseline CIFAR10 dataset to obtain a fine-tuned version of the models. To improve the representations learned by the ResNet student network, we provide response-based knowledge distillation from the teacher network. The student is then fine-tuned along with the soft labels from the Teacher network. We compare a transfer learning only variant of the model with the variant with knowledge distillation with a teacher to investigate our research questions. Since we refer to these cases frequently, we sometimes use the following short notation. TL: transfer learning, KD: knowledge distillation, and TL + KD: transfer learning variant with knowledge distillation. 

\begin{figure}[!htb]
\centering
\includegraphics[width=\textwidth]{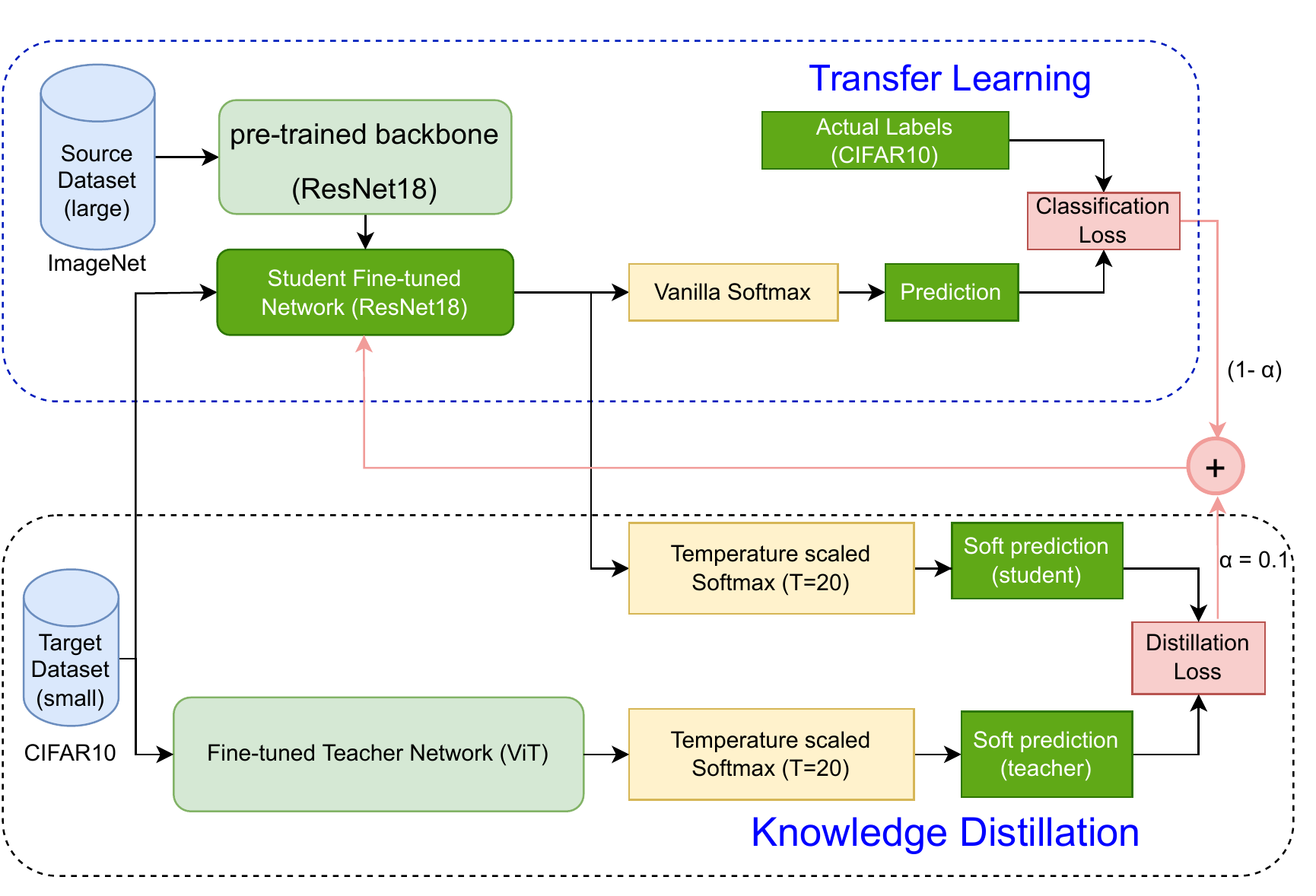}
\caption[Fine-tuning process of Student network with KD]{Fine-tuning process of the student network with KD. The upper half represents vanilla Transfer Learning, and the lower half represents the knowledge distillation process. Temperature-scaled soft predictions (for T=20) are obtained from the pre-trained teacher network and the student network, respectively, and compared to obtain distillation loss. The student network improves with a combination of the weighted sum of distillation loss and original student loss with $\alpha=0.1$. Dark green blocks are used during fine-tuning, and light green blocks are not trained during any of these processes. The red arrow shows the propagation of loss and optimization using both losses. \FIX{The process of obtaining finetuned version of teacher and student networks are shown in Figure~\ref{fig:kd_dist_arch}.}}
\label{fig:KD_arch_approach}
\end{figure}

Figure \ref{fig:KD_arch_approach} describes the fine-tuning process in more detail. The lower half section of the figure shows the process of both getting knowledge and distilling knowledge, whereas the upper half shows how the process of vanilla transfer learning process takes place. Knowledge here represents the pre-trained Teacher network and its predictions. \sm{In general, the teacher network is large and trained on a dataset with a strong regularizer for long epochs, whereas the student network is typically a smaller network more suitable for deployments.} The transfer learning process is the same for both TL-only and TL+KD models, and the lower half is our contribution. The student tries to imitate soft labels provided by the Teacher with the help of temperature-scaled Softmax simultaneously, as it tries to imitate hard labels from ground-truth with logits from vanilla Softmax. The classification problem with vanilla TL+KD learns with the combined loss of soft and hard labels during fine-tuning. KL Divergence Loss is used as a loss function for distillation, where standard cross-entropy loss is used for learning to classify ground-truth labels. 

\section{Hypothesis 1}
\label{sec:ex_training}
We describe the details of various training procedures in this section. First, we describe how we train vanilla networks for students and teachers, then how the teacher is used for distilling knowledge to provide soft labels to students, and finally, some more details on obtaining the evaluation metrics scores with interpretations \sm{for Hypothesis 1. Hypothesis 1 says, ``TL+KD architecture improves image classification over the
TL architecture on the CIFAR10 dataset".} Most of the part explained here overlaps with other hypotheses, so we explain common details here and refer to them later when needed in other hypotheses. 

\subsection{Vanilla Student Training}
\label{TL}
For the baseline student model for all hypotheses 1-4 in Chapter~\ref{chap:results}, we use a student ResNet~\cite{he2016deepresnet} architecture. We fine-tune the baseline student ResNet model without any knowledge distillation or logits of teacher models initially to get a TL-only model. The detailed architecture we use for the student network for all experiments is ResNet-18~\cite{he2016deepresnet}. We chose resnet18, described in detail in Chapter~\ref{chap:model}, specifically for its small size yet reasonable performance and because of its popularity for transfer learning. First, we train ResNet18 using SGD optimizer with a momentum of 0.9 and a weight decay rate of 0.0005 for 200 epochs. Momentum helps to improve the convergence speed of the model by helping it accelerate in the right direction. Weight decay is a regularization technique that adds a penalty to the model's weight to keep it within reasonable ranges. 

For further regularization, we used dropout with zero probability of 0.2 and early stopping with the patience of 10 epochs. With early stopping, the model stops training and exits the loop if the validation performance does not improve for \sm{10 epochs as provided by patience value}. Using a dropout of 0.2, that is, turning off 20\% nodes \sm{randomly} in a layer, allows us to add regularization to the model and improve overfitting by allowing individual nodes to be stronger~\cite{JMLR:v15:srivastava14a}. \sm{The learning rate was reduced by $10\%$ if model performance specified by validation accuracy did not improve for 3 epochs}. We set the initial learning rate and mini-batch size to 0.01 and 64, respectively. Most of these hyperparameters were chosen from the previous literature~\cite{xie2020understandingweightdecay}. Some of them, such as learning rate, learning rate(LR), scheduler patience, and early stopping patience, were changed through manual hyperparameter tuning by training multiple runs and evaluating the convergence and final performance. 

We also follow standard data augmentation techniques: 32x32 pixels images are resized to 224x224 pixels with bilinear interpolation to make it consistent with the input of the teacher network later and also consistent with the size of the source ImageNet dataset~\cite{imagenet} which is followed by a random horizontal flip. \sm{Adding horizontal random flip helps to improve the robustness of the model towards the left and right direction in images across all classes in the dataset}. Finally, all scores are converted to the PyTorch Tensor for processing and normalized with mean values
$[0.4914, 0.4822, 0.4465]$ and standard deviation of $[0.2471, 0.2435, 0.2616]$ in three channels. We get these values by selecting all the training sets of the CIFAR10 dataset and calculating its mean and standard deviation of pixels across all three channels separately. 

For fine-tuning, the last layer defined by 'fc' in CNN is replaced by a fully connected classifier head with output size as the number of classes, that is, 10 for CIFAR10. All the parameters in the backbone except the newly added fully connected head are frozen and not updated during optimization. For Transfer Learning, instead of starting with random weights, we start the training initialized with pre-trained weights for ImageNet classification.  ImageNet weights are provided by the PyTorch~\cite{paszke2019pytorch} TorchVision package. For both training and validation, we use the train-test split provided by the CIFAR10 dataset. These splits are consistent with all other experiments in this domain for image classification tasks. 

The ResNet model trained here is the TL Only variant reported in all the Results~\ref{chap:results} chapter. Likewise, the ViT model trained here will serve as a Teacher network when training ResNet under the TL+KD paradigm later. 

\subsection{Vanilla Teacher Training}
For the knowledge distillation for all hypotheses 1-4, we train a teacher network to provide soft labels to the student. The detailed architecture that we consider for the teacher is called ViT-B/16~\cite{dosovitskiy2020imagevit} and is a vision transformer described in chapter~\ref{chap:model}. We chose the ViT model for its performance, which is among the highest top-1 accuracy for ImageNet weights in the TorchVision package~\cite{paszke2019pytorch}. Another reason for choosing a Transformer-based vision model is that it would be compatible and relatively easier to transfer to other modalities of the dataset where Transformer-based models are very effective~\cite{bommasani2021opportunities}. The hyperparameters used for this training are consistent with the ResNet student training above, except for a few things. During training the teacher model, we replace the last classifier layer named "head" instead of "fc" in resnet18. All other details of the training for the vanilla teacher network are the same as those of the vanilla student network in Section~\ref{TL}.

\subsection{Knowledge Distillation}
\label{KD}
For TL+KD in all hypotheses 1-4, we use the saved trained version of the ViT model in evaluation mode. The weights of all layers in ViT are frozen and are not updated during this training. The primary purpose of ViT in this training paradigm is to use the images in batches and to generate soft labels that will be used while training a new ResNet student. A temperature value of $T=10$ is used after manual hyperparameter tuning to obtain Softmax soft labels for teacher and student networks. The loss of KL divergence is weighted by the distillation weight hyperparameter $\alpha$ of $0.1$ based on previous literature~\cite{efficacykd}. \FIX{We verified that this $\alpha$=0.1 gave better validation accuracy ($96.34\%$) than either of the two pure trainings corresponding to $\alpha$=0 (validation accuracy 93.09\%) and $\alpha$=1 (validation accuracy 91.99\%) with the same setup over a single run}. A detailed setup for both vanilla TL only and TL+KD training framework is given in Section~\ref{ex_approach}.

\label{ex_details_pytorch}All experiments were performed with PyTorch~\cite{paszke2019pytorch} on the NVIDIA A100 PCIe GPU system provided by New Mexico Tech and Google Cloud Platform clusters. It takes around 1-2 hours to train each of these models.

\subsection{Evaluation}
\label{ex_metrics}
We use top-1 accuracy, precision, recall, and \sm{F1} scores as percentages as standard performance measures for the multiclass image classification task in Hypothesis 1. In top-1 accuracy, the model prediction should have the same winning class as the ground truth labels. For the \sm{F1} score, we calculate the precision and recall of the classification and calculate the harmonic mean of the precision and recall value with the same weights. Precision provides insights on the percentage of correct predictions from all true positive classes, while recall calculates the correct predictions from all positive classes. F1 score provides better insight into incorrectly classified cases than the accuracy scores because the harmonic mean penalizes extreme values in precision and recall. 

The exact formulation of accuracy and F1 score can be further described with the help of a confusion matrix. 
If TP, TN, FP, and FN are the True Positive, True Negative, False Positive, and False Negative values of a Confusion Matrix, 
$$Accuracy = \frac{\text{True Predictions}(TP + TN)}{\text{All Instances}(TP+TN+FP+FN)}$$ 

$$ Precision = \frac{\text{Correct Prediction}(TP)}{\text{All Positive Predictions}(TP+FP)}$$ 

$$Recall = \frac{\text{Correct predictions}(TP)}{\text{All Positive Labels}(TP+FN)}$$

$$\sm{F1\ score} = \frac{2*Precision*Recall}{Precision+Recall}$$
Here, true positive means the number of positive predictions that are actually positive. True negative means the number of negative predictions that are actually negative. False positive, on the other hand, refers to the case in which images are predicted as positive but actually were negative. Lastly, false negative refers to the case in which images are predicted to be negative but are actually positive. 
We provide all accuracy, precision, recall, and F1 scores for the quantitative evaluation of baselines with and without the influence of transfer learning for transparent evaluation for Hypothesis 1. Since we are dealing with balanced classes and all other evaluation metrics showed similar responses, we use validation accuracy only for hypotheses 2-4. 
A confusion matrix is also used to provide some more insights on the classification task with both TL only and TL+KD models in hypothesis 1. It helps us identify the confusions that occur in classification prediction. It is represented by a $N*N$ matrix($N$ is the number of classes) where rows are the actual ground-truth classes from the CIFAR10 validation set for each class, while columns are the predicted classes. Each of the classes has $1k$ images to evaluate, and when all the values in the row are added, we get the total images for each class. Numerically, the confusion matrix gives a matrix. For our case, since we have 10 classes in the CIFAR10 dataset, we get a 10*10 matrix of scores with shaded cells according to the number of images in that particular case. We also provide this matrix as part of the quantitative reporting results for the effect of knowledge distillation on our baseline problem.

\label{ex_interpret}
\subsection{Interpretability}
Out of all post-hoc explanation techniques, we use the gradient-based method to avoid the dependence of perturbation-based methods on the randomness of perturbation of input samples while testing hypotheses 1 and 3. Due to such dependence, methods such as LIME~\cite{ribeiro2016should} tend to produce unreliable interpretations that often change with different runs. We also do not explore the attention-based method, even if we use a transformer-based model, due to the debate about its validity as an explanation and its vulnerability to manipulation~\cite{wiegreffe-pinter-2019-attention, serrano-smith-2019-attention}.
For qualitative evaluation and obtaining interpretations, we use the 'DeepExplainer' module from SHAP~\cite{lundberg2017unified} library, which is described in detail in Chapter~\ref{intro_shap}. With this method, for an image, we get a \sm{multidimensional} array of contribution values for all the classes. Each interpretation of classes has positive and negative values shown in red and blue colors, respectively. Red color for a particular class represents the pixels that are responsible for predicting it as that class, i.e., positive contribution, and blue represents the pixels that are responsible for not predicting it as the same class, i.e., negative contribution. We used the first $100$ images from the data loader to get the mean values of the dataset to be used as a baseline/background, as suggested by the library. \FIX{Passing the background gives a prediction for each class which is called as expected values of 'DeepExplainer' for each class. We then pass the test images to calculate the approximate SHAP values with DeepExplainer on a particular sample of interest. As a result, we get an array of SHAP value contributions to predicting the image as all $10$ classes of the CIFAR10 dataset. These contributions are additive in nature, i.e., when all of the contributions for a particular class are added together with an expected value of the DeepExplainer on the background for that class, we get the actual output prediction of the model for that class. If the model is a classification model with the last layer as a Softmax, the model's prediction adds up to 1. Otherwise, if it is regression model or classification model without softmax as the last layer, model's output can be any real value.\label{sec:shap_addition}}

To visualize the SHAP contributions, we first need to convert the shape of the SHAP value array from (N, 1, C, H, W) to (N, 1, H, W, C), where N = the number of classes, C = Number of channels, H and W is Height and width of the image which is 224*224 after being resized as motivated in Section~\ref{sec:ex_training}. To \sm{export} images with contribution representations, we also need to reorder the axis of the test array like that of the SHAP array and denormalize the images with the mean and \sm{standard deviation} values used to normalize the images as described in the \sm{vanilla student training Section ~\ref{TL}.}

\section{Hypothesis 2}
\label{ex_interpret_quant}
\begin{figure}[!htbp]
\centering
\begin{subfigure}[b]{\textwidth}
   \includegraphics[width=0.84\linewidth]{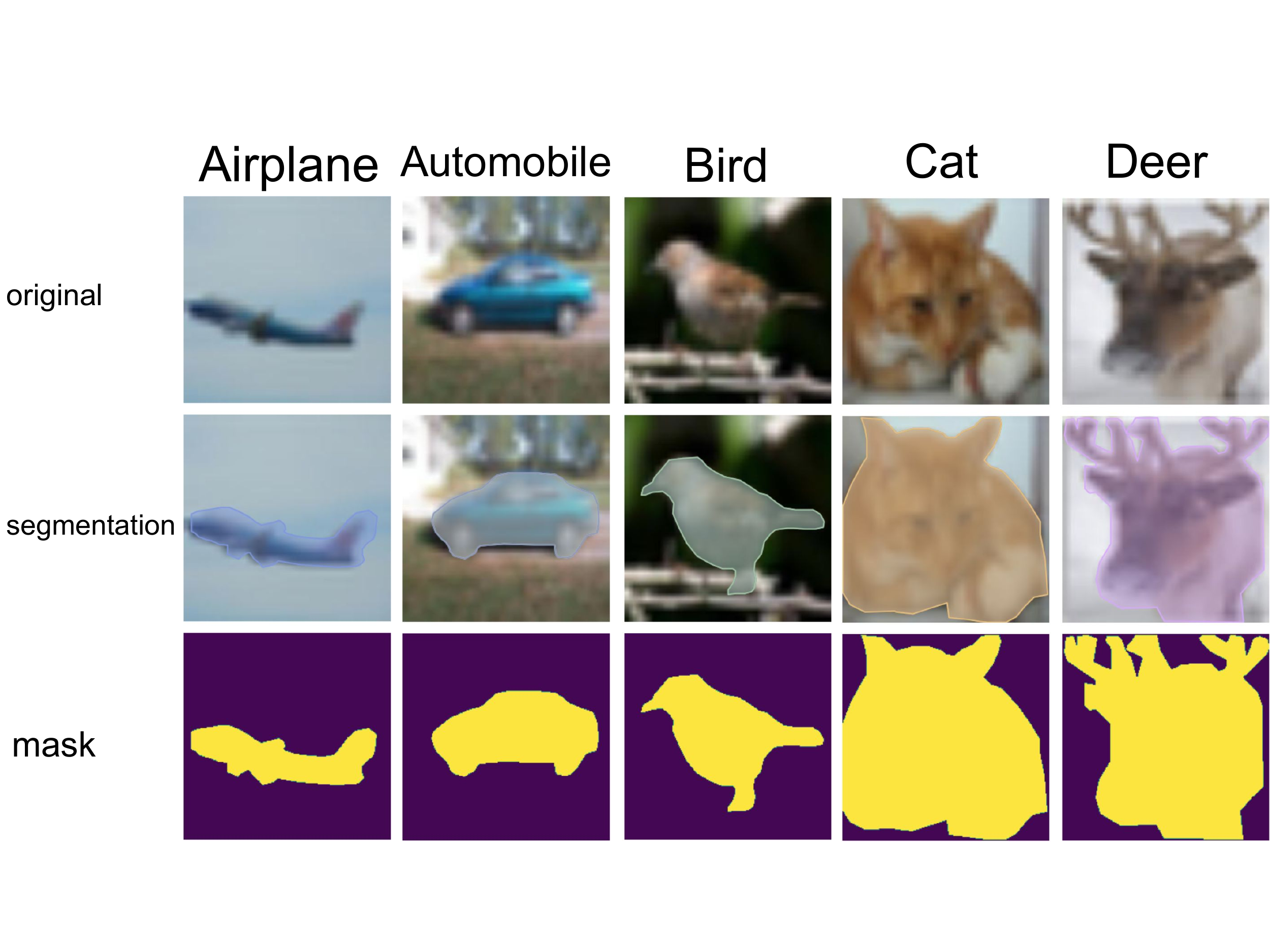}
\end{subfigure}
\begin{subfigure}[b]{\textwidth}
   \includegraphics[width=0.84\linewidth]{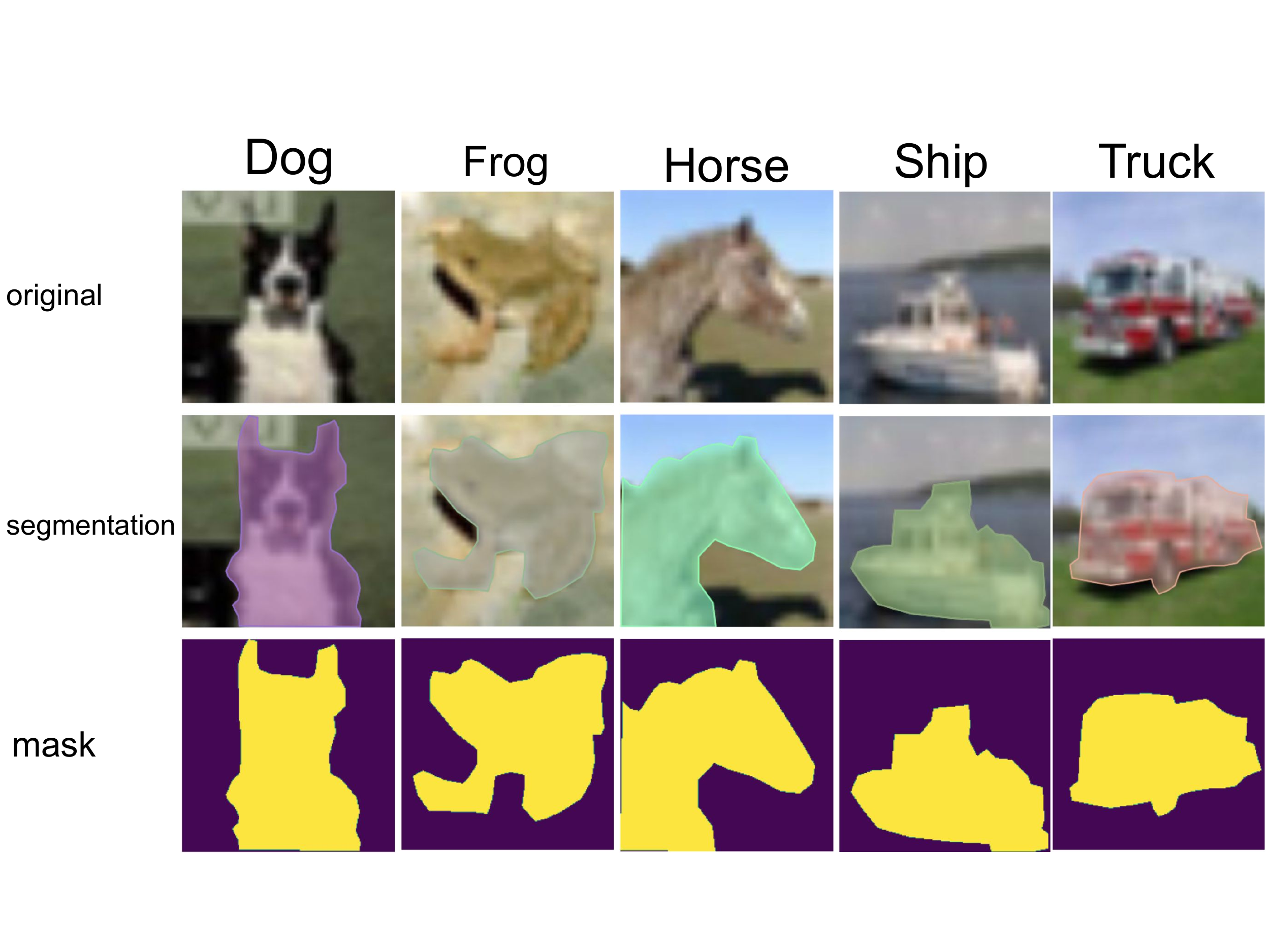}
\end{subfigure}
\caption[Three-step Image Segmentation Process]{Three-step segmentation process for samples for each class. top half = segmentation for the first five classes of CIFAR10, bottom half = last five classes. The original image from the top row is used to draw a segmentation map manually like the middle row, and the segmentation mask, like the bottom row, is exported.}
\label{fig:segmentation_process}
\end{figure}

Hypothesis 2 says, ``TL+KD provides more contribution on foreground image features measured by SHAP contribution than TL only". To test this hypothesis, we analyze the local interpretations of one example each for all classes. The SHAP interpretations of student and teacher networks for selected images for each class are obtained from an experiment from hypothesis 1. Every single details to train these models are same as in hypothesis 1. Additionally, to quantify the SHAP interpretations of student networks trained as both TL only and TL+KD setup, we assign each pixel of the images to foreground and background based on if it falls over the actual object or outside it. We use the VGG Image Annotation tool~\cite{dutta2016vgg} to draw and collect the annotation of the image due to its lightweight and ability to run only in browsers. We export the annotations as the Microsoft COCO~\cite{lin2014microsoftcoco} image annotation format and save the vertices that define the concave hull surrounding the region defining the class foreground. Once the map is loaded, we build a foreground mask that differentiates the foreground from the background. We extract the SHAP interpretations within the mask, i.e., from the foreground, and analyze those further. The results of all these processes are summarized in Figure~\ref{fig:segmentation_process}. 

To compare the absolute values of the SHAP interpretations, we divide the SHAP values according to positive and negative signs. Values greater than zero are positive contributions, and those less than zero are negative contributions. Take the sum of positive contributions within the foreground and negative contributions in the foreground to get the values \textit{pos} and \textit{neg}. The primary purpose of using absolute sums instead of means is to avoid losing information on the actual range of the peaks so that we can compare the absolute differences \textit{diff} between them. These values \textit{pos}, \textit{neg} and \textit{diff} for the background and foreground are calculated separately for the TL only and the TL + KD variant and compared to obtain the results in Section~\ref{res_shap_table_explain}. \FIX{Adding all \textit{pos} and \textit{neg} (or just adding \textit{diff}) for both foreground and background together for an image sample passed through a particular model gives net SHAP contribution of the sample responsible for predicting it as a particular class. Adding that net contribution for the winning class with an expected value of background prediction for that same class gives the prediction of the model as described in Section~\ref{sec:shap_addition}. Since the SHAP interpretations are local in nature and are different for different samples, we only compare them within the same sample and avoid comparing across samples/classes in this hypothesis. We can compare across TL only and TL+KD model because the student model's architecture for both scenarios is the same (the only difference being the inclusion of soft labels during training).}

One of the limitations of this experimental setup is that since the segmentation map of CIFAR10 is not easily available, we had to annotate the mask over the pixels of images manually, and it was very time-consuming. For this reason, we randomly select one image from each class to represent their cases and SHAP interpretations. We believe that this should be sufficient to test the main hypothesis of Hypothesis~\ref{res_segmention} i.e., to explore the changes of focus on the foreground by different models. 

\section{Hypothesis 3}
Hypothesis 3 says, ``TL+KD achieves similar validation accuracy faster with fewer training epochs than TL only model". The process of training models is the same as that of hypothesis 1 trained over 3 runs with different random seeds(0, 7, 42). We track the validation accuracy at different iterations to test out these hypotheses. Any remaining details are explained as needed in section~\ref{res_faster_convergence}.

\section{Hypothesis 4}
\label{ex_complexity}
Apart from the common things above, we add a few additional elements for Hypothesis 4.  Hypothesis 4 says, ``TL+KD improves validation accuracy even after adding training complexities such as training with a fraction of data or corrupted images". When the training fraction is mentioned, only a subset of training data per train-fraction hyperparameter is used to add complexity to the network \sm{training}. The validation set is not touched on for any of these additional complexities. To test hypothesis 4 in Section~\ref{res_training_frac} during our initial investigation, we used a different teacher network, i.e., WideResNet, to teach the student described in detail in Chapter~\ref{chap:model}. The whole process of building a teacher network is the same as building the student and teacher, as described in Section~\ref{TL} except for a few things. We worked with the original size of the input, i.e., 32*32, since this time, we did not have to make it consistent with the teacher network as we did for the ViT network. Some hyperparameters were also different from those reported in other experiments above. The initial learning rate of 0.1 was used from literature~\cite{zagoruyko2016wide} and decayed with the same LR scheduler above. Early stopping was not yet employed. The mean and standard deviation values for normalization were the same as reported above. The initial weights for both networks were also taken from PyTorch ImageNet pre-trained weights~\cite{paszke2019pytorch} like before. All these hyperparameters are the values that gave the optimal validation performance for this setup. Since this was our initial experiment, we focused more on the differences between the TL and TL+KD models than on optimizing the absolute values of the validation performance of each of the models.

In the second part of the experiments with label noise, we select a fraction of the training dataset(using train\_fraction \sm{hyperparameter}) from the batch and randomly assign different labels to that subset during training. Everything else during training is the same as before. 

\begin{figure}[!htb]
\centering
\includegraphics[width=\textwidth]{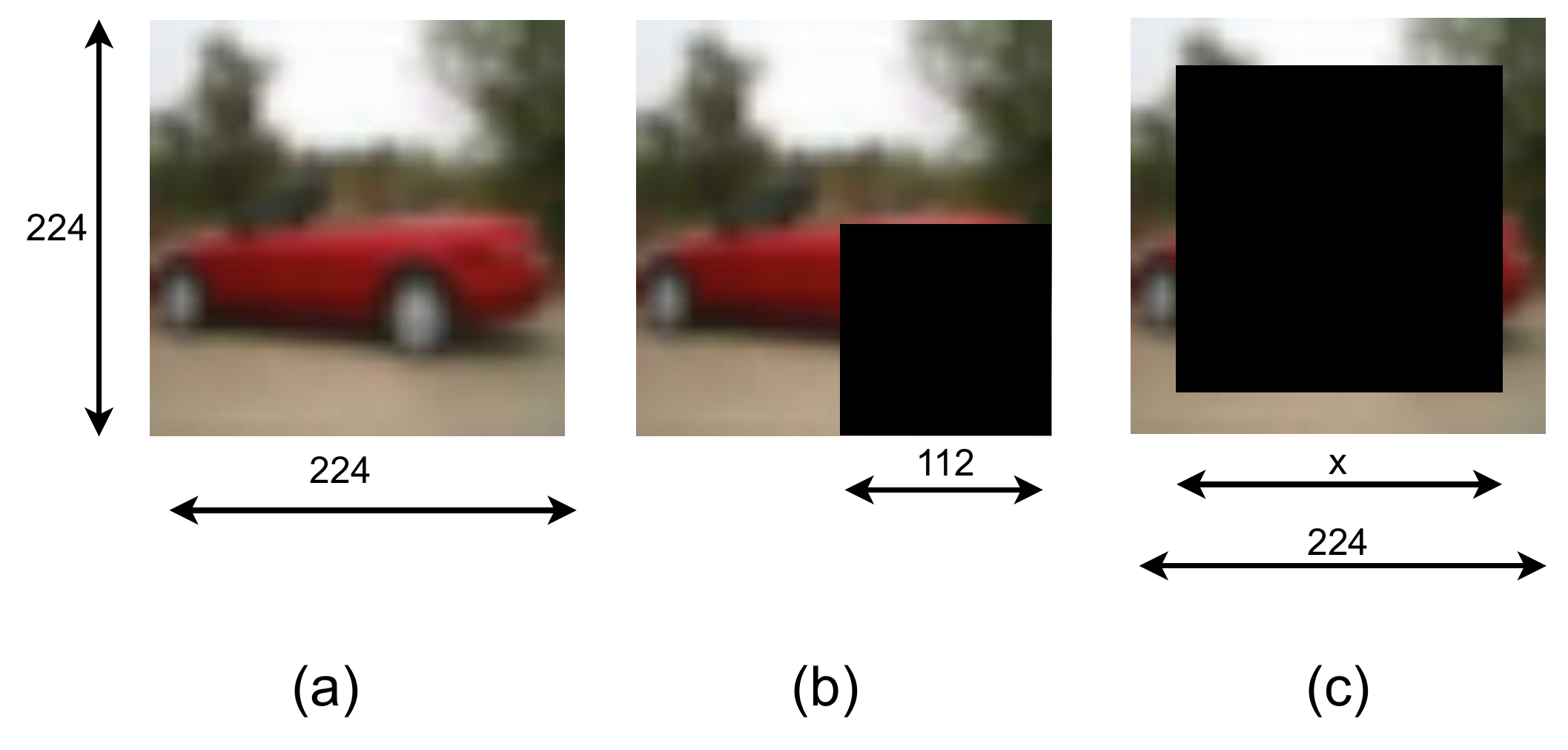}
\caption[Examples of QuarterBlack and CenterBlack augmentations]{Examples of QuarterBlack and CenterBlack augmentations. a) original image of size (224,224) pixels, b) QuarterBlack image corruption method with a random black patch on the bottom right. c) CenterBlack image corruption method with random size $(x,x)$. The center of both the patch and the image falls at the same point.}\label{fig:complexity_aug}
\end{figure}

When CenterBlack is mentioned in hypothesis 4, we introduce random CenterBlack augmentation with  \sm{square patches with uniform random side length with range} $x<224$ to the full $224*224$ size of the image. The central portion of the images is completely replaced by a pixel value $0$ or completely black pixels, as intuitively shown in Figure~\ref{fig:complexity_aug}. 

For Section~\ref{res_complexity} in hypothesis 4, after trying multiple values of $x$ between 112 and 220, we selected $x=200$ because it was the minimum value that pushed the accuracy of the network to the lowest side that we intended. A detailed description of the CenterBlack image corruption method is given in~\ref{centerblack}. The CenterBlack is applied to a fraction of training images according to a range of probabilities provided through the 'image-noise-train-fraction' hyperparameter. The 'image-noise-train-fraction' here basically refers to the fraction of training samples that have the CenterBlack augmentation. 

Likewise, when QuarterBlack is mentioned in hypothesis 4, we introduce a random QuarterBlack augmentation, where we assign black patches sized ($112, 112$) to a random quarter \sm{of area} of images. An instance of QuarterBlack augmentation is shown in Figure~\ref{fig:complexity_aug}. The QuarterBlack is also applied randomly according to \sm{probabilities} provided through the 'image-noise-train-fraction' hyperparameter. The 'image-noise-train-fraction' here also refers to the fraction of training samples that have the QuarterBlack augmentation method applied.

\chapter{Results and Discussion}


\label{chap:results}
We run multiple experiments to evaluate the performance differences of Transfer Learning only and with Knowledge Distillation against various research questions in this section. We summarize the results after testing the hypotheses corresponding to these research questions and evaluate whether and how TL+KD is better than transfer learning below.

\section{Does TL+KD provide better validation performances than TL-only on the CIFAR10 image classification?}
\label{res:improves}
We test this hypothesis by comparing transfer learning only and transfer learning with knowledge distillation models for image classification in terms of accuracy, precision, recall, F1 score, confusion matrix, and SHAP interpretations. This is the main hypothesis of this thesis and provides multiple pieces of evidence to prove it. We evaluate the performance by comparing predictions from the same Student network with the validation ground truth after training. The experimental setup to test this hypothesis is explained in Section~\ref{sec:ex_training}.

\subsection{In terms of Evaluation Metrics}
The  mean final accuracy and other scores for quantitative comparison of the experiments are summarized in Table \ref{tab:results_quant}. We report the average values and their standard deviation on three runs with different random seeds $(0, 7, 42)$. 

\begin{table}[!htbp]
\centering
\begin{tabular}{||c | c c | c||} 
 \hline
  & TL Only($\mu_{\sigma}$) & TL with KD($\mu_{\sigma}$) & Mean Improvements\\ [0.5ex] 
 \hline\hline
Accuracy & $93.46_{0.91}$ &  $97.02_{1.24}$ & $+3.56$\\
 Precision & $93.57_{1.08}$ & $96.91_{1.62}$ & $+3.34$\\
 Recall & $92.99_{0.86}$ & $96.46_{1.40}$ & $+3.47$\\
 F1 Score & $93.18_{1.12}$ & $96.62_{1.17}$ & $+3.44$\\
 \hline
\end{tabular}
\caption[Evaluation Metrics performance on TL variants]{Evaluation metrics performance on TL variants. The base represents the mean, whereas the subscript represents the standard deviation between three runs. We see consistent improvements in all four evaluation metrics for TL with KD compared to TL alone.}
\label{tab:results_quant}
\end{table}

The first column describes the types of evaluation metrics described in detail in Section~\ref{ex_metrics}. In addition to the accuracy score to calculate the correctness, we also evaluate the precision, recall, and F1 scores to verify that the model is not biased against certain positives/negatives and to support the insights provided by accuracy. The second column of Table~\ref{tab:results_quant} shows the corresponding evaluation metrics scores with the vanilla transfer learning setup. 
The third column provides the performance after augmenting the same student network with Knowledge Distillation. The second and third columns report the mean value of the metrics as a base and the standard deviation across multiple runs as a subscript. The last column summarizes the improvements in mean values obtained with Knowledge Distillation, where a positive value indicates that the Student Network performed better with KD.  Because we are dealing with a balanced CIFAR10 dataset, all evaluation metrics like precision, recall along with the F1 score are consistent and behaved similarly throughout our experiments with accuracy. From Table~\ref{tab:results_quant}, we observe that TL+KD training performs better than the baseline ResNet model with a ViT teacher with more than 3\% in terms of the four evaluation metrics, i.e., accuracy, precision, recall, and F1 score. Standard deviation scores are significantly higher for TL+KD compared to the counterpart TL Only. This shows, along with improving the correctness of the model, that the soft labels also introduce slightly more randomness into the training. Looking at the \sm{mean improvement columns, we see that the accuracy improvements are the highest compared to other metrics, and the precision score is slightly lower than the recall scores. However, by observing the standard deviation of both models, the differences do not seem significant and could change for different 3 runs.} 

\subsection{In terms of Confusion Matrix}
\begin{figure}[!htbp]
\centering
\begin{subfigure}{.57\textwidth}
  \centering
  \includegraphics[width=0.90\linewidth]{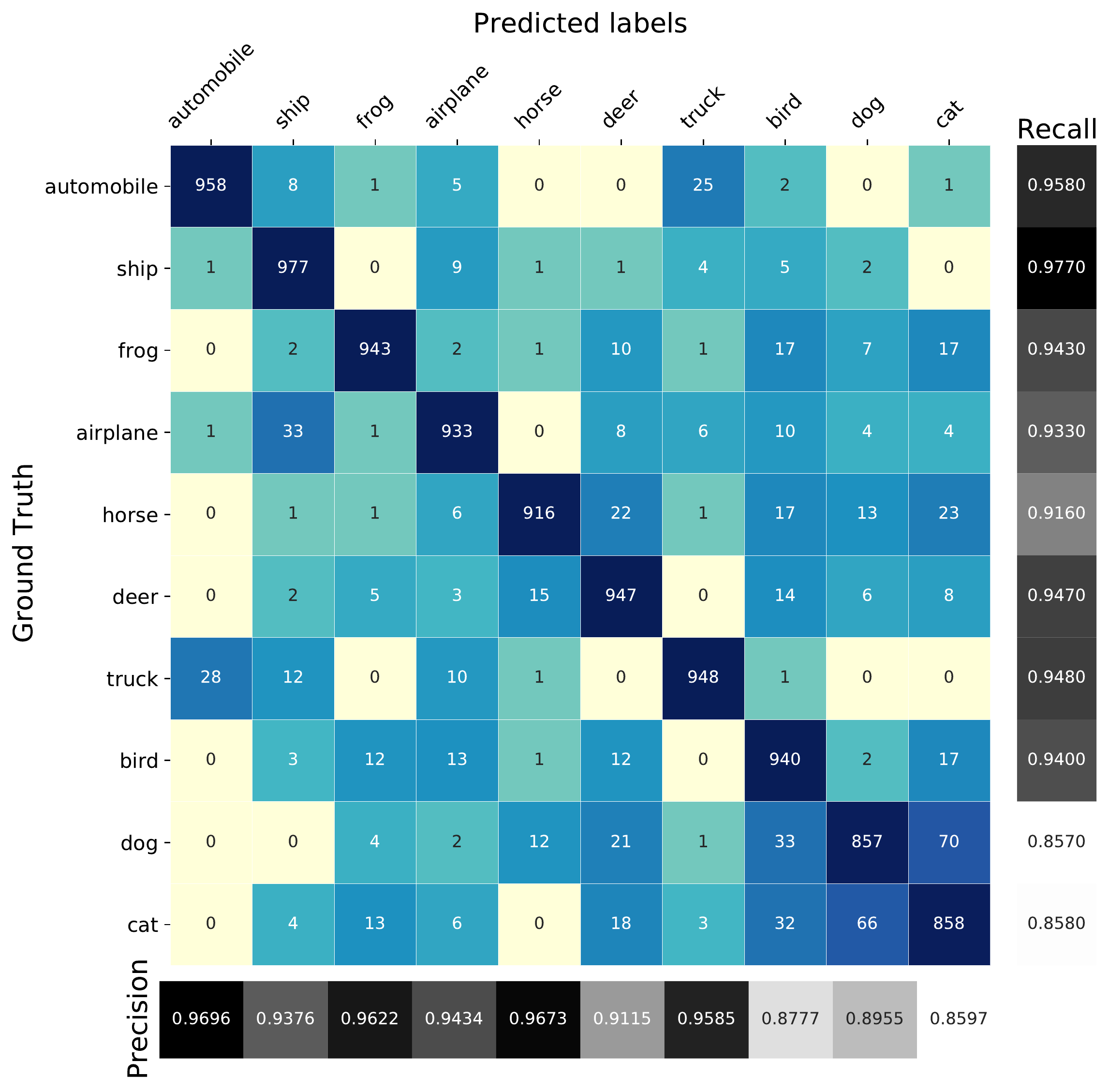}
  \caption{TL Only}
  \label{fig:sub12}
\end{subfigure}%
\vspace{0pt}
\begin{subfigure}{.57\textwidth}
  \centering
  \includegraphics[width=0.90\linewidth]{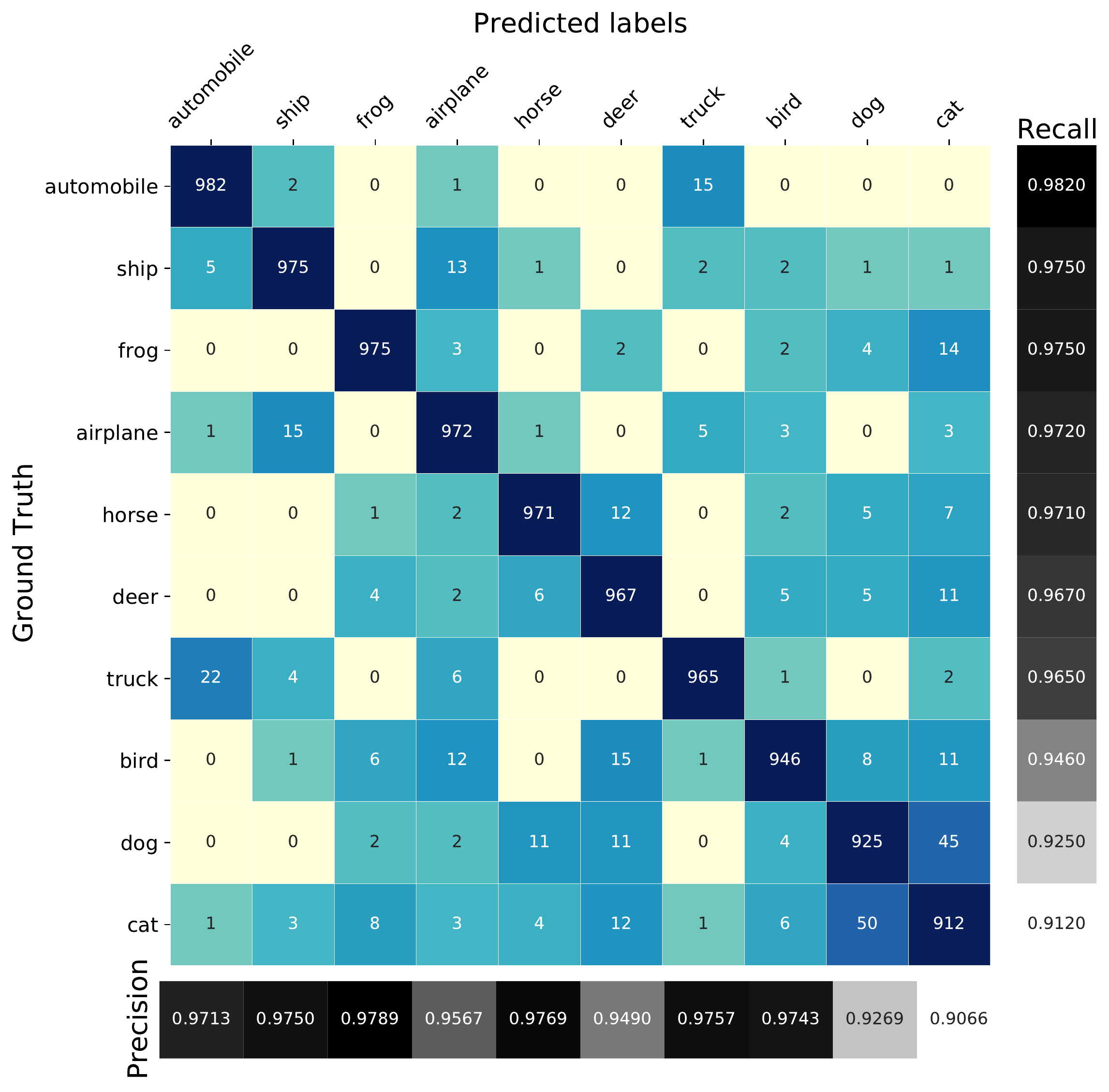}
  \caption{TL+KD}
  \label{fig:sub22}
\end{subfigure}
\caption[Comparison of confusion matrix between TL only and TL+KD]{Comparison of the confusion matrix between TL Only and TL+KD. Darker shades refer to high values, and lighter shades refer to lower values. Color scheme across two images is not necessarily in the same intensity. Classes are sorted according to the recall score of TL + KD where diagonal values are in decreasing order. Non-diagonal incorrect predictions improved and moved towards diagonal on TL+KD. Precision values for all classes and Recall values for all classes except Ship improved as shown in Table~\ref{tab:cmatrix_changes}.}
\label{fig:conf_matrix}
\end{figure}

Apart from calculating evaluation metrics, we also generate a confusion matrix of the models in both sets of figures in Figure \ref{fig:conf_matrix}. The confusion matrix on the top is for the ResNet Student network with vanilla Transfer Learning, whereas the bottom one is TL+KD applied using ViT as a teacher. The experimental setup is the same as the setup used to generate Table~\ref{tab:results_quant} but is a different run. 

In the confusion matrix, rows are the actual images from the CIFAR10 validation set for each class, while columns are the predicted classes. Each class has $1k$ images to evaluate, and each of the cells has a color to choose from, a gradient that goes from dark to light for both precision \& recall and the actual count of images. The dark color gradient in any cell outside the diagonal represents the actual class of the row that does not match the predicted class of the columns. Since the classifier has a validation accuracy of $92+$, most cells are diagonally dark, representing most of the actual positive classes that were correctly predicted to be positive. Here, the more prominently dark the diagonal cells, the better the classification model. 

\begin{table}[!htb]
\centering
\begin{tabular}{|l|r|r|r|r|r|}
\toprule
Class &TL only ($TP_1$) &TL+KD ($TP_2$) &$TP_2 - TP_1$ &$\frac{TP_2 - TP_1}{N-TP_1}$ \\\midrule
Automobile &958 &982 &24 &57.14\% \\
Ship &977 &975 &-2 &-8.7\% \\
Frog &943 &975 &32 &56.14\% \\
Airplane &933 &972 &39 &58.21\% \\
Horse &916 &971 &55 &65.48\% \\
Deer &947 &967 &20 &37.74\% \\
Truck &948 &965 &17 &32.69\% \\
Bird &940 &946 &6 &10\% \\
Dog &857 &925 &68 &47.55\% \\
Cat &858 &912 &54 &38.03\% \\
\midrule
Mean &927.7 &959 &31.3 &39.42\% \\
\bottomrule
\end{tabular}
\caption[Changes in true positives for TL only and TL+KD models.]{Changes in true positives (diagonal values) for TL only and TL+KD models. $TP_1$ column represents True Positives or diagonal values for TL Only for all classes. $TP_2$ represents the same for the TL+KD model. The fourth column shows the improvements in true positives after KD. The last column shows the \sm{class-wise error decrement percentage for TL only model after KD is added. The denominator in the last column is obtained by subtracting TL only the true positives from the total number of samples ($N=1000$) for each class. We calculate this denominator to compare changes in errors for each class}. We see a consistent decline in error for all classes except Ship.}\label{tab:cmatrix_changes}
\end{table}

In inspection, we can see that the diagonal cells for the TL+KD model are slightly better or homogeneous compared to vanilla transfer learning, with each corresponding cell having values higher in TL+KD. In vanilla transfer learning, there are a few prominent cases outside the diagonal cells, like the actual cat being predicted as the dog and the actual truck being recognized as the automobile. There are also some cases where the actual frog is recognized as a cat, and the actual horses are recognized as deer and cat. Similarly, there are also some cases in TL+KD where actual dogs are being confused with cats, but these cases, along with others, do not seem as prevalent as the vanilla TL model. Most of the wrong predictions, as shown in the confusion matrix, seem to be either because of visual similarities of classes like Truck and Automobile or environmental(background) similarities that classes like Cat and Dog have. In TL+KD, confusion is now more with intuitively related concepts rather than associated things, like ships now being confused with automobiles and airplanes rather than birds. With this observation, we can infer that TL+KD helps vanilla TL work better in cases where the model gets confused with visually similar classes. We \sm{guess that such a thing happens because a temperature scaled softmax model in TL+KD} learns to focus on more robust features and avoid focusing on less relevant features like background. This hypothesis is further explored in another Section~\ref{res_segmention}.

\subsection{In terms of Interpretability}

\begin{figure}[!htbp]
\centering
\includegraphics[width=0.86\linewidth]{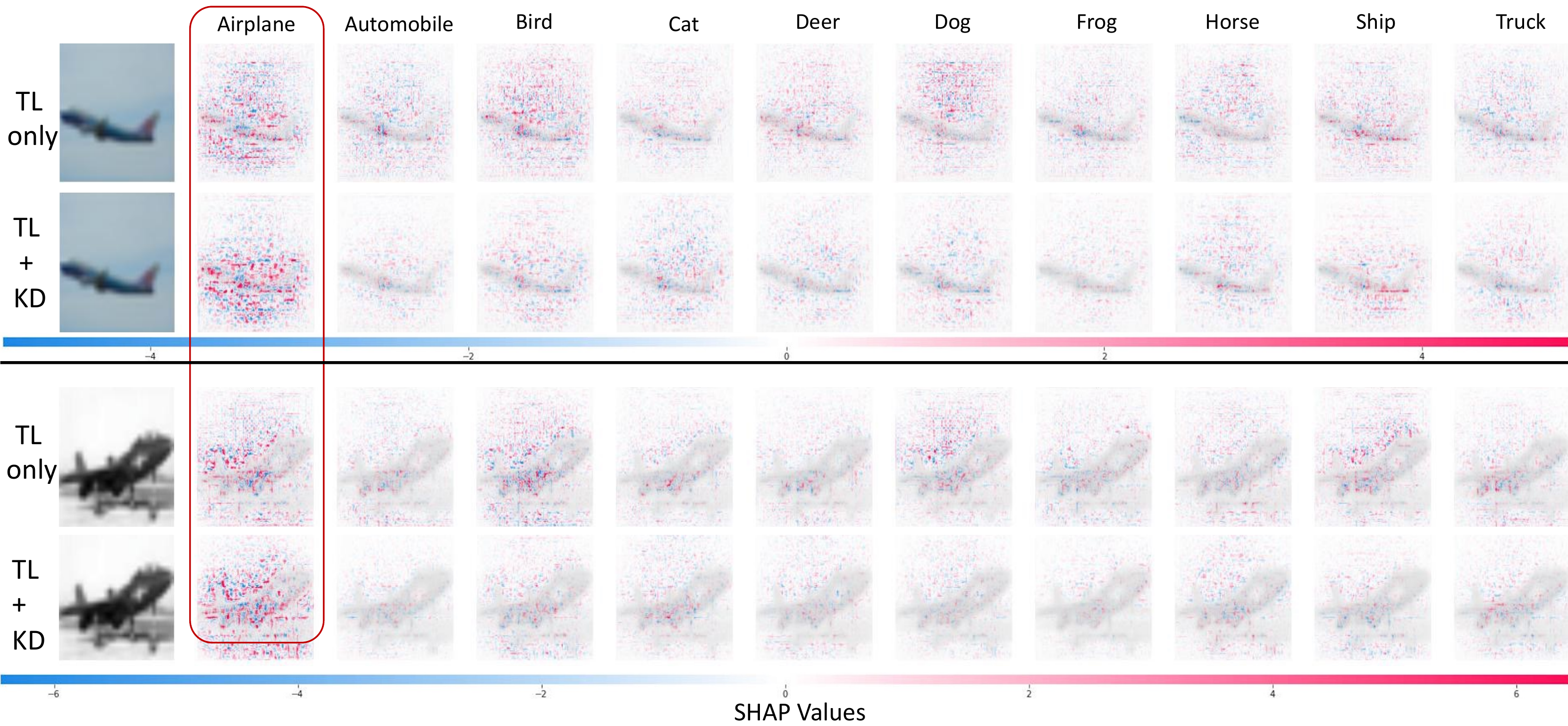}
\includegraphics[width=0.86\linewidth]{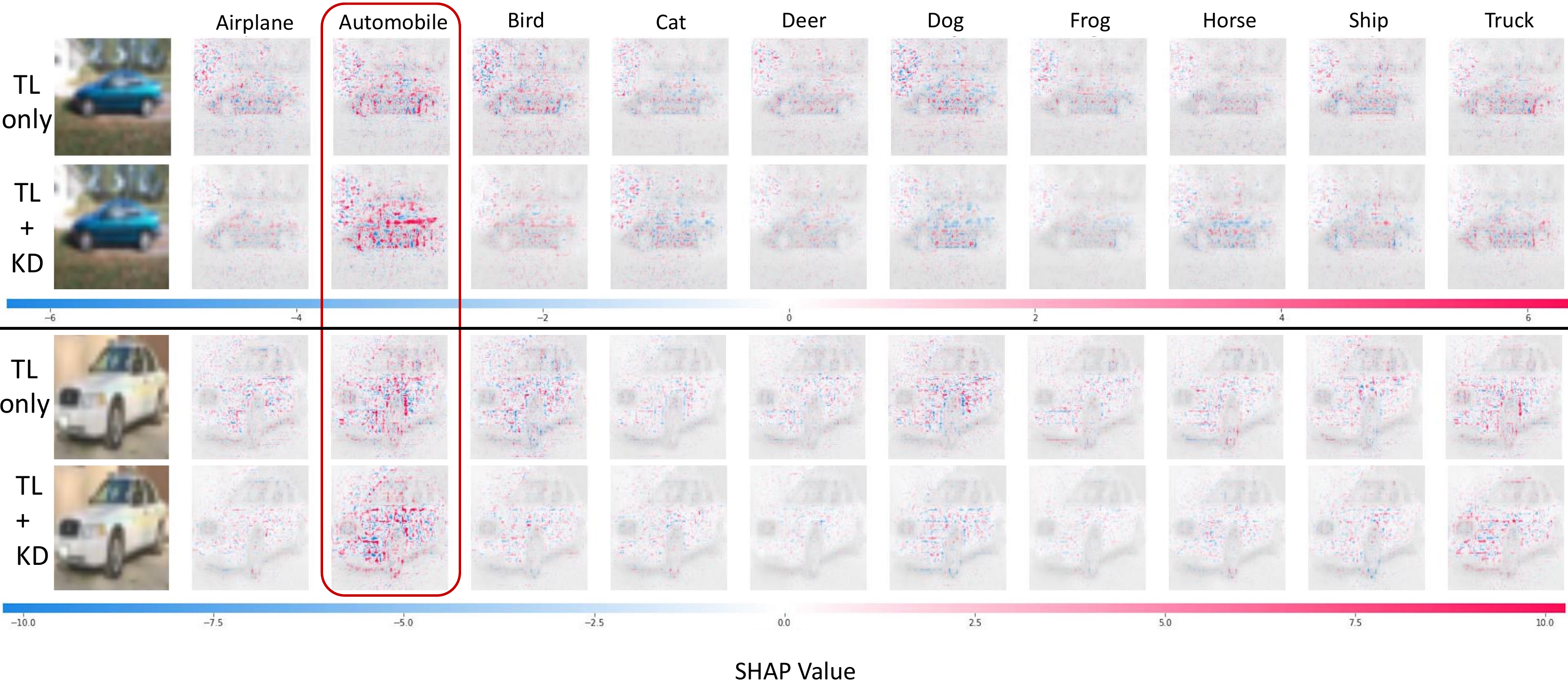}
\includegraphics[width=0.86\linewidth]{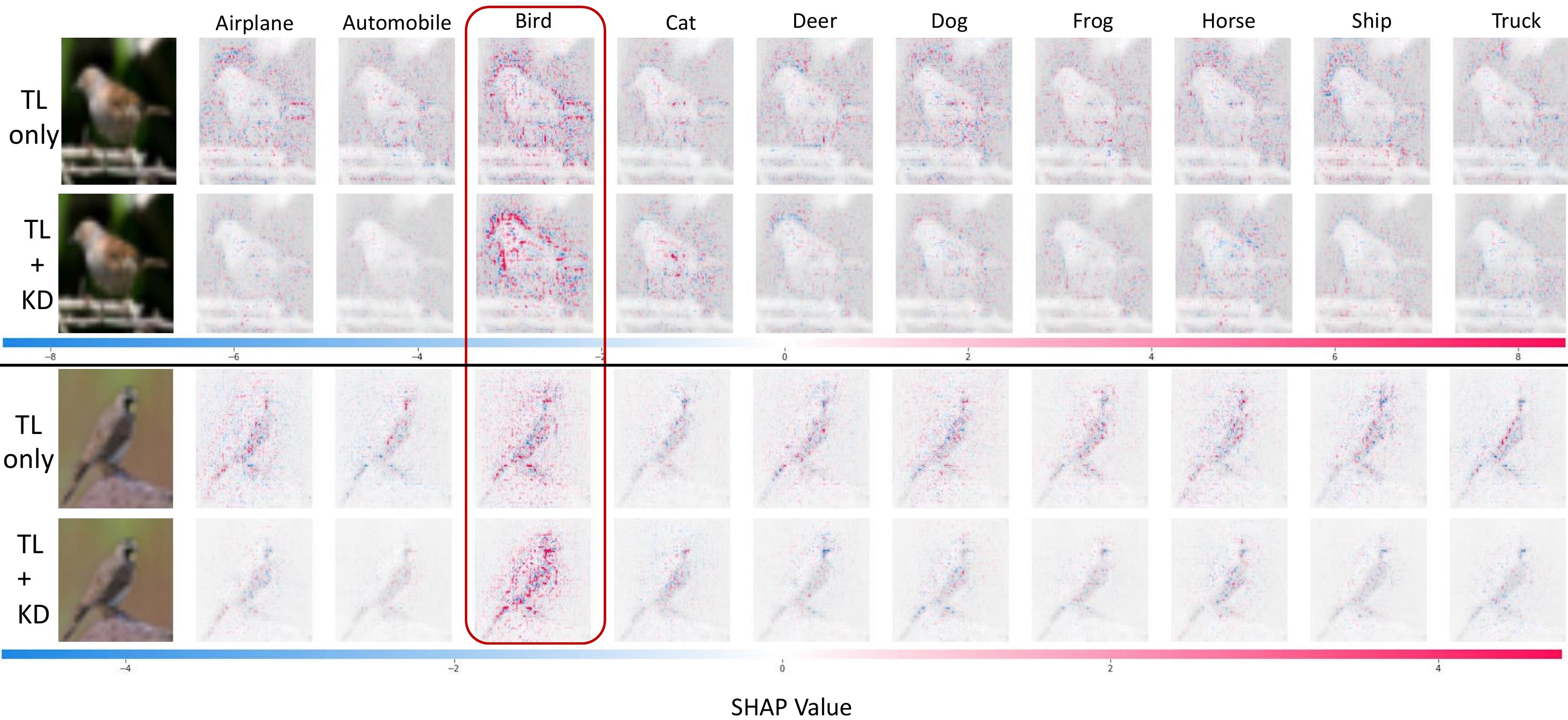}
\caption[SHAP contribution for test images across all classes]{SHAP contribution for two test images per class(airplane, automobile, bird) with and without KD. The winning classes are marked with a red box. Positive contributions are colored red, whereas negative contributions are colored blue. The upper row is the image and contribution for all classes for TL only, whereas the lower row is the corresponding TL+KD variant. The KD winner class appears to have the brightest red colors compared to TL only. SHAP values range across samples are not necessarily in the same scale.}\label{fig:shap_all_3}
\end{figure}
\begin{figure}[!htbp]
\centering
\includegraphics[width=0.89\linewidth]{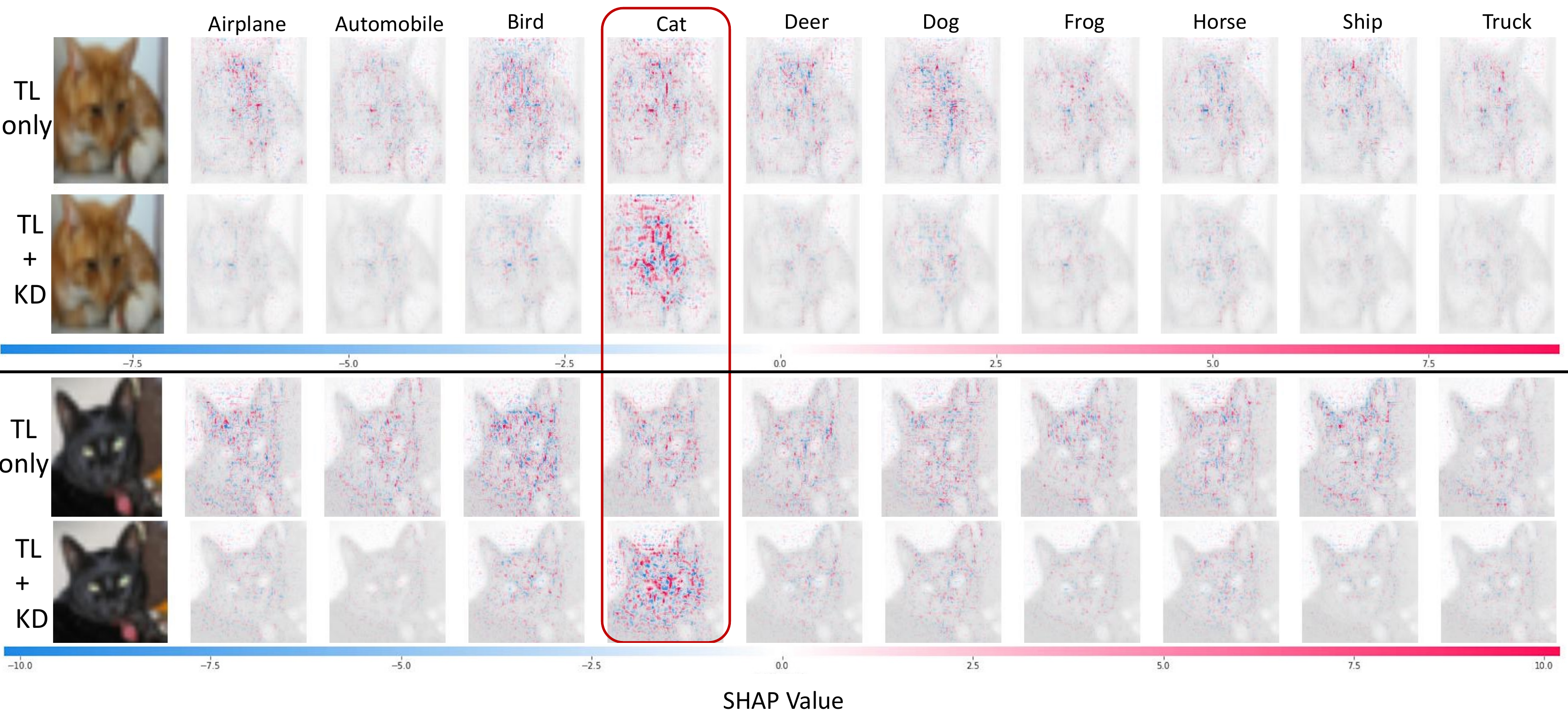}
\includegraphics[width=0.89\linewidth]{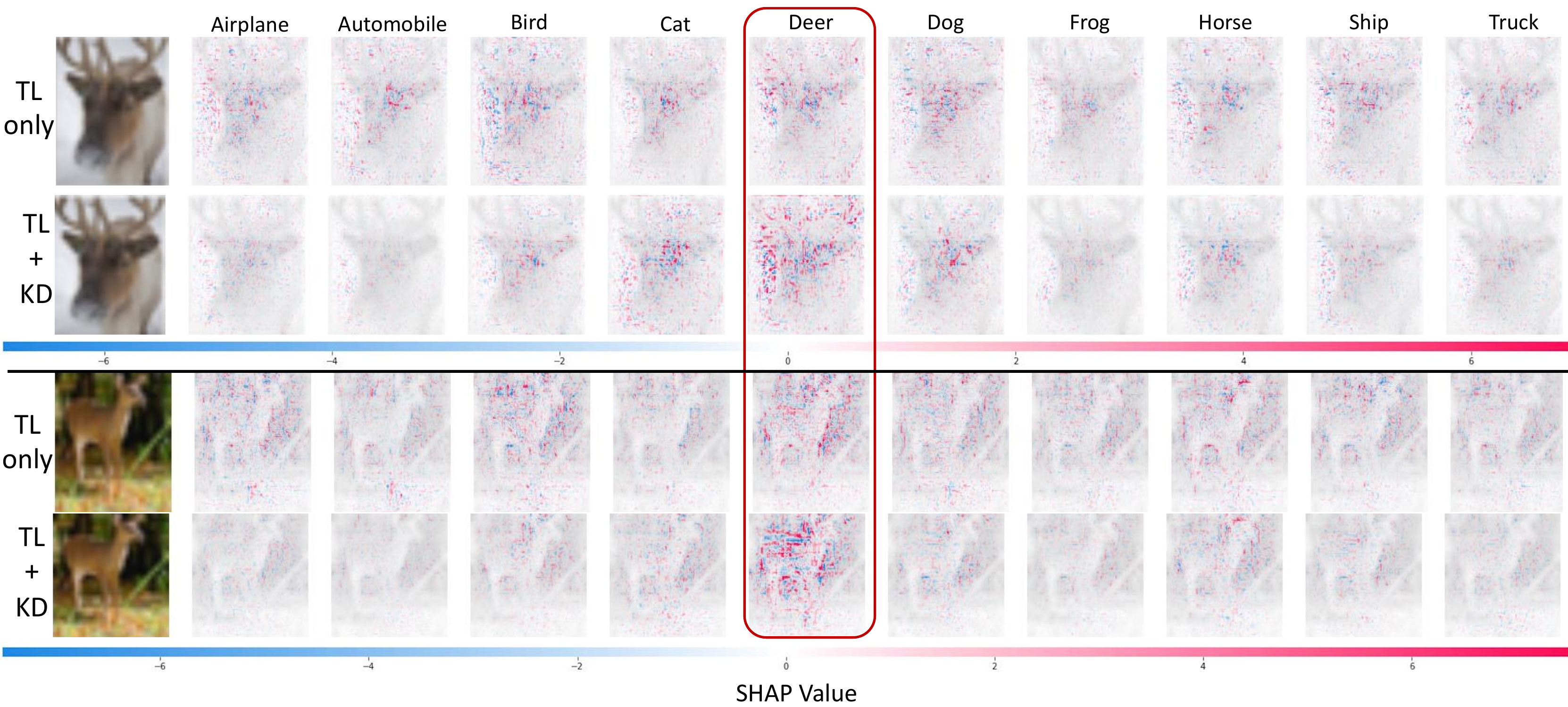}
\includegraphics[width=0.89\linewidth]{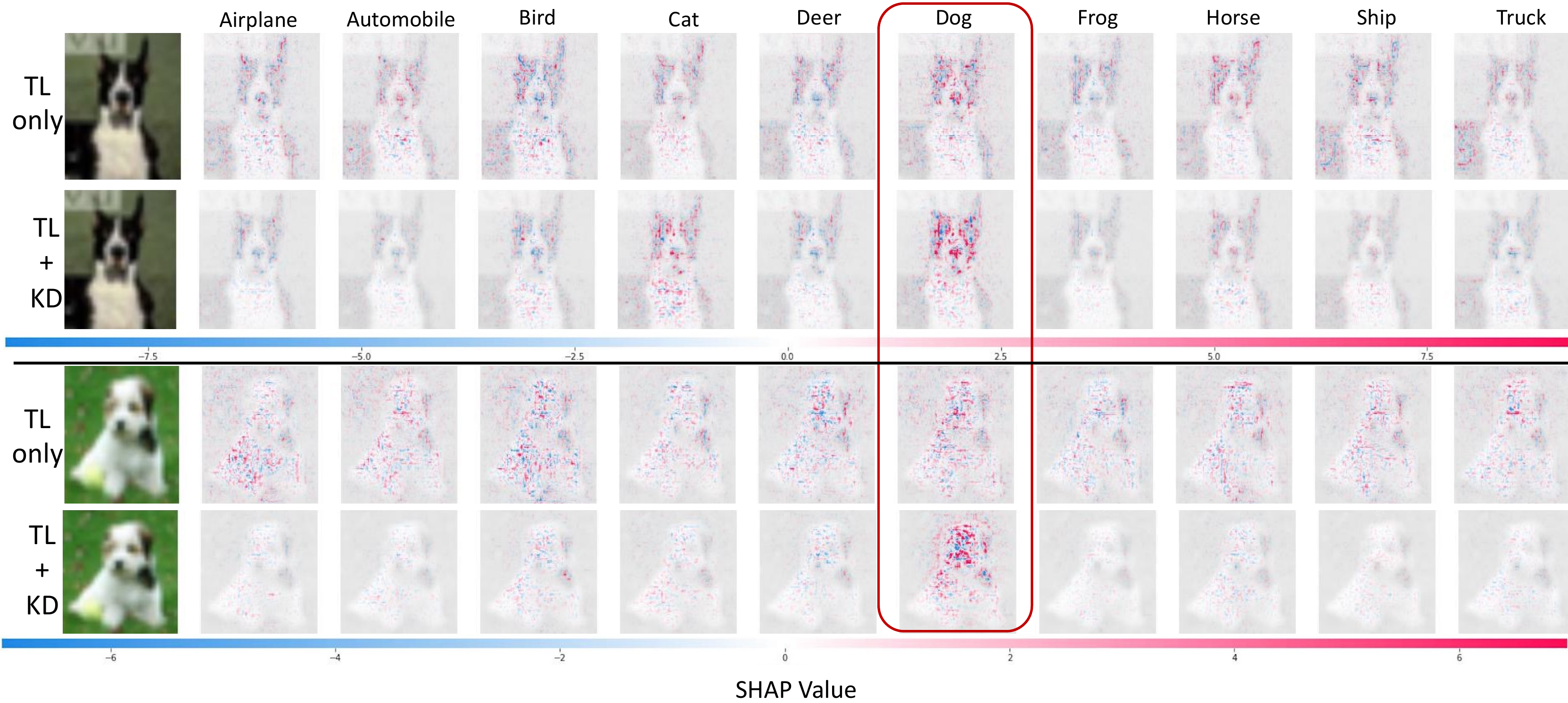}
\caption[]{SHAP contribution for two test images per class(cat, deer, dog) with and without KD. Winning classes are marked with a red box. The top row is the image and contribution for all classes for TL only, whereas the bottom row is the corresponding TL+KD variant. The KD winning class appears to have the brightest red colors compared to TL only. SHAP values range across samples are not necessarily in the same scale.}\label{fig:shap_all_6}
\end{figure}
\begin{figure}[!htbp]
\centering
\includegraphics[width=0.88\linewidth]{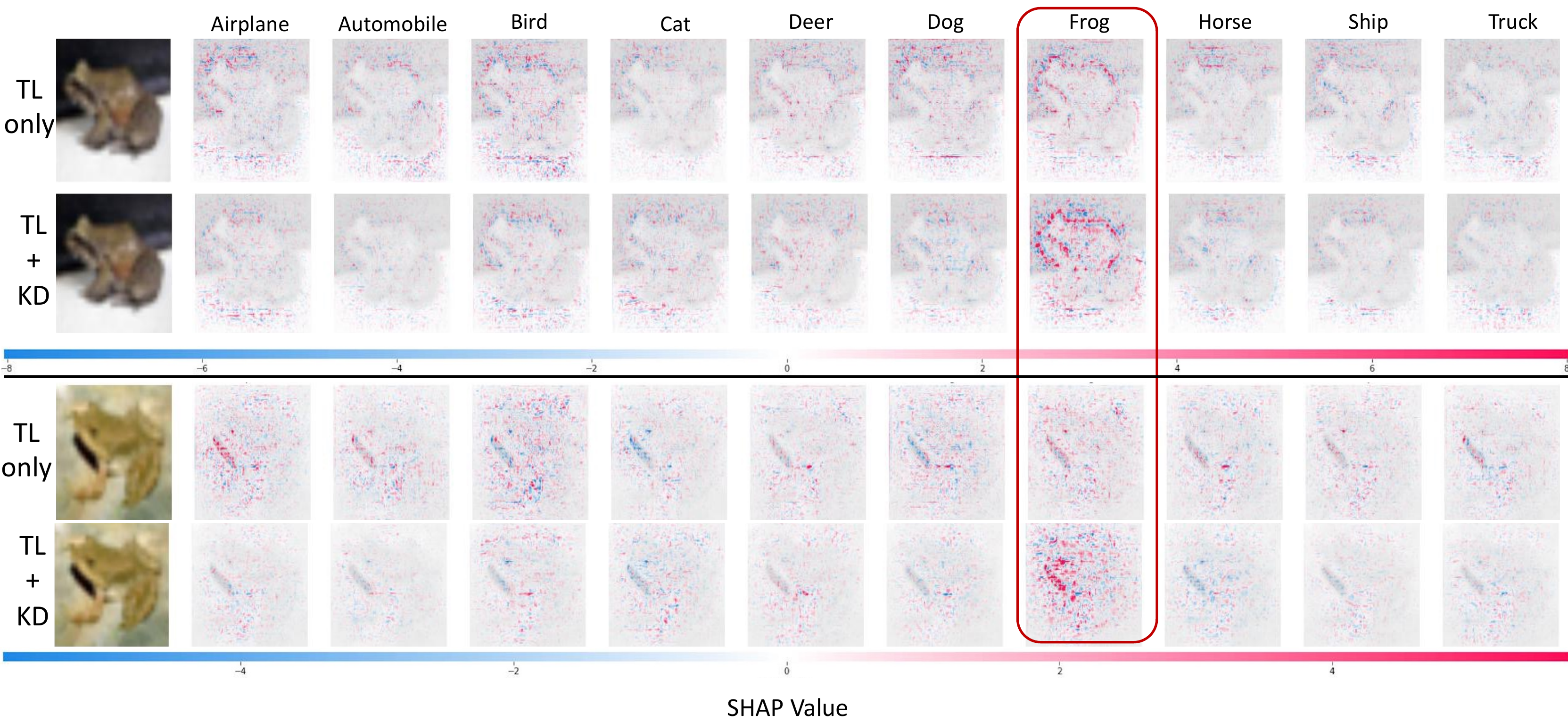}
\includegraphics[width=0.88\linewidth]{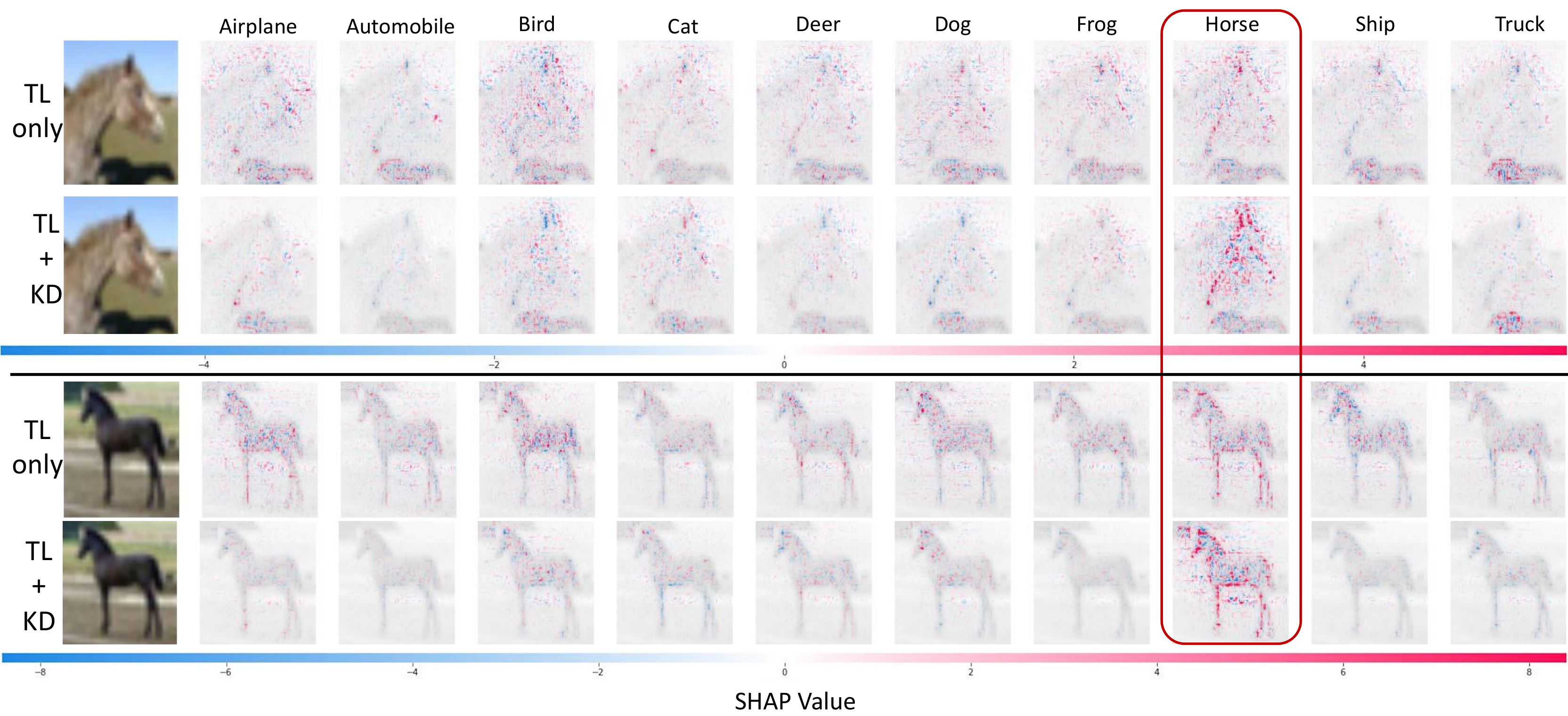}
\includegraphics[width=0.88\linewidth]{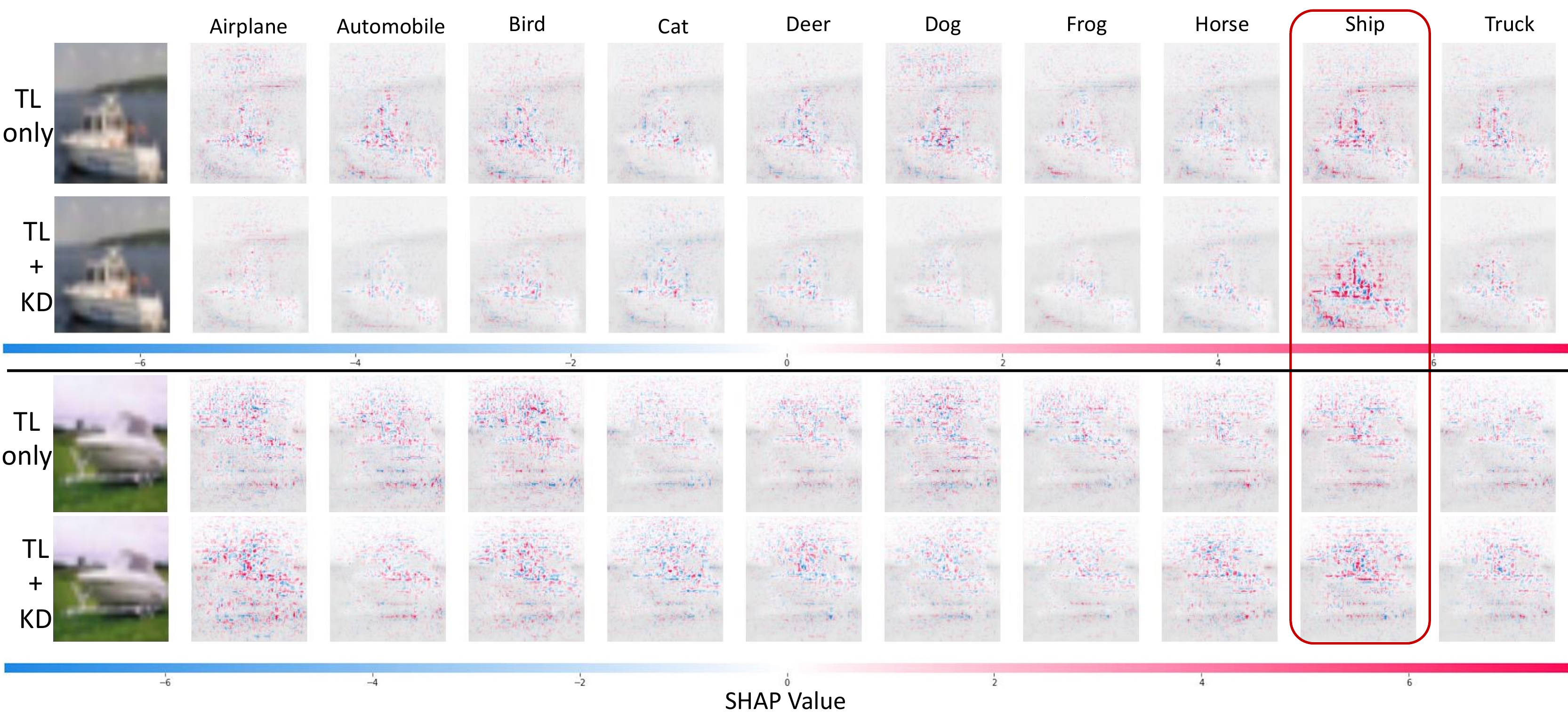}
\caption[]{SHAP contribution for two test images per class(frog, horse, ship, truck) with and without KD. Winning classes are marked with a red box. The top row is the image and contribution for all classes for TL only, whereas the bottom row is the corresponding TL+KD variant. The KD winning class seems to have the brightest red colors compared to TL only. SHAP values range across samples are not necessarily in the same scale.}\label{fig:shap_all_9}
\end{figure}
\begin{figure}[!htbp]
\centering
\includegraphics[width=\linewidth]{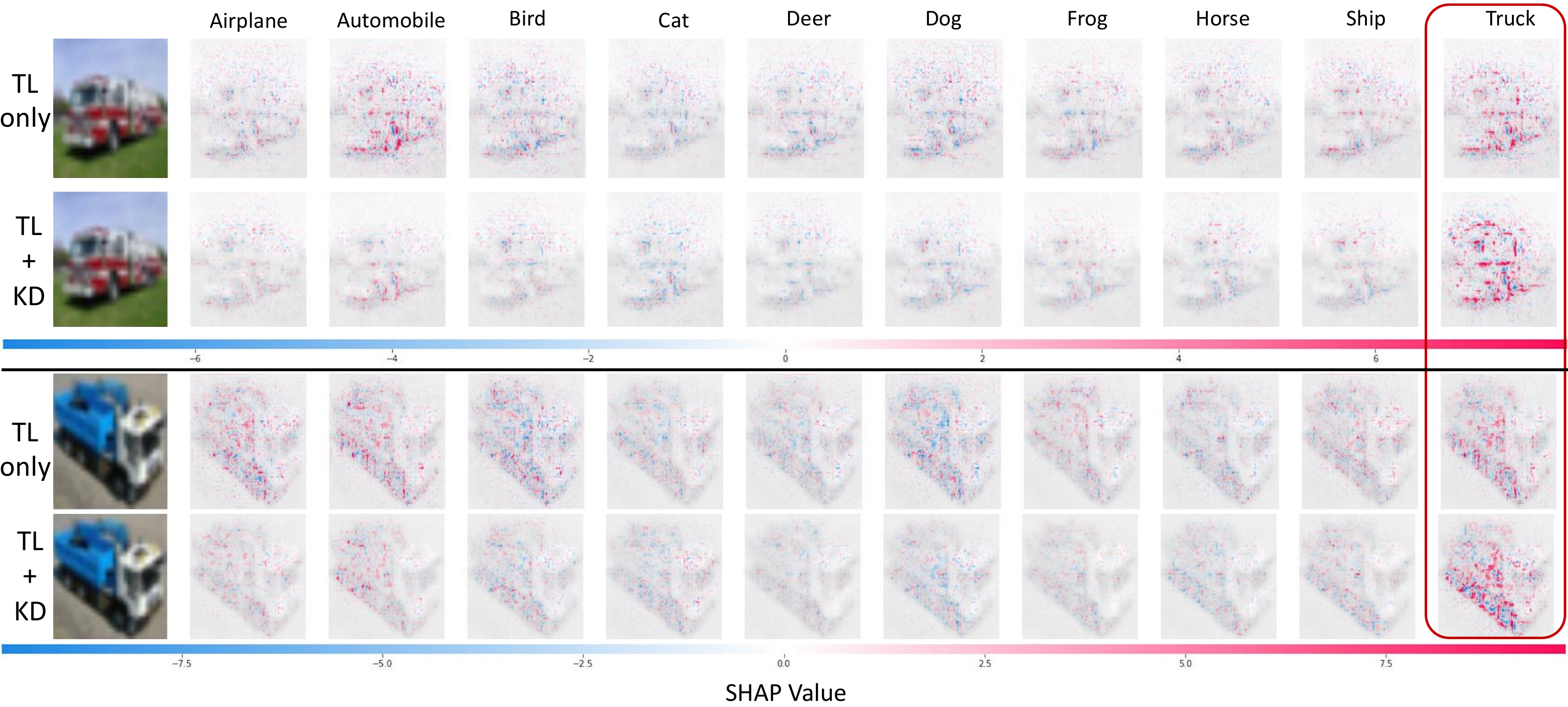}
\caption[]{SHAP contribution for two test images per truck class with and without KD. The winning classes are marked with a red box. The upper row is the image and contribution for all classes for TL only, whereas the lower row is the corresponding TL+KD variant. The KD winning class seems to have the brightest red colors compared to TL only. SHAP values range across samples are not necessarily in the same scale.}\label{fig:shap_all_10}
\end{figure}

For interpretation, we follow the process described in the Experiment Setup Chapter~\ref{ex_interpret}. We export multiple images with the explanations for both TL only and TL+KD models and compare the visual differences and similarities in those contribution plots. 
All figures~\ref{fig:shap_all_3} to ~\ref{fig:shap_all_10} in this section below show the Shapley values for each pixel, which represent their contribution of them to pushing the model prediction to be that particular class. The red color indicates a positive contribution, whereas the blue color represents a negative contribution. A detailed description of the absolute values and a further quantification of the interpretation are given in Section~\ref{res_segmention}. For now, we explore the visual cues and differences in such interpretations for all available classes of CIFAR10. The absolute SHAP value range varies depending on the input samples and the model.

In the figures~\ref{fig:shap_all_3} to ~\ref{fig:shap_all_10}, we can see that both variants of the model focus on distinctive regions for different classes.  The first common observation in all of the images is that the contributions to winning classes seem to be comparatively more prominent visually. We can also see that the patterns and edges of the image are more distinctive for the model TL + KD compared to the TL-only section in most of the examples.  More exploration and quantification of the positive and negative contributions to the winning classes carried out in Figure~\ref{fig:shap_all_one} also provides further evidence of this.

\label{res_interpret}

In summary, with the help of multiple evaluation metrics and a confusion matrix, it is evident that Knowledge Distillation improves the qualitative performance of the transfer learning model.

\newpage

\section{Does TL+KD provides more contribution on foreground image features when measured by SHAP contribution?}
\label{res_segmention}
\begin{figure}[!htb]
\centering
\includegraphics[width=\linewidth]{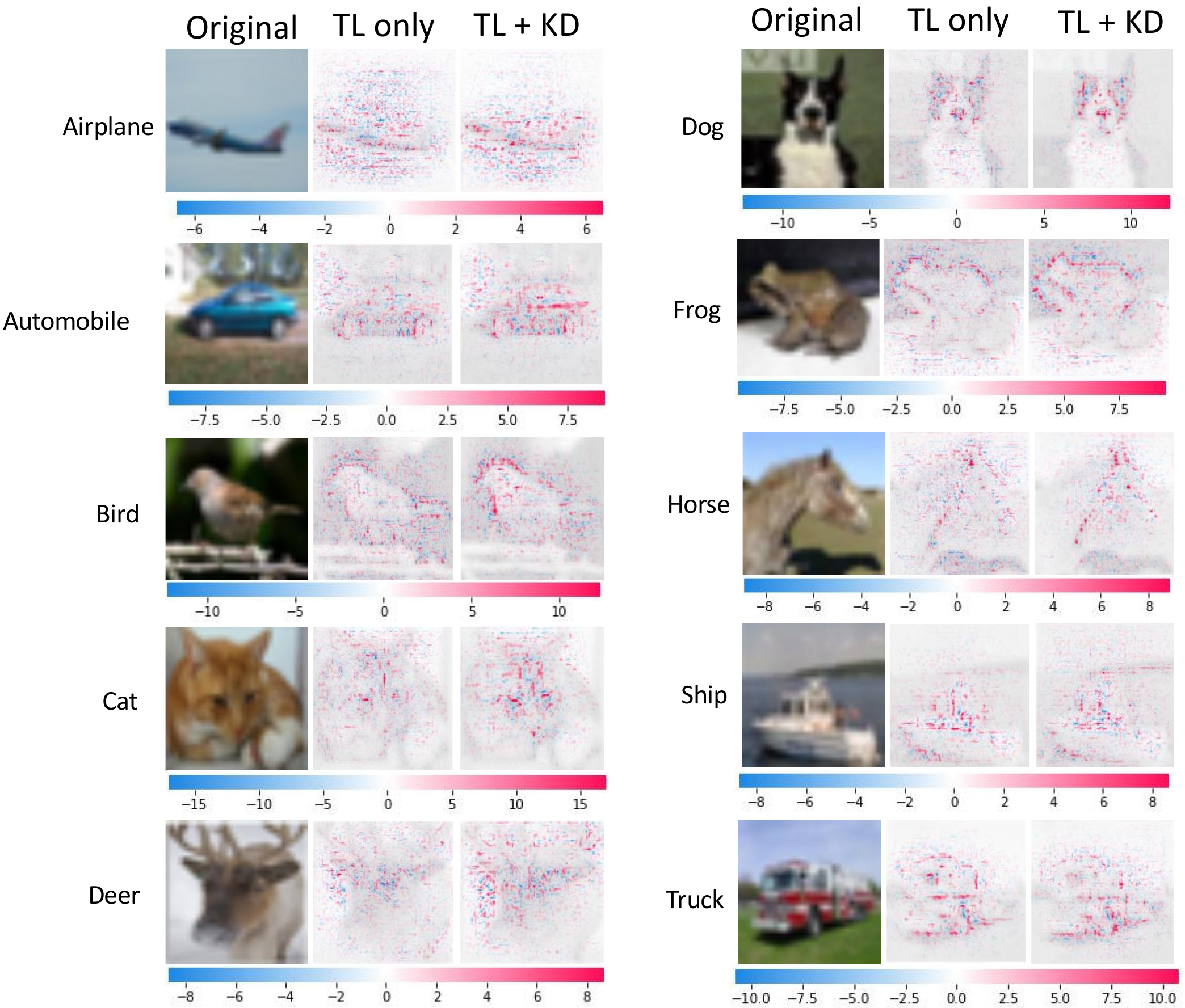}
\caption[Winning Class SHAP contribution for image samples for each class]{Winning class contribution for Image Samples for each class. The first column is the original image, the middle column is the SHAP contributions for the TL-only variant, and the last column shows the contributions for TL+KD.Visual inspection hints at higher red patches (positive contribution) in the foreground section of the image. SHAP values across samples are not necessarily on the same scale.}\label{fig:shap_all_one}
\end{figure}

In this section, we further explore interpretability in terms of positive and negative contributions of pixels of winning classes. Initially, we select the SHAP values of the winning classes for a selected sample from each class. The blown version of the selected SHAP interpretations from Section~\ref{res_interpret} is shown in Figure~\ref{fig:shap_all_one}. 

\begin{figure}[!htbp]
\centering
\begin{subfigure}{.849\textwidth}
  \centering
  \includegraphics[width=\linewidth]{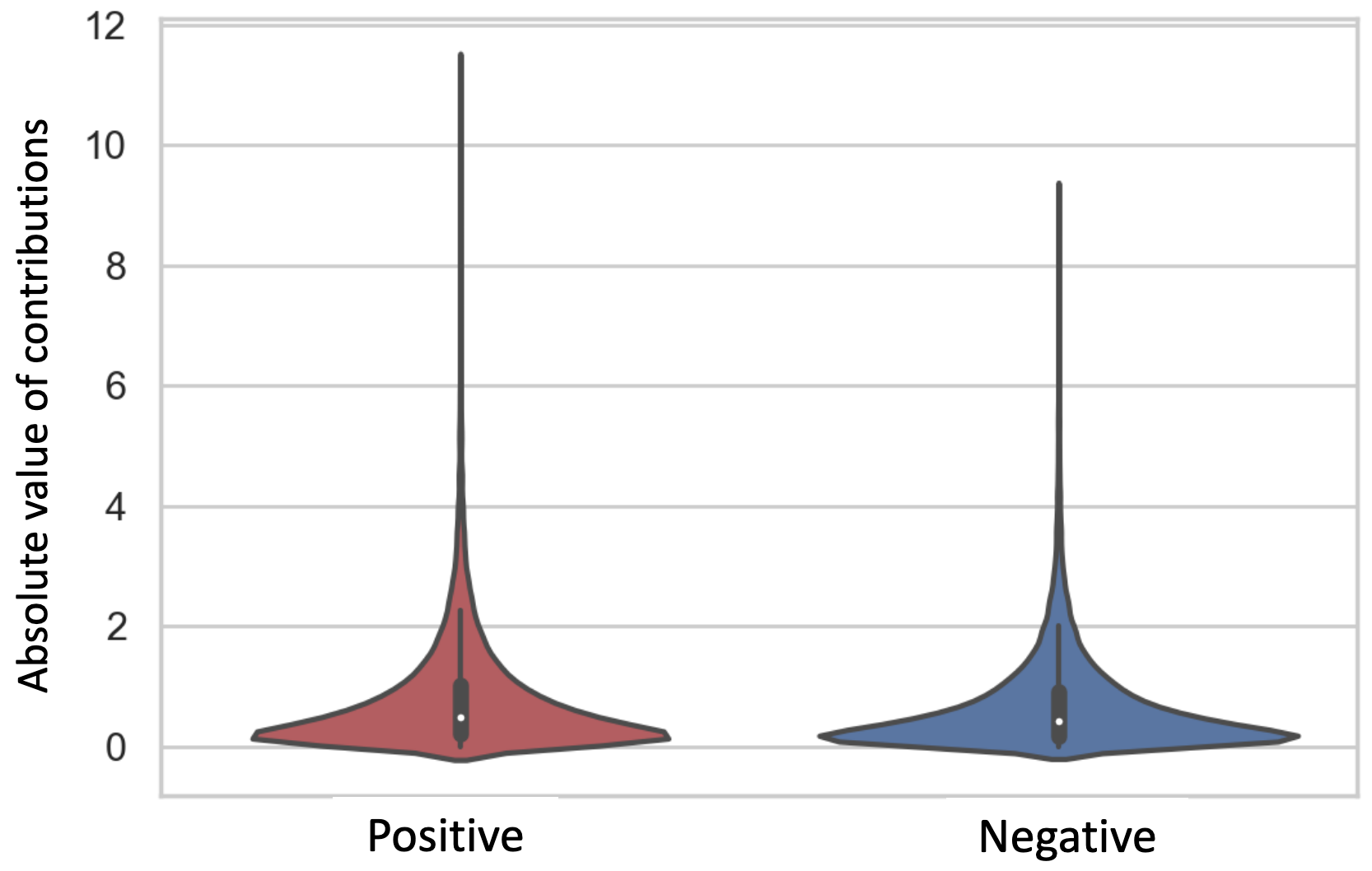}
  \caption{Foreground}
  \label{fig:sub11}
\end{subfigure}%
\vspace{0pt}
\begin{subfigure}{.849\textwidth}
  \centering
  \includegraphics[width=0.90\linewidth]{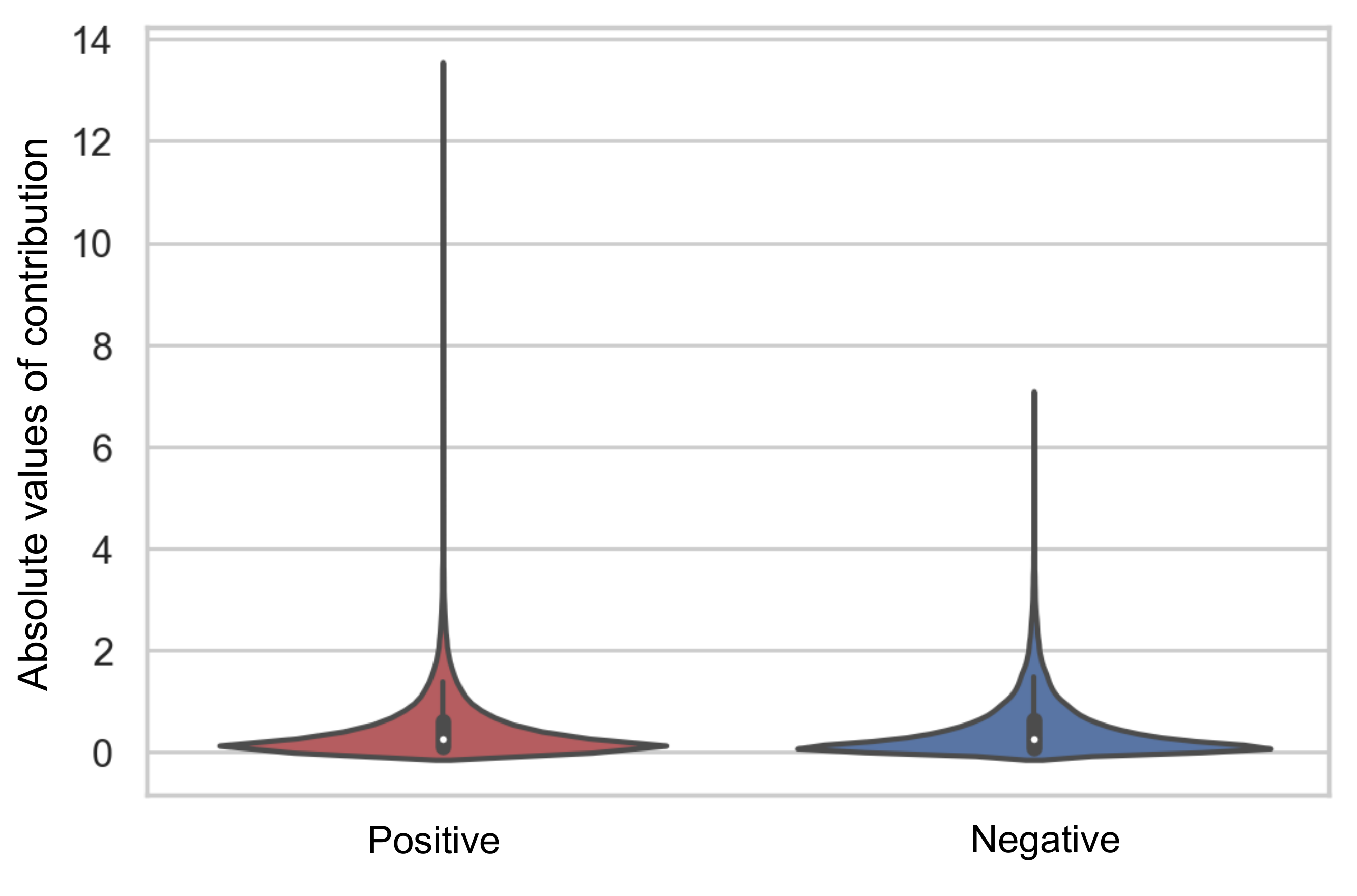}
  \caption{Background}
  \label{fig:sub21}
\end{subfigure}
\caption[Violin plot of absolute values of positive and negative contributions of the SHAP interpretations of a random sample from TL+KD model]{Violin plot of absolute values of the positive and negative contributions of the SHAP interpretations \sm{of a random sample from TL+KD model. a) Contributions for the foreground and b) contributions for the background. \FIX{This shows the comparison of absolute values rounded to the nearest integer value} for a random sample. The median for all these plots is very close to zero. Positive and negative violin plots are very similar in the foreground and background.} The summary statistics and density of both sides are larger/higher for positive than for negative contributions.}
\label{fig:shap_violin}
\end{figure}

Along with visual cues of differences in positive and negative contributions for the TL and TL+KD models, we also quantified the differences in positive and negative contributions numerically. First, to differentiate the foreground and background of the image, we find the segmentation map of all classes as described in Section~\ref{ex_interpret_quant}. Once we get the mask and can divide the image into foreground and background, we extract contributions of the winning class from the foreground and collect insights from it. Figure~\ref{fig:shap_violin} shows the violin plot for the absolute values of the positive and negative contributions. The violin plot shows the kernel density estimate that shows the peaks in the data along with the box plot with summary statistics. We can observe that the peaks of absolute values of positive contributions are comparatively higher than negative contributions. This indicates the reason why the class is selected as a prediction. 


\label{res_shap_table_explain}
We obtained the values of {\itshape pos,} {\itshape neg} and {\itshape diff} for foreground and background to compare and quantify the interpretation of SHAP between the TL+KD model and the TL-only model based on the setup explained in Section~\ref{ex_interpret_quant}.

Table~\ref{tab:results_shap_abcd} also shows the respective ratios calculated to better compare the contributions in different scenarios. The column $\frac{A}{B}$ shows the ratio of the difference in the positive and negative contributions of the TL models alone and TL+KD in the foreground. The column $\frac{C}{D}$ shows the ratio of the difference in the positive and negative contributions of the TL and TL+KD models in the background. The ratio $\frac{A/B}{C/D}$ shows the significance and comparison of the ratios in the previous two ratios $\frac{A}{B}$ and $\frac{C}{D}$ and gives insight into the ratio of foreground changes to background changes in TL+KD. The negative sign in the ratio $\frac{C}{D}$ and $\frac{A}{B}$ / $\frac{C}{D}$ indicates that the negative contributions in the background (outside pixels) are higher in one of the models and not the same in its counterpart. 


\begin{table}[!htbp]
\centering
\normalsize
\resizebox{\linewidth}{!}{%
\begin{tabular}{|l|c|c|c|c|c|c|c|c|c|c|c|c|c|c|c|} 
\hline 
\multicolumn{1}{|c|}{} & \multicolumn{6}{c|}{TL Only} &  \multicolumn{6}{c|}{TL+KD} & \multicolumn{3}{|c|}{Changes in Contribution} \\
\cline{2-16}
\multicolumn{1}{|c|}{Class} & \multicolumn{3}{c|}{foreground (inside)} & \multicolumn{3}{c|}{background (outside)} & \multicolumn{3}{c|}{foreground (inside)} & \multicolumn{3}{c|}{background (outside)}  & \multirow{2}{*}{$\frac{A}{B}$} & \multirow{2}{*}{$\frac{C}{D}$}	& \multirow{2}{*}{$\frac{A/B}{C/D}$} \\
\cline{2-13}
\multicolumn{1}{|c|}{} & \multicolumn{1}{c|}{pos} &  \multicolumn{1}{c|}{neg} &  \multicolumn{1}{c|}{diff$(A\uparrow)$} & \multicolumn{1}{c|}{pos} &  \multicolumn{1}{c|}{neg} &  \multicolumn{1}{c|}{diff$(C\downarrow)$} &\multicolumn{1}{c|}{pos} &  \multicolumn{1}{c|}{neg} &  \multicolumn{1}{c|}{diff$(B\uparrow)$} & \multicolumn{1}{c|}{pos} &  \multicolumn{1}{c|}{neg} &  \multicolumn{1}{c|}{diff$(D\downarrow)$}  &&&\\
\hline 
airplane   &                5532 &               -5248 &                  284 &                18587 &               -19403 &                  -816 &               5071 &              -3562 &                1509 &               12327 &              -12161 &                  166 &  0.19 & -4.92 &  -0.04 \\
automobile &               12010 &               -8869 &                 3141 &                13821 &               -14118 &                  -297 &              12100 &              -6698 &                5402 &                9726 &               -9336 &                  390 &  0.58 & -0.76 &  -0.76 \\
bird       &               12293 &              -10611 &                 1682 &                30261 &               -24876 &                  5385 &              10776 &              -7850 &                2926 &               21938 &              -16875 &                 5063 &  0.57 &  1.06 &   0.54 \\
cat        &               21465 &              -16568 &                 4897 &                 3138 &                -3111 &                    27 &              43021 &             -30318 &               12703 &                4802 &               -5177 &                 -375 &  0.39 & -0.07 &  -5.35 \\
deer       &               18192 &              -15582 &                 2610 &                 7269 &                -6650 &                   619 &              20361 &             -15231 &                5130 &                8670 &               -7758 &                  912 &  0.51 &  0.68 &   0.75 \\
dog        &               19277 &              -16120 &                 3157 &                 9059 &                -8018 &                  1041 &              27104 &             -17838 &                9266 &                6302 &               -5672 &                  630 &  0.34 &  1.65 &   0.21 \\
frog       &               11510 &               -8829 &                 2681 &                 8131 &                -7631 &                   500 &              13361 &              -9659 &                3702 &                9657 &               -7905 &                 1752 &  0.72 &  0.29 &   2.54 \\
horse      &               11404 &               -9848 &                 1556 &                 8003 &                -7792 &                   211 &              14054 &             -11173 &                2881 &                8272 &               -7476 &                  796 &  0.54 &  0.27 &   2.04 \\
ship       &               14029 &              -11110 &                 2919 &                10825 &                -8518 &                  2307 &              15519 &             -10471 &                5048 &                7560 &               -5719 &                 1841 &  0.58 &  1.25 &   0.46 \\
truck      &               17514 &              -12479 &                 5035 &                10968 &                -9594 &                  1374 &              17247 &             -11066 &                6181 &                9585 &               -7585 &                 2000 &  0.81 &  0.69 &   1.19 \\
\hline
\end{tabular}
}
\caption[SHAP contribution analysis for each winning class sample]{SHAP contribution analysis for each class. \FIX{One random sample was selected for each class and the SHAP contribution for winning class prediction was used for this analysis. All values are rounded to the nearest integer}. \textit{diff} represents the difference in absolute values of the corresponding {\itshape pos} and {\itshape neg} of the same side. The upward arrow represents that the higher difference is better for the winning class. Changes in contribution are the ratio of differences($A,B,C \& D$) from earlier columns. The negative sign in the ratio $\frac{C}{D}$ represents the cases of classes where the negative contributions in the background were higher and different from another model. The major observations from this table are listed in Section~\ref{observations_interpret_quant_tab}.}
\label{tab:results_shap_abcd}
\end{table}

\begin{figure}[!htb]
\centering
\includegraphics[width=\linewidth]{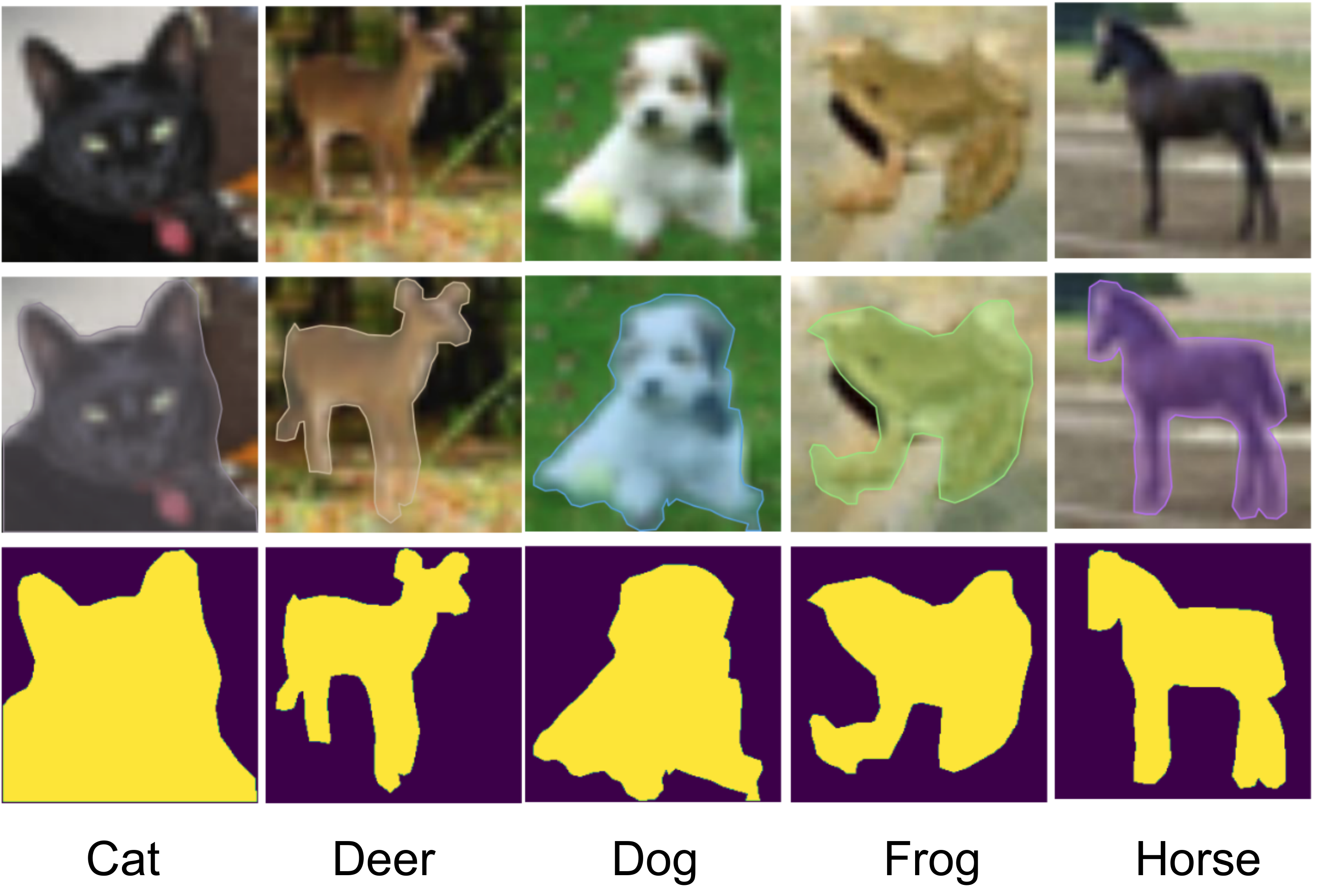}
\caption[Three-step image segmentation Process for additional five samples from ambiguous classes]{Three-step image segmentation process as in Figure~\ref{fig:segmentation_process} for an additional five samples from ambiguous classes.}
\label{fig:segmentation_process2}
\end{figure}

After looking at the scores {\itshape pos}, {\itshape neg} and {\itshape diff} and the ratios of {\itshape diff}, for the four variants of the foreground-background combination for TL and TL+KD, we find the following observations in Table~\ref{tab:results_shap_abcd}.
\FIX{
\begin{enumerate}
\label{observations_interpret_quant_tab}
    \item Absolute values of positive contribution({\itshape pos}) are always higher than the absolute value of negative contributions({\itshape neg}) for the foreground. \textit{diff(A) \& diff(B)} are always positive across the scenarios in both models for these samples. The possibility of this happening for all the cases that we tested by random chance is negligible: a paired Wilcoxon test gives the highly significant p=0.001953 in all comparisons.
    \item \textit{diff($B$)} is always greater than \textit{diff($A$)} as shown by the ratio \textit{diff}$(A)$/\textit{diff}$(B)$ being less than $1$ in all class samples. Comparison of \textit{diff($A$)} and \textit{diff($B$)} gives P-value=0.001953 by paired Wilcoxon test which suggests it is not likely to be due to random choice. This shows the difference between positive and negative contributions for the foreground increased in TL+KD.
    \item No significant difference was found between positive(\textit{pos}) and negative(\textit{neg}) contributions for the background when compared across TL and TL+KD. Comparison of \textit{diff($C$)} and \textit{diff($D$)} gives P-value=0.1934 by paired Wilcox test and is not significant.
    \item TL+KD learns to reject the background more in all cases, except for the frog, cat, deer, and horse classes. {\itshape pos} absolute value of the background or outside segmentation map is lower for TL+KD compared to the TL-only counterpart for six classes. Six out of ten comparisons giving the same sign are reasonably consistent with random fluctuations: the p-value by a paired Wilcoxon test is 0.1602. More work is needed to understand the pattern in these comparisons. 
    \item Absolute value of Negative contribution({\itshape neg}) within foreground also decreased in TL+KD in all cases, except for the cat, dog, frog, and horse. The p-value by a paired Wilcox test is not significant. We note that out of four exceptions, cats and dogs are the classes that were confused the most during the prediction and have the lowest two recalls in the confusion matrix Figure~\ref{fig:conf_matrix}. 
\end{enumerate}
}
For Observations 3, 4, and 5, we are not confident about the reason why some classes like Cat, Dog, etc. are not consistent compared to the rest for negative contributions and background. We believe this has to do with the exact samples that we selected for this analysis, where the backgrounds of those classes have some resemblances to the remaining classes. For example, as described before, the background of cats is somewhat visually similar to that of dogs, like houses. To test this hypothesis, we repeated the same process with different samples for classes(cat, dog, deer, horse, frog) that seemed inconsistent with Observations 3 and 4 in~\ref{observations_interpret_quant_tab}. We first randomly selected different samples for each four classes and generated a segmentation map for each class in Figure~\ref{fig:segmentation_process2} as in Figure~\ref{fig:segmentation_process}. 


We then calculated the values of {\itshape pos}, {\itshape neg} and {\itshape diff} for the foreground and background and their ratios to compare and quantify the interpretation of SHAP between the TL+KD model and the TL-only model for ambiguous classes from observations 4 and 5 in Table~\ref{tab:results_shap_abcd2}.

\begin{table}[!htbp]
\centering
\normalsize
\resizebox{\linewidth}{!}{%
\begin{tabular}{|l|c|c|c|c|c|c|c|c|c|c|c|c|c|c|c|} 
\hline 
\multicolumn{1}{|c|}{} & \multicolumn{6}{c|}{TL Only} &  \multicolumn{6}{c|}{TL+KD} & \multicolumn{3}{|c|}{Changes in Contribution} \\
\cline{2-16}
\multicolumn{1}{|c|}{Class} & \multicolumn{3}{c|}{foreground (inside)} & \multicolumn{3}{c|}{background (outside)} & \multicolumn{3}{c|}{foreground (inside)} & \multicolumn{3}{c|}{background (outside)}  & \multirow{2}{*}{$\frac{A}{B}$} & \multirow{2}{*}{$\frac{C}{D}$}	& \multirow{2}{*}{$\frac{A/B}{C/D}$} \\
\cline{2-13}
\multicolumn{1}{|c|}{} & \multicolumn{1}{c|}{{\itshape pos}} &  \multicolumn{1}{c|}{{\itshape neg}} &  \multicolumn{1}{c|}{{\itshape diff}$(A\uparrow)$} & \multicolumn{1}{c|}{pos} &  \multicolumn{1}{c|}{neg} &  \multicolumn{1}{c|}{diff$(C\downarrow)$} &\multicolumn{1}{c|}{pos} &  \multicolumn{1}{c|}{neg} &  \multicolumn{1}{c|}{diff$(B\uparrow)$} & \multicolumn{1}{c|}{pos} &  \multicolumn{1}{c|}{neg} &  \multicolumn{1}{c|}{diff$(D\downarrow)$}  & &&\\
\hline 
cat   &               24765 &              -21466 &                 3299 &                 4237 &                -4479 &                  -242 &              49298 &             -43424 &                5874 &                7153 &               -7396 &                 -243 &  0.56 &  1.00 &   0.56 \\
deer  &               12288 &               -9765 &                 2523 &                19005 &               -16255 &                  2750 &              16241 &             -12929 &                3312 &               18028 &              -15376 &                 2652 &  0.76 &  1.04 &   0.73 \\
dog   &               17579 &              -14367 &                 3212 &                 6619 &                -6181 &                   438 &              32591 &             -21485 &               11106 &                7521 &               -5459 &                 2062 &  0.29 &  0.21 &   1.36 \\
frog  &               11045 &               -8494 &                 2551 &                 8597 &                -7966 &                   631 &              12914 &              -9538 &                3376 &               10104 &               -8025 &                 2079 &  0.76 &  0.30 &   2.49 \\
horse &               16842 &              -12207 &                 4635 &                 9709 &                -9219 &                   490 &              27939 &             -18069 &                9870 &               13364 &              -11887 &                 1477 &  0.47 &  0.33 &   1.42 \\
\hline
\end{tabular}
}
\caption[SHAP contribution analysis for additional five winning class samples]{SHAP contribution analysis for additional five class samples. \FIX{All values are rounded to the nearest integer}. {\itshape diff} represents the difference in absolute values of the corresponding {\itshape pos} and {\itshape neg} of the same side. The upward arrow represents that the higher difference is better for the winning class. Changes in contribution are the ratio of differences($A,B,C \& D$) from earlier columns. The main observations from this table are listed in Section~\ref{observations_interpret_quant_tab2}}
\label{tab:results_shap_abcd2}
\end{table}

Additional observations from Table~\ref{tab:results_shap_abcd2} are listed below.
\begin{enumerate}\label{observations_interpret_quant_tab2}
    \item Observations 1 and 2 of observation ~\ref{observations_interpret_quant_tab} above are consistent and correct in these additional samples as well.
    \item Observations 3, 4, and 5 were still inconsistent with these additional samples as well. 
\end{enumerate}

In summary, there were some changes in the observation from the previous one; however, the ambiguity remains after investigating additional random samples for selected classes. Therefore, there is much room for research to investigate other unclear observations. Another unclear open question we faced while evaluating this table was should a model look at the background to classify an image. From a trustworthy machine learning perspective, we should only look at the foreground of an image that avoids the background because a model could be attacked with adversarial examples with confusing backgrounds and forced to produce wrong predictions. On the other hand, from the context perspective, if we need to differentiate a ship with a visually similar truck image from a small 32*32 pixel image like CIFAR10, their background sea vs. road provides context to classify the image correctly. Due to this ambiguous question, we do not take the background scores from the tables above in the equation to evaluate this hypothesis. 

In summary, because of the consistent pattern of higher positive contribution in the foreground for a model with KD in observations 2 and 3 in all cases, we treat this hypothesis that TL+KD provides more contribution on foreground image features as correct and accepted when measured by SHAP contribution. 

\section{Does TL+KD achieve similar validation accuracy faster in fewer training epochs than TL-only?}
\label{res_faster_convergence}Our other hypothesis was to see if TL+KD trains faster than TL only model in fewer iterations for similar validation accuracy. To assess that, we look at the convergence graph of the models in both scenarios and compare the time it takes to achieve a certain validation performance, such as accuracy. 

\begin{figure}[!htbp]
\centering
\includegraphics[width=\linewidth]{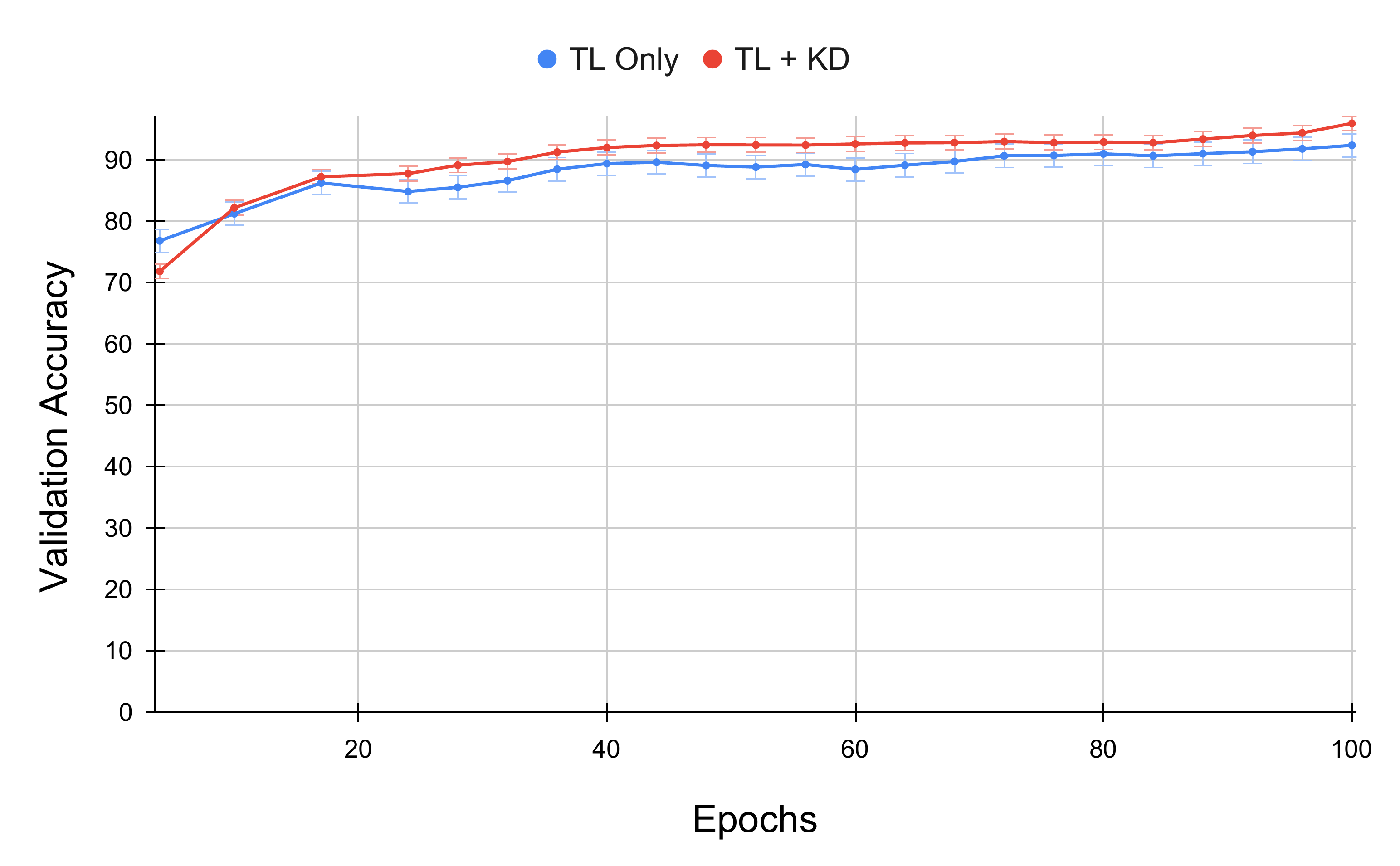}
\caption[Convergence speed comparison of TL with and without KD]{Convergence speed comparison of TL Only and TL+KD. The red line (TL+KD) reaches high accuracy thresholds such as 90\% before the TL-only counterpart. Interestingly, the TL+KD model starts out with lower accuracy and optimizes faster.}
\label{fig:faster_convergence}
\end{figure}

Figure~\ref{fig:faster_convergence} shows two line graphs that describe the progression of validation accuracy for the TL and TL + KD models.

Figure~\ref{fig:faster_convergence} shows the accuracy of the validation in columns and the training epochs in rows. The blue line refers to TL only without any teacher and the red line refers to TL + KD with ViT as a teacher. The experiment setup is described in Section~\ref{sec:ex_training} and is the same as the setup used for the previous hypothesis~\ref{res:improves}. The mean standard deviation of 1.94\% and 1.25\% calculated from 3 different runs with different seeds are included in the plots for TL only and TL+KD, respectively.  

On inspection, we see that both red and blue lines start at relatively high accuracy for a few epochs. This is because the pre-trained weights of the feature extractor learned from the ImageNet dataset seem quite useful to be used by newly appended classifier layers. For that reason, within a few epochs, both models reach 70+ accuracy values. Another interesting thing we see is that TL+KD has a slower start and has lower accuracy than the TL-only counterpart, opposing our hypothesis in initial epochs. We believe this is due to the additional term of KL divergence loss in knowledge distillation as specified in Equation~\ref{eq:kd_loss}. Because the initial accuracy is low and there is a remarkably high initial difference in the softened logits of the teacher network and student network, it is difficult for the student network to learn faster during initial epochs. However, we see that TL+KD improves pretty quickly and surpasses the TL-only version of the model. Once the gap decreases, we see a faster convergence of the student network with KD, which supports our hypothesis. In this particular Figure~\ref{fig:faster_convergence}, the student with ViT as a teacher(KD) reaches higher accuracy values such as 90\% validation accuracy at only 35 epochs, while vanilla training without any teacher takes almost double epochs, i.e. 70, to reach that same accuracy.  We also see similar consistent results for all other accuracy values higher than ~85\% validation accuracy. This supports the hypothesis that Knowledge Distillation helps student networks converge faster for high accuracy but fails or even hurts during the initial stage of training.

\section{Is TL+KD validation accuracy better than TL-only even after adding different training complexities?}
\label{res_training_frac}
\begin{figure}[!htb]
\centering
  \includegraphics[width=0.9\linewidth]{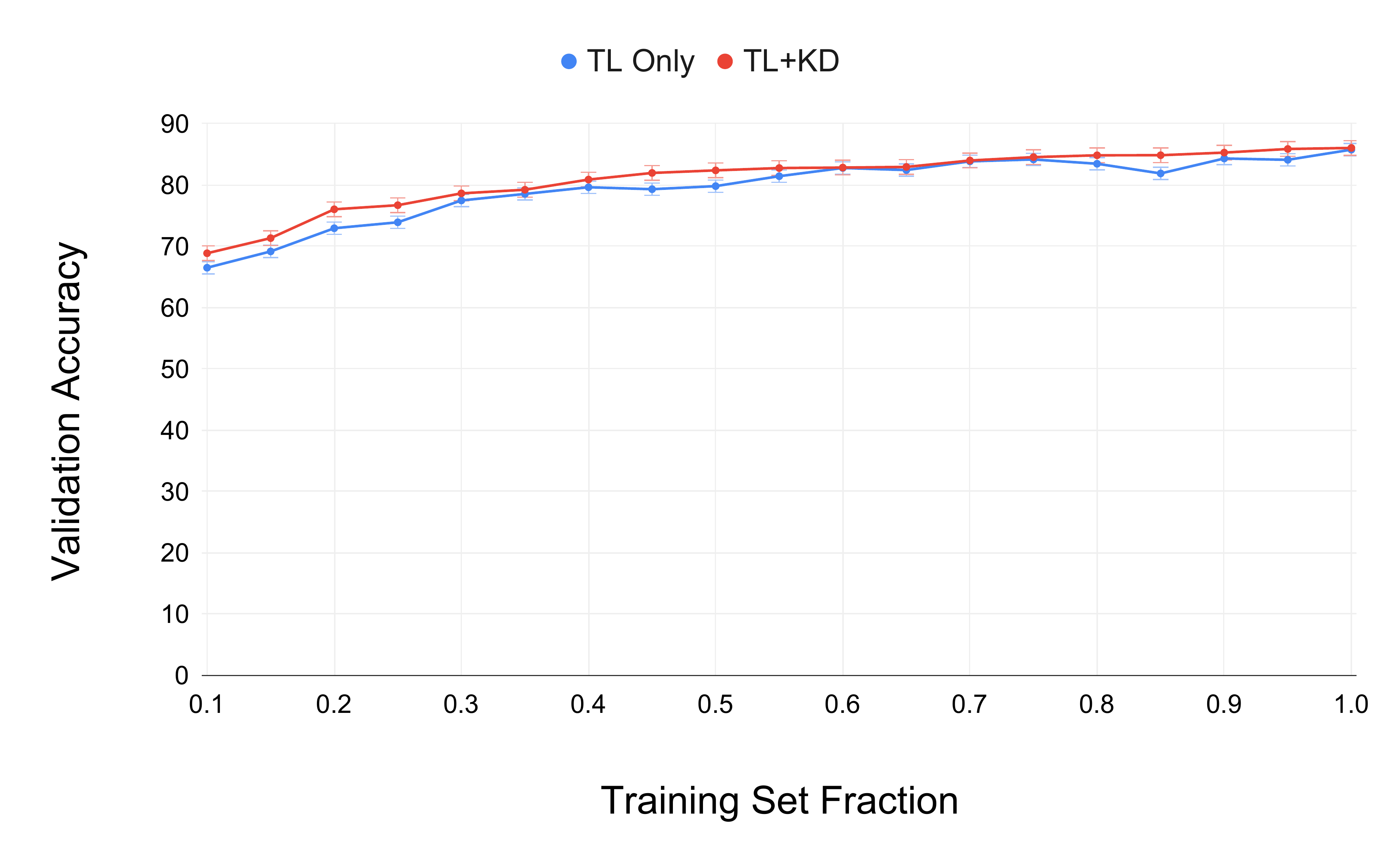}
\caption[Final Validation accuracy changes for different training fractions]{Changes in the accuracy of the final validation for different training fractions. KD (red) is consistently higher than TL only (blue), suggesting that TL+KD is better for added complexities and is more prominent for more complex setups (training fraction).}
\label{fig:training_frac}
\end{figure}

While exploring the effects of knowledge distillation on transfer learning, we saw certain gains in models trained with the full dataset. We wanted to see if these gains are consistent even when the training setup is changed and made more complex. We hypothesized that with a strong teacher network, the student network should be able to learn and perform better despite the added complexities. For this reason, we first limited the amount of training data to which the model has access by limiting the training fraction parameter. The experimental setup is described in Section~\ref{ex_complexity}. With this setup, we could explore the effect of KD on transfer learning under the added complexity of limited data. Figure~\ref{fig:training_frac} shows the performance of the TL-only model 'ResNet18' and TL+KD with soft labels from the WideResNet teacher. 

In Figure~\ref{fig:training_frac}, the X-axis represents the fraction parameter of the training set, while the Y-axis reports the final validation accuracy after training. Upon inspection, we observe that the red line (KD) model is consistently higher than the blue line (TL only) model, and we see higher gains for more complex problems. 

Note that the graph and scores are slightly different from those of other experiments. The highest validation accuracy for both TL and TL+KD is consistently less than $90\%$. The gains in TL+KD do not seem significant here. In Figure~\ref{fig:training_frac}, we note that even with a dataset of only 10\% (500 samples for each class), both models consistently have a good accuracy score of almost 70\%. We believe this is because the CIFAR10 dataset is easy to learn from. For this reason, we tried other ways to add more complexities. 

As described in Section~\ref{ex_complexity} the reason is that this experiment differs from the ones reported above. The teacher 'WideResNet' and some of the hyperparameters for this experiment are different from the experiments reported in other sections of this chapter. The reason we changed the teacher for other experiments is that the stronger teacher and hyperparameters used there yielded higher accuracy scores comparable to those of the literature. We observe a notable dependence of the network on hyperparameters, such as image sizes and learning rates. This is why the maximum accuracy achieved by the ResNet (both student and teacher) networks here is limited to approximately 86\%, while it is 90\%+ in other sections. However, since we are more motivated to see how KD affects TL than what the maximum TL performance is, we also report these scores in this section. 

\begin{figure}[!htb]
\centering
  \includegraphics[width=\linewidth]{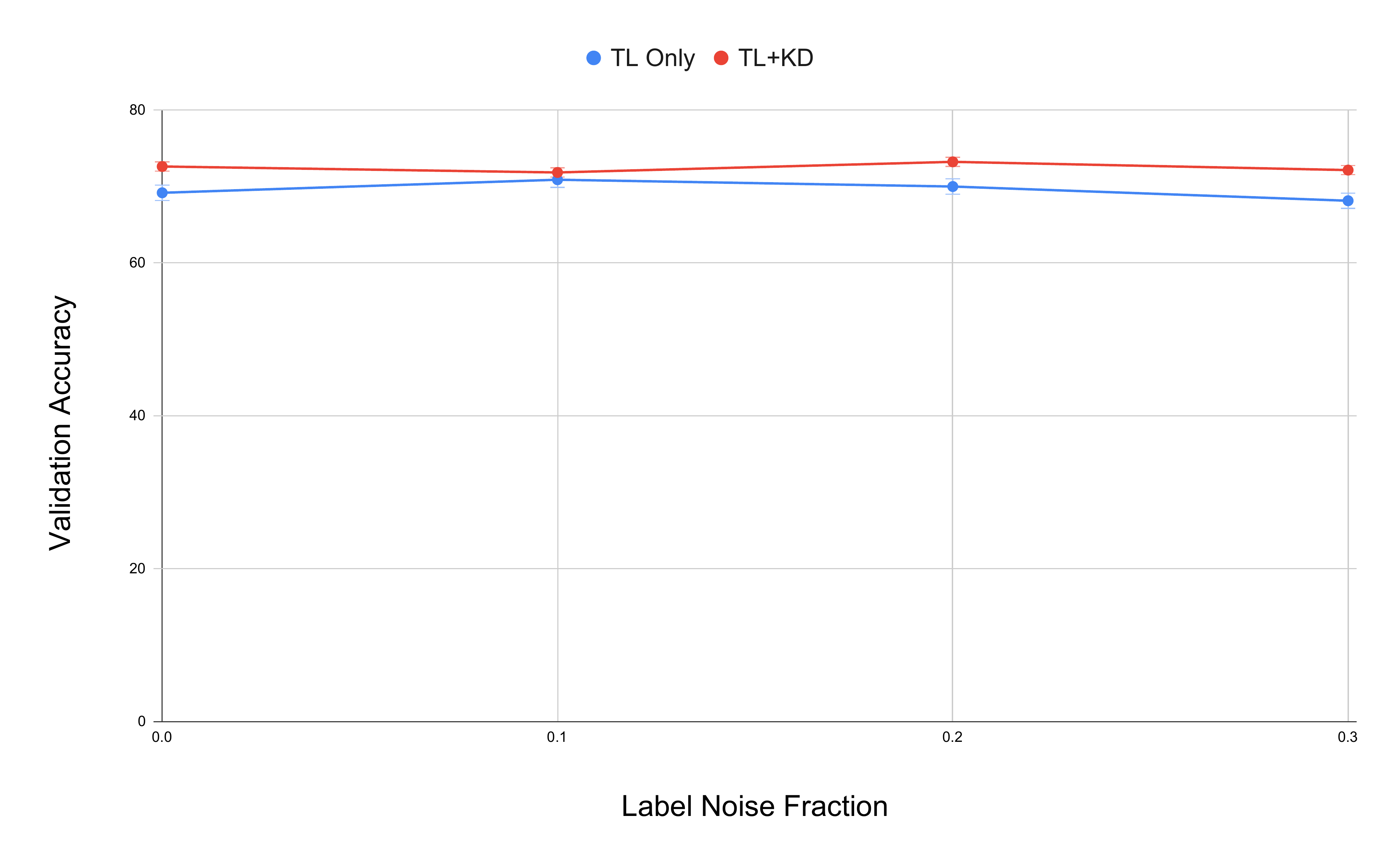}
\caption[Final Validation accuracy changes for different label noise levels]{Changes in the accuracy of the final validation for different levels of label noise. KD(red) is consistently higher than TL only (blue), suggesting that TL+KD is better for added complexities and is more prominent for more complex setups (label noise).}
\label{fig:label_noise}
\end{figure}

The first alternative was to explore the noise of the label as described in Section~\ref{ex_complexity}. To add additional complexity and to perform the experiment faster, we performed these experiments with only a subset of the training set (15\% for Figure~\ref{fig:label_noise}). In Figure~\ref{fig:label_noise}, the X-axis represents the label noise fraction parameter, while the Y-axis reports the final validation accuracy after training. Upon inspection, we observe that the red line (KD) model is consistently higher than the blue line (TL only) model, and we see higher gains for more complex problems. We see that with added label noise, the network can still maintain the original accuracy values. One thing to note is that the red line is still higher than the blue line, indicating that the TL+KD is consistently superior to the TL-only counterpart. 

Since the performance was not degrading enough with the training fraction, and we wanted to introduce a real-world-like distortion method instead of randomizing labels, we next tried other different ways to corrupt the images next. The primary motivation to search for such corruption methods was again to push the accuracy scores of the model down for the CIFAR10 image classification. One can also think of these manipulations as trying to reflect "real" data instead of overly clean, well-delineated test datasets and exploring how KD affects such scenarios. After trying Gaussian blurring and noise, QuarterBlack corruption, etc., we finally settled on the CenterBlack\label{res_complexity} corruption method as described in Section~\ref{ex_complexity} since this parameter could control the amount of foreground to which the model has access. The QuarterBlack corruption method, on the other hand, only blocked a quarter of images and always exposed parts of the images that were enough to learn good representations and did not add much complexity to the model. 

\begin{figure}[!htb]
\centering
\begin{subfigure}{.795\textwidth}
  \centering
  \includegraphics[width=\textwidth]{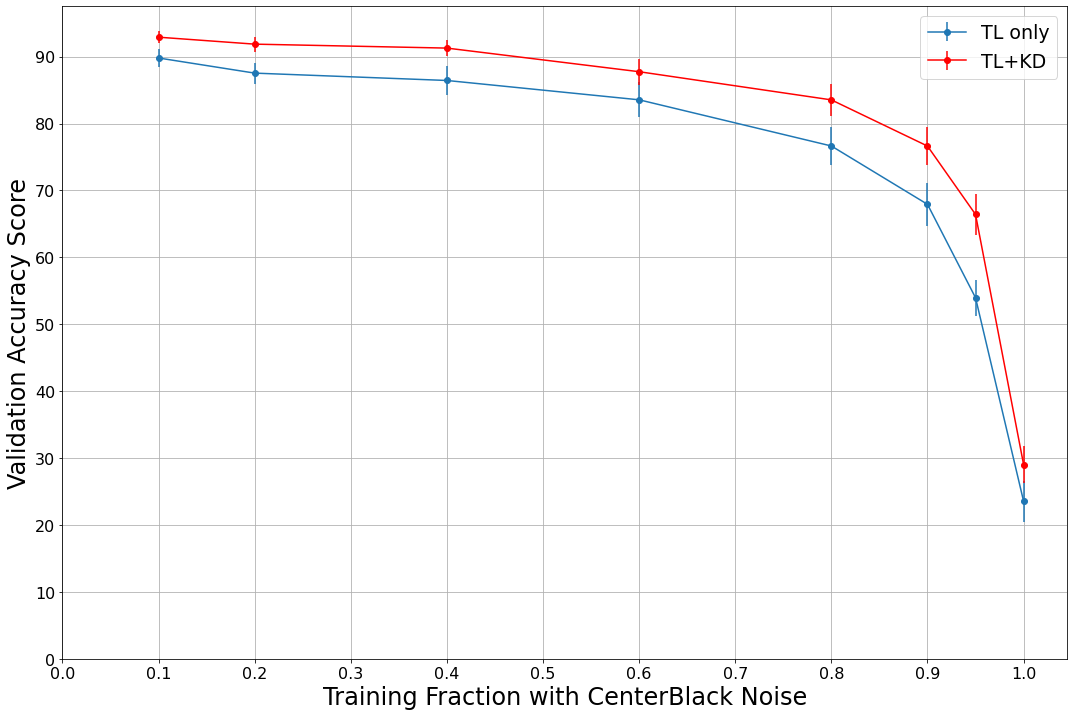}
  \caption[]{}\label{fig:kd_lowerval_both}
\end{subfigure}%
\vspace{0pt}
\begin{subfigure}{.795\textwidth}
  \centering
  \includegraphics[width=\textwidth]{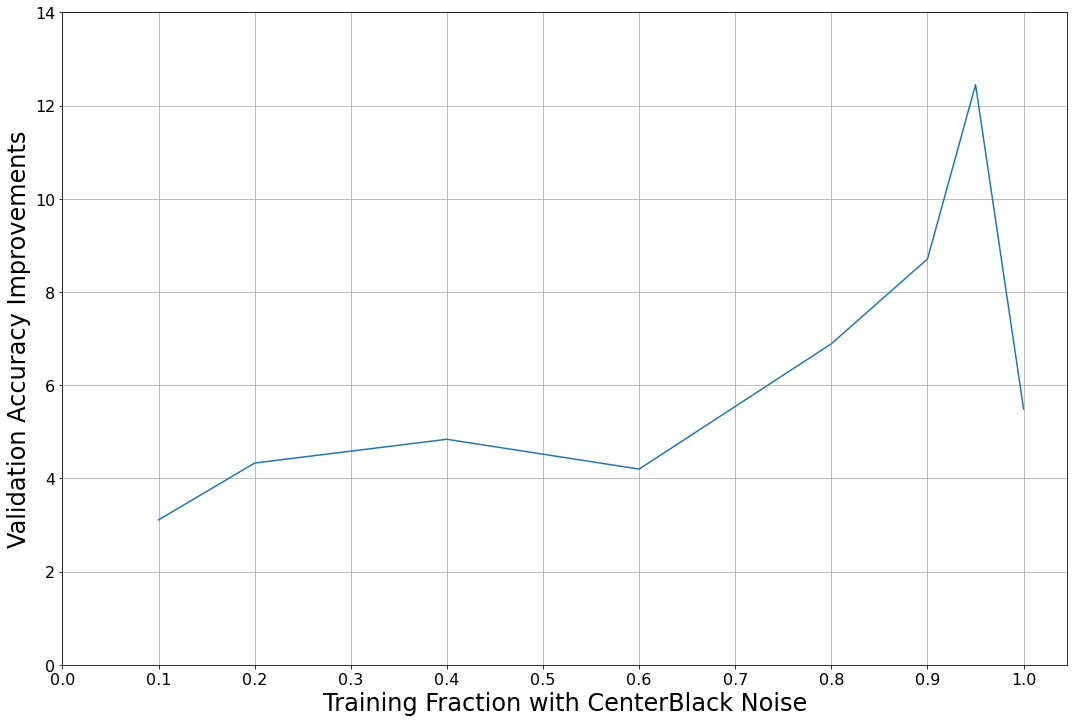}
  \caption[]{}\label{fig:kd_lowerval_improv}
\end{subfigure}
\caption[Accuracy Comparison for a different fraction of training images with CenterBlack corruption]{Accuracy comparison for different fractions of training images with CenterBlack corruptions with patches of random sizes between 200px and 224px (a) Accuracy comparison of two models, TL+KD is consistently better and has more gains for added difficulties at lower accuracy scores. (b) Accuracy improvements in TL+KD calculated by subtracting the validation scores of TL+KD with TL only from (a), we see a notable gain at higher complexity around 95\% training images CenterBlack noise. However, the gains with maximum corruption seem consistent with those when they are minimum.}\label{fig:kd_lowerval}
\end{figure}

\begin{figure}[!htb]
\centering
\begin{subfigure}{.77\textwidth}
  \centering
  \includegraphics[width=\textwidth]{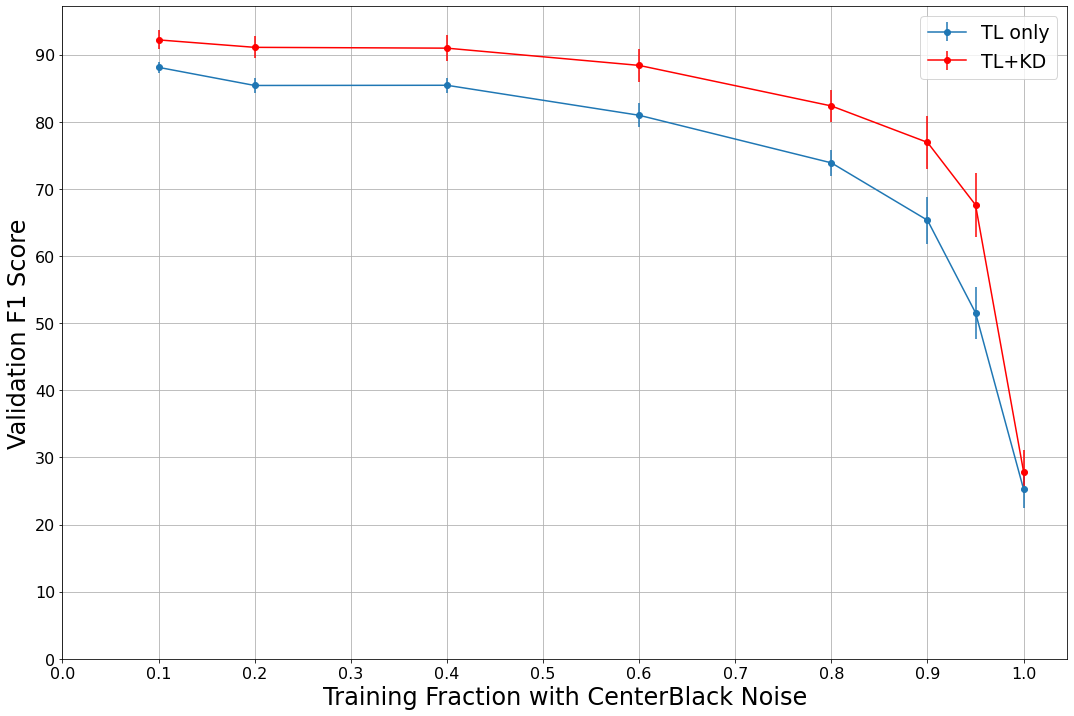}
  \caption[]{}\label{fig:kd_lowerval_both_f1}
\end{subfigure}%
\vspace{0pt}
\begin{subfigure}{.77\textwidth}
  \centering
  \includegraphics[width=\textwidth]{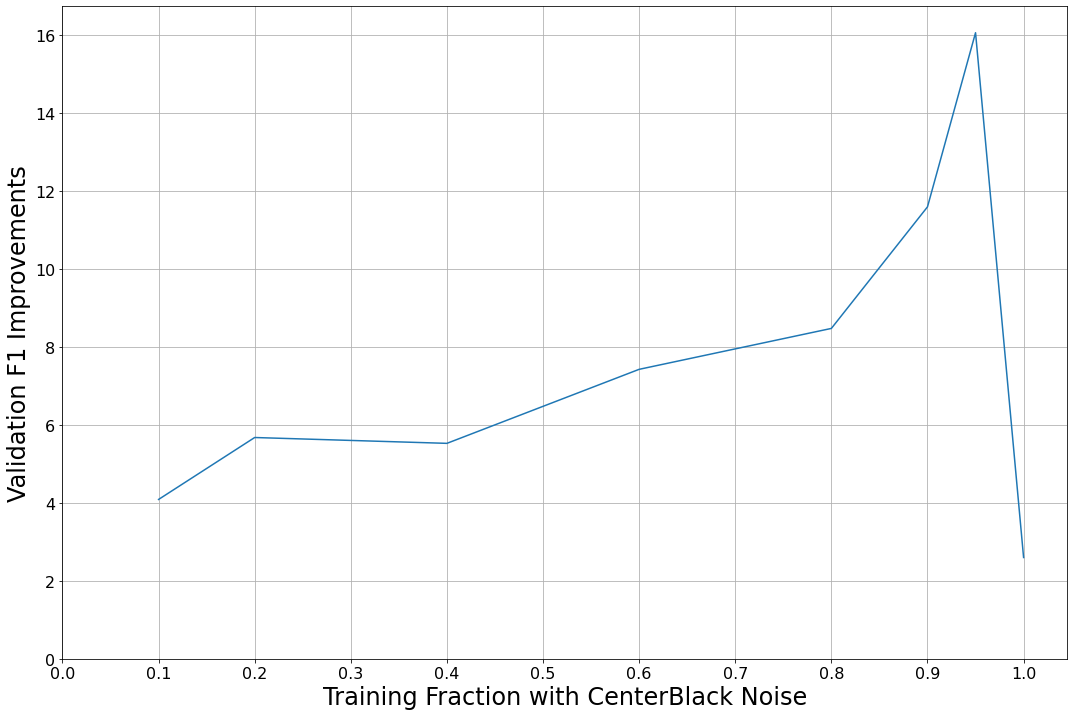}
   \caption[]{}\label{fig:kd_lowerval_improv_f1}
\end{subfigure}
\caption[F1 score Comparison for a different fraction of training images with CenterBlack corruption]{Comparison of the F1 score for different fractions of training images with CenterBlack corruptions with random size patches between 200px and 224px (a) Comparison of the F1 score of two models, TL+KD is consistently better and has more gains for added difficulties at lower f1 scores. (b) Improvements in the TL+KD F1 score in TL calculated by subtracting the TL+KD f1 validation scores with TL only from (a), we see a notable peak in the gain in the f1 score at higher complexity around 95\% training images CenterBlack noise. However, the gain with maximum corruption appears to be lower and is consistent with that when it is minimum. All these observations are consistent with Figure~\ref{fig:kd_lowerval}.}
\label{fig:kd_lowerval_f1}
\end{figure}

In Figure~\ref{fig:kd_lowerval} and Figure~\ref{fig:kd_lowerval_f1}, 
we visualize the gains of knowledge distillation on the same baseline model with added complexity. Here, we introduce image corruption/CenterBlack patches described in Subsection ~\ref{centerblack} to move the Validation Accuracy to low values. In this particular setup, described in Section~\ref{ex_complexity}, we observe that knowledge distillation is also effective with a more corrupt dataset and is more prominent for lower validation accuracy. Figure~\ref{fig:kd_lowerval} shows the progression of knowledge distillation gains when applying CenterBlack patches in different fractions ranging from 10\% to 100\% in the available training set. Please note that CenterBlack image corruption is applied only to training samples. 

We see an interesting behavior shown by the network in this corruption method when the fraction is 95\%. It is seen that TL+KD provides the maximum gain (i.e., +12.45\% accuracy and 13.07\% F1 score) when CenterBlack corruption is applied to 95\% of the available training samples. It is also interesting that TL+KD provides similar improvements to Transfer Learning without CenterBlack when all 100\% of the training images are corrupted with CenterBlack augmentation. To add complexity and push performance to lower accuracy and f1 space, we only take 10\% of the available training samples, just as for label noise. We verify these conclusions through multiple runs by training the network with different random seeds(1, 7, 42). Apart from the conclusion being consistent in multiple runs, it also provided some insight into the high deviation of accuracy and f1 scores with different fractions. Due to such a high standard deviation, we cannot expect the exact value of 12\% gains. However, with multiple runs, it consistently shows that the gain in the 95\% fraction of data with CenterBlack has the highest gains among other fractions. 

From all these experiments, we can observe that TL+KD does not behave the same way for different complexities of training. TL+KD behaves consistently for simpler corruptions and complexities but provides significant gains for high complexities. We tested a few types of corruption to support this hypothesis. Still, we believe that we can try various image-corrupting methods to explore and learn more about the effects of TL+KD with those complexities. Some of the work we think would be insightful for future work is exploring how the pixels on the edges of images affect the predictions with huge CenterBlack patches and how different filters like Low Pass, Band Pass, High Pass, etc. affect the learning process, etc.





\chapter{Conclusion}
\label{chap:conclusion}In this work, we present a study of the effects of knowledge distillation on transfer learning in an image classification problem exploring some research questions. All our conclusions are for transfer learning and transfer learning with knowledge distillation in an image classification task. Our summary of the findings based on each hypothesis of Section~\ref{hypothesis_list} is as follows.
\begin{itemize} \label{hypothesis_list_conclusion}
    \item Hypothesis 1: TL+KD architecture improves image classification over the TL architecture on the CIFAR10 dataset.
        \begin{description}
            \item[\hspace*{19pt}] We investigated this hypothesis by evaluating the performance of the student model on the same CIFAR10 image classification task, first with vanilla transfer learning only and then transfer learning with knowledge distillation model. We evaluated both models on multiple metrics beyond accuracy such as precision, recall, and F1 score, and with a confusion matrix and SHAP interpretability. We saw improvements in metrics (more than +3\%) for all accuracy, precision, recall, and F1 scores with the TL+KD model over the vanilla TL model consistently across three runs with different seeds. \sm{Using the confusion matrix, we saw that previously incorrect predictions were consistently predicted correctly with the help of TL+KD: there were higher numbers in diagonal entries and smaller ones  off-diagonal}. With SHAP interpretations, we observed visually that the model looked at relevant features within the foreground with more distinctive contribution patterns.
            Each of these evaluation methods infers that the performance of TL+KD is better than that of TL-only training, validating this hypothesis successfully. 
        \end{description}
    \item Hypothesis 2: TL+KD provides more contribution on foreground image features measured by SHAP contribution than TL only.
        \begin{description}
            \item[\hspace*{19pt}] We investigated this hypothesis by differentiating the foreground and background of the images and evaluating the interpretability scores of the image classifier models separately with and without KD. Across all samples taken for all available CIFAR10 classes, we found a higher difference in the absolute value of the positive and negative contribution of the model within the foreground in TL+KD. The positive/negative contribution for the background section was not consistent and varied with the samples selected. Our test validates this hypothesis for the foreground section but demands further investigation of factors affecting the background portion of images. 
        \end{description}
    \item Hypothesis 3: TL+KD achieves similar validation accuracy faster with fewer training epochs than TL only model.
        \begin{description}
            \item[\hspace*{19pt}] We investigated this hypothesis by looking at the convergence of the same image classifier student with and without KD and assessed the iterations taken to achieve similar accuracy thresholds. TL-only model reached an accuracy threshold slower than the TL+KD setup for higher accuracy values that validate the hypothesis. However, during initial epochs, TL+KD showed lower performance, rejecting the proposed hypothesis. However, TL+KD improves with additional iterations faster and surpasses the TL-only version fairly quickly validating the hypothesis, as shown in Figure~\ref{fig:faster_convergence}. We believe that one possible reason for this might be the high difference in capacity between the teacher and student network, which hurts the network during initial runs.
        \end{description}
    \item Hypothesis 4: TL+KD improves validation accuracy even after adding training complexities such as training with a fraction of data or corrupted images.
        \begin{description}
            \item[\hspace*{19pt}] \sm{We investigated this hypothesis by training the same student image classifier model with two additional complexities:  training with a fraction of data and training with corrupted data, for both TL and TL+KD}. Looking at the model's performance with different complexities, we saw consistent gains in validation accuracy for all complexity methods, which validates the hypothesis. 
        \end{description}
\end{itemize}
\chapter{Future Work}
\label{chap:future_work}
Some of the possible future iterations of this work could be summarized as follows.
\begin{enumerate}
    \item Since our methodology is designed to be domain-agnostic, we can also apply the same framework to explore the effects of KD on transfer learning in other modalities like text, genomics, medical imaging, etc. 
    \item We use a few types of complexity and corruption methods to test our Hypothesis 4 which is not exhaustive. However, we can try a myriad of other image corruption methods, such as applying different filters, other types of augmentation, etc., and see how KD behaves in those scenarios.
    \item We currently use SHAP interpretations for all our interpretability needs. We can also explore how our conclusions compare with the use of various other types of interpretation techniques mentioned in Section~\ref{intro_interpret}.
    \item Our experiments were limited by compute resources and time. A full hyperparameter sweep with grid search or Bayesian optimization could be helpful to achieve optimal performance for each model. Using other datasets for transfer learning instead of CIFAR10 could also be an interesting direction to explore. 
    \item We primarily explored the response-based offline knowledge distillation method for our experiments. Exploring other types of knowledge distillation, as mentioned in Section~\ref{related_work_kd}, and how they behave could also be an interesting research find.
\end{enumerate}


\bibliographystyle{unsrt}
\bibliography{references}

    \appendix
    \chapter{Code}
    All the experiments done for this thesis were implemented in Python programming language. The deep learning experiments and optimizations were done in \textit{PyTorch}. Models and data augmentation were done with \textit{TorchVision} library. Image manipulations were done with \textit{imageio} and \textit{OpenCV} library. Data manipulation and splits were done with \textit{Scikit-Learn} library. Plotting and visualizations were done with \textit{Matplotlib}, \textit{Seaborn} and Google \textit{draw.io}.  Monitoring, MLOps, and dashboard were built with \textit{Weights and Biases}. The details of these libraries and their version are given below. 
\begin{Verbatim}[frame=single]
  Library and Version                    Task (same onwards)
--------------------------------------------------------------------
- python==3.7                          # programming language
- opencv-python==4.6.0.66              # image manipulation
- imageio==2.16.2
- pillow==9.0.1                       
- scikit-image==0.19.2
- shap==0.41.0                         # interpretation
- matplotlib==3.5.1                    # plotting, visualization
- seaborn==0.11.2                 
- numpy==1.21.5                        # data manipulation
- scikit-learn==1.0.2
- scipy==1.7.3
- pandas==1.1.5               
- torch==1.11.0                        # model development
- torchvision==0.12.0
- torchsummary==1.5.1
- tensorflow==2.9.1     
- wandb==0.12.11                       # MLOps
- via==2.0.11                          # data collection and tagging
- pycocotools==2.0.4   
\end{Verbatim}
All codes for the experiments and analysis in this thesis will be available at:\\ 
\href{https://github.com/Sushil-Thapa/distill-transfer}{https://github.com/Sushil-Thapa/distill-transfer}
    

    
   
   
   
   
%
\copyrightpage

\end{document}